\documentclass{article}

     \PassOptionsToPackage{numbers, compress}{natbib}
\usepackage[preprint]{neurips_2026}

\usepackage[utf8]{inputenc} 
\usepackage[T1]{fontenc}    
\usepackage{hyperref}       
\usepackage{url}            
\usepackage{booktabs}       
\usepackage{amsfonts}       
\usepackage{nicefrac}       
\usepackage{microtype}      
\usepackage{xcolor}         

\usepackage{graphicx}
\usepackage{subcaption}
\usepackage{epigraph}
\usepackage{amsmath}
\usepackage{lipsum}
\usepackage{tikz}
\usepackage{cleveref}
\usepackage[table]{xcolor}
\usepackage{wrapfig}
\usepackage{xcolor}
\usepackage[misc]{ifsym}

\definecolor{plblue}{HTML}{527EC7}
\definecolor{regreen}{HTML}{67A374}
\definecolor{mtpurple}{HTML}{8250BB}

\newcommand{\benchmarkname}{{{\color{magenta}V}{\color{cyan}I}{\color{orange}B}{\color{teal}E}}}

\def\eg{\emph{e.g.}} 
\def\ie{\emph{i.e.}} 
 
\def\etc{\emph{etc.}}

\def\etal{\emph{et al.}}

\definecolor{1}{RGB}{0,104,55}
\definecolor{2}{RGB}{18,138,73}
\definecolor{3}{RGB}{110,192,100}
\definecolor{4}{RGB}{254,255,190}
\definecolor{5}{RGB}{254,218,134}
\definecolor{6}{RGB}{234,87,57}
\definecolor{7}{RGB}{165,0,38}

\newcommand{\1}{\cellcolor{1} 1}
\newcommand{\2}{\cellcolor{2} 2}
\newcommand{\3}{\cellcolor{3} 3}
\newcommand{\4}{\cellcolor{4} 4}
\newcommand{\5}{\cellcolor{5} 5}
\newcommand{\6}{\cellcolor{6} 6}
\newcommand{\7}{\cellcolor{7} 7}

\newcommand{\pertext}[1]{\textcolor{plblue}{\textbf{#1}}}
\newcommand{\restext}[1]{\textcolor{regreen}{\textbf{#1}}}
\newcommand{\mantext}[1]{\textcolor{mtpurple}{\textbf{#1}}}

\title{Quo Vadis, Visual In-Context Learning? \\ 
A Unified Benchmark Across Domains and Tasks}

\author{%
  Pradnya Halady\thanks{equal contribution} $\phantom{00}$ Jiale Wei $\phantom{00}$ Zdravko Marinov $\phantom{00}$ Alexander Jaus $\phantom{00}$ Simon Rei{\ss}$^*$$^{\text{\Letter}}$\\
  Karlsruhe Institute of Technology \\ \Letter~\texttt{simon.reiss@kit.edu} \\
}

\begin{document}

\maketitle

\vspace{-1cm}

\begin{figure}[h!]
    \centering
    \includegraphics[width=\textwidth,page=4]{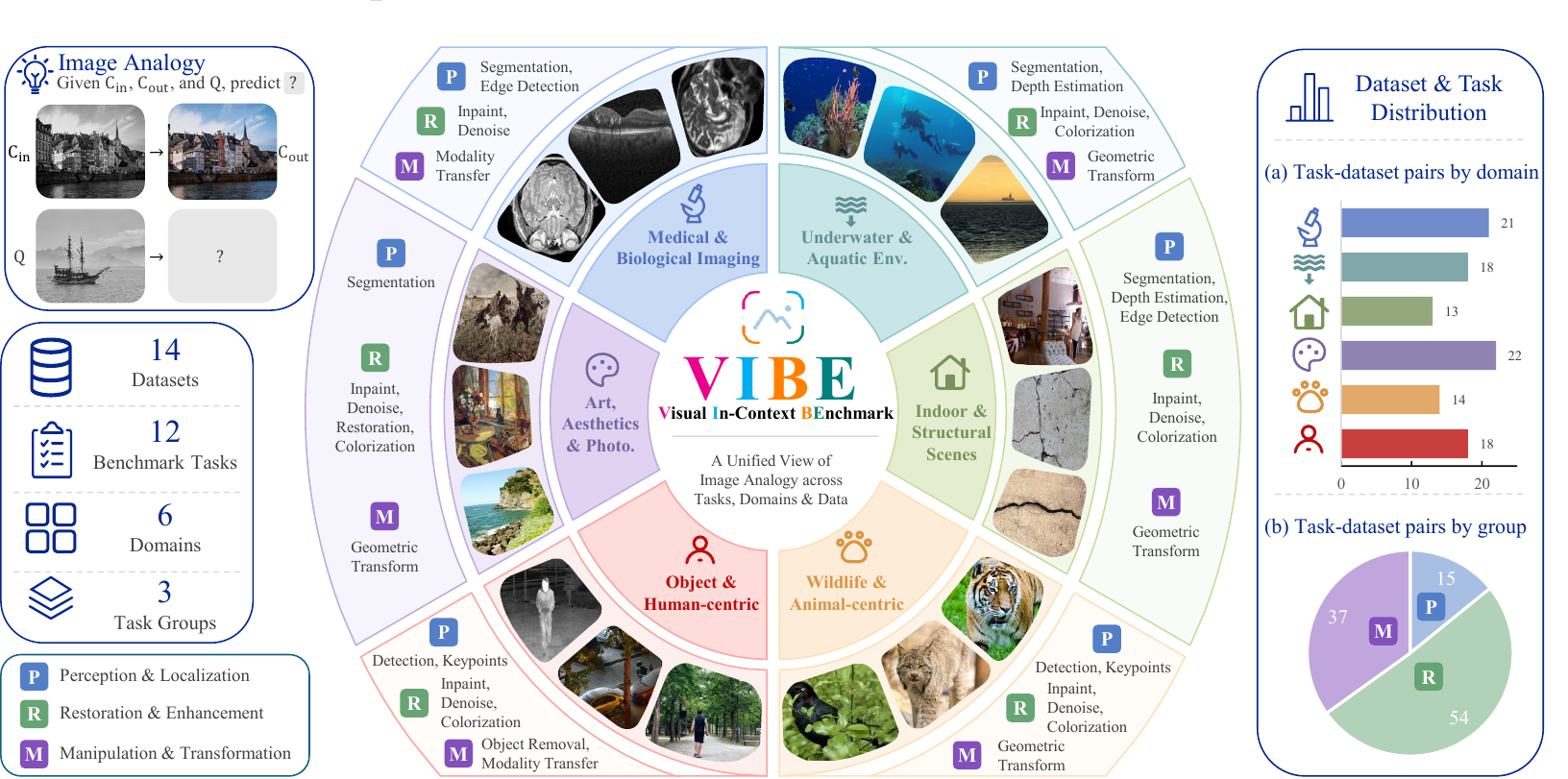}
    \caption{Overview of our {\color{magenta}V}isual {\color{cyan}I}n-Context {\color{orange}B}{\color{teal}E}nchmark (\benchmarkname) with diverse tasks and domains.}
    \label{fig:placeholder}
\end{figure}

\begin{abstract}
    Visual in-context learning has been proposed as a pathway towards dynamic models that can generate predictions based on a provided context and thereby can adapt to new vision tasks at test-time.
    Yet, the evaluation of the \emph{adaptation capabilities} of these models has been limited to narrow setups that mainly mirror tasks or image domains from pre-training for which real adaptation is not required.
    We address this gap by constructing a broad {\color{magenta}V}isual {\color{cyan}I}n-Context {\color{orange}B}{\color{teal}E}nchmark (\benchmarkname) with a focus on diverse imaging domains and a wide range of tasks.
    With this, we are able to get a much clearer picture of the adaptive capabilities of visual in-context models when faced with new image- and task distributions.
    We stress test six models on $14$ datasets and $12$ tasks (in total, we explore $106$ dataset-task combinations) and compare them under a unified, reproducible evaluation protocol, in an one-shot setting.
    Our evaluation uncovers key insights on the state of visual in-context learning, including limitations, systematic failure modes and promising directions.
    To foster broader evaluation, we openly release our \benchmarkname~toolkit (\href{https://github.com/Simael/visual-in-context-benchmark}{source code}).

\end{abstract}

\section{Introduction}
In the bygone decade, much of the success in computer vision research centered around using deep neural networks~\cite{krizhevsky2012imagenet,long2015fully,girshick2015fast,ronneberger2015u} to address new tasks through supervised training on task-specific curated datasets.
While offering a generic pathway towards vision systems for different applications, a fundamental prerequisite for this paradigm is the access to large, labeled datasets~\cite{everingham2010pascal,deng2009imagenet,lin2014microsoft,cordts2016cityscapes}.

Advancing vision models towards requiring fewer labeled examples for adapting to new tasks, similar to humans~\cite{walker2014toddlers}, is still a fundamental open question.
Different prominent research strains have tried to address this, \eg, by self-supervised representation learning~\cite{vincent2008extracting,chen2020simple} and successive transfer learning with a small set of labeled examples~\cite{thrun1995learning,caruana1997multitask,donahue2014decaf}, semi-supervised learning where few labeled examples are used in conjunction with a large set of unlabeled samples~\cite{lee2013pseudo,arazo2020pseudo,sohn2020fixmatch}, or, few-shot learning~\cite{snell2017prototypical,finn2017model,ravi2017optimization}, where models converge with few examples, \eg, through meta-learning~\cite{813180}.

Recently, visual in-context learning (VICL) has been explored as a different pathway towards data-efficient adaptation~\cite{bar2022visual}.
VICL models are given a few examples for a new task (the context), as input to the model at test-time as well as a query image for which the new task shall be solved.
Thereby, the model needs to infer the task from the given context first and successively apply it to the query image to produce a fitting output.
This setup can be seen as solving a `visual analogy'~\cite{hertzmann2001imageanalogies} in the form:
\begin{equation} \label{eq1}
\begin{split}
C_{\mathrm{in}} &\to C_{\mathrm{out}} \\
Q &\to~~?\phantom{_{\mathrm{out}}}
\end{split}
\end{equation}
where $C_{\mathrm{in}}$ and $C_{\mathrm{out}}$ are the contextual images, that showcase a transformation by example, which then has to be carried out on the query image $Q$ in order to fill the placeholder $?$ with a corresponding output image.
This is a significant shift as it views models as inherently contextual and thereby turning them into dynamic analogy-making systems rather than static, task-specific models.
As visual analogies can already be defined with one example pair of images, data-efficient adaptation is at the center of this learning paradigm.
Interestingly, this view coincides with findings in human cognition, where analogies have been argued to be a driver of conceptualizations of wholly new situations~\cite{mitchell2021abstraction} and may even lie at the core of human thought~\cite{hofstadter2001analogy}.

While early on, the problem setting of image analogies has been explored for texture synthesis~\cite{hertzmann2001imageanalogies}, more recently, researchers have been striving for deep learning models that are capable of solving such visual analogies.
Sadeghi \etal~\cite{sadeghi2015visalogy} trained a model to score each image in a pool of candidate images in order to select the candidate image that fits the analogy pattern in the context best.
Similarly, visual analogies were explored in terms of Raven-style Progressive Matrices~\cite{raven1941standardization}, \ie, simple visual IQ tests~\cite{barrett2018measuring,hill2019learning}, where the deep models also scored candidate images.

While probing the contextual capabilities of vision models through scoring candidate images remains an important area of exploration today~\cite{born2026context,bitton2023vasr}, the next step in building deep models that solve visual analogies was to move on from selecting the correct image from a candidate pool towards generating the actual output image.
This was explored on visual analogies composed of simple geometric patterns by Webb \etal~\cite{webb2020learning} with a model that first encodes the context and then uses the encoding as condition for an image-generation model to produce the analogous output image.




The recent culmination of vision foundation models~\cite{rombach2022high,vit_visiontransformer,esser2021taming,he2022masked} that capture rich representations of images and observations on few-shot prompting in natural language processing~\cite{brown2020language}, have sparked new interest in visual analogies under the term visual in-context learning~\cite{bar2022visual} and have created space for imagination to capture more complex tasks.
Rather than operating on image textures~\cite{hertzmann2001imageanalogies}, geometric patterns~\cite{barrett2018measuring} or line drawings~\cite{lu2019seeing}, this time, the analogy-making models operate on computer vision tasks such as segmentation, colorization or denoising, which are rearranged into visual analogies.

In this new attempt at training data-efficient, adaptive vision models, the focus has mostly been on architectural designs for conditioning the models on context and query images, on the training strategy and the data for training these models.
Now that the community trained a wide bouquet of models~\cite{bar2022visual,bai2024sequential,wang2023images,oorloff2025stable,promptdiffusion,delvm,improv,reiss2025visual,czolbe2023neuralizer,schmidt2026static,negrini2025conquering}, what remains unexplored is an in-depth evaluation and analysis of their capabilities, limitations and failure modes.
Previously, evaluation has been limited to image distributions present in training and simplified task evaluation settings that conceal the extent to which models can draw contextual predictions rather than taking shortcuts (\eg, as observed for Raven’s Progressive Matrices~\cite{hu2021stratified}).

We address this gap by evaluating the most prominent openly available visual in-context learning models on multi-domain visual analogies, which has been identified as important factor for making progress in the field~\cite{mitchell2021abstraction}.
Specifically, we stress test VICL models on our {\color{magenta}V}isual {\color{cyan}I}n-Context {\color{orange}B}{\color{teal}E}nchmark (\benchmarkname), on out-of-domain images from medicine, underwater- and aquatic environments, on images from art and aesthetics, wildlife as well as object- and human-centric imagery in natural- and infrared imagery.
Further, we include a wide range of vision tasks, including perception and localization tasks, restoration and image enhancement tasks, generative image manipulation tasks and we also probe model's capability with simple geometric transformations.
Coupled with a detailed analysis and interpretation of the results, our contributions amount to:
\begin{itemize}
    \item We introduce the \benchmarkname~toolkit with a new evaluation protocol that includes deterministic context sampling, unified task encodings and unified post-processing to obtain metric scores.
    \item We evaluate and stress test the six most prominent VICL models on the currently broadest set of 12 tasks, 14 datasets across 6 domains in 106 one-shot task-dataset pairings.
    \item We derive key insights of the current state of visual in-context learning, including model properties and failure modes in this broad evaluation,  starting to isolate effects of design choices made with respect to model architectures, training data and training strategies.
\end{itemize}


\section{Related work}


\paragraph{Visual in-context learning}
In natural language processing, specifically in the GPT-3 model~\cite{brown2020language}, notable few-shot learning capabilities were observed, without weight updates, but by merely providing the model with few-shot examples of the task to solve in the input.
Thereafter, the first approach to adapt this setup to the vision domain is MAE-VQGAN~\cite{bar2022visual}, which designed the conditional input as an image grid, where the context input-output examples are placed at the top next to each other, while the query image is placed in the bottom left.
The model is then tasked with inpainting the masked bottom right region.
Following a similar design, but moving away from uncurated Computer Vision Figures as training data and towards a collection of curated vision datasets is the Painter model~\cite{wang2023images}.
There, the training process is designed as pixel-wise regression utilizing a Vision Transformer (ViT)~\cite{vit_visiontransformer} with a masked image modeling strategy~\cite{xie2022simmim,he2022masked}.
The IMProv model~\cite{improv} added a way to integrate textual task prompts into the grid-based in-context modeling for multi-modal in-context prediction. 
Similar to prior work, Visual Token Matching model~\cite{kim2023universal} uses a ViT, but the self-attention mechanism is re-designed to hard-wire contextual prediction on an indoor dataset.

To more closely mirror the in-context learning setting in natural language processing~\cite{brown2020language} the Large Vision Model (LVM)~\cite{bai2024sequential} is a billion parameter model trained on a large amount of vision datasets and formulates VICL as sequence prediction.
Further, an image quantization model~\cite{esser2021taming} is utilized to operate on a visual vocabulary rather than pixels -- mirroring language models.
In an effort to cope with the data intensive compute of the LVM training recipe, DeLVM~\cite{delvm} was introduced with an focus on data-efficient training, utilizing a much smaller amount of training datasets and leveraging the idea of knowledge distillation~\cite{hintondistill} to distill smaller, parameter-efficient model variants.

Diffusion models~\cite{sohl2015deep,ho2020denoising,rombach2022high} have seen some application in VICL through the approach PromptDiffusion~\cite{promptdiffusion} which is trained on the synthetic InstructPix2Pix dataset~\cite{brooks2023instructpix2pix} and uses a ControlNet~\cite{zhang2023adding} to integrate visual context.
Similar to PromptDiffusion, Analogist~\cite{gu2024analogist} utilizes diffusion models, similar to MAE-VQGAN it builds an image grid as input and re-designs the attention as Visual Token Matching~\cite{kim2023universal}.
It further automatically extracts a text prompt to support generating the correct output of the visual analogy.
SD-VICL~\cite{oorloff2025stable} is a diffusion model-based VICL approach that does not utilize text and works training-free, expanding upon ideas from image editing~\cite{alaluf2024cross} and attention re-wiring~\cite{kim2023universal,gu2024analogist}.
Lastly, VIRAL~\cite{li2026viral} addresses VICL by training a Mixture of Experts model~\cite{jiang2024mixtral} using Low-Rank Adaptation~\cite{hu2022lora} on a multi-modal image-editing model~\cite{wu2025qwen,esser2024scaling,peebles2023scalable}




\paragraph{Cross-domain benchmarks}
Benchmarks containing imaging data from diverse domains have been proposed in adjacent fields, \eg, few-shot classification~\cite{guo2020broader}, few-shot object detection~\cite{qiu2026second}, open vocabulary segmentation~\cite{blumenstiel2023mess}, or, zero-shot learning with vision-language prompts~\cite{wiedemer2025video}.
Such benchmarks provide a clearer view on the applicability of methods in new environments, which aligns with the goal to develop adaptive methods for imaging domains where data scarcity is prevalent.

For visual in-context classification~\cite{chan2022data,bratulic2025unlocking}, evaluation protocols often use few-shot protocols on individual existing datasets~\cite{lake2019omniglot}.
Zong \etal~took a step further and propose a multi-modal benchmark on in-context classification with vision language models~\cite{zong2024vl}.
While not multi-modal, the Visual analogy dataset~\cite{yiu2024kiva} matured visual analogies towards more diverse image types with focus on task diversity.
Closest to our work, Xia \etal~\cite{xia2025ideal}, propose a real-world dense prediction benchmark where fine-tuning models with few labeled examples for semantic segmentation and depth estimation is explored.
Our efforts also consider dense prediction but on a much broader set of tasks and imaging domains, and, we are set in VICL, \ie, prediction without any parameter optimization.

\section{\benchmarkname: {\color{magenta}V}isual {\color{cyan}I}n-Context {\color{orange}B}{\color{teal}E}nchmark}
\label{sec:vibe}

\begin{figure}[b]
    \centering
    \includegraphics[width=\textwidth,page=4]{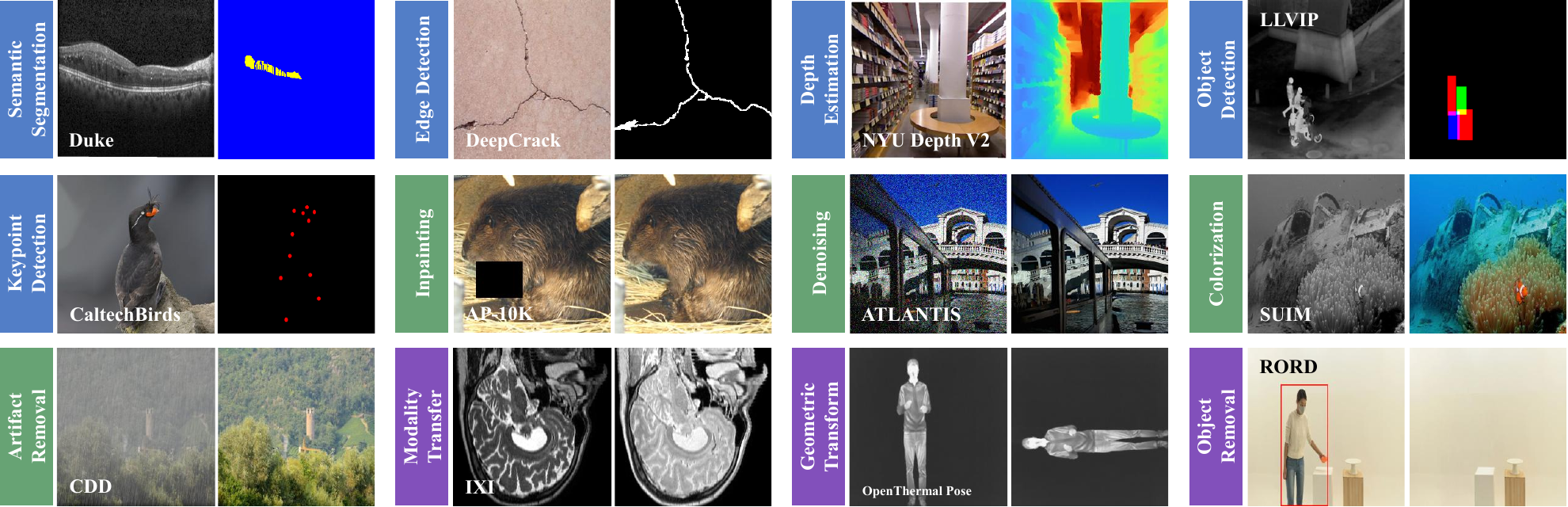}
    \caption{
    Visualization of task encodings in the \benchmarkname~toolkit.
    \pertext{Perception \& Localization}, \restext{Restoration \& Enhancement},
    and \mantext{Manipulation \& Transformation} tasks tinted in their respective colors.
    }
    \label{fig:benchmark_overview}
\end{figure}

The goal of our \benchmarkname~toolkit is to provide a unified, reproducible and extensible evaluation protocol for VICL across heterogeneous tasks and diverse imaging domains, which we introduce next.

\subsection{Preliminaries}
\label{sec:preliminaries}
\textbf{Notation} A dataset with $N$ samples can be defined as $\mathcal{D} = \{(q_i, y_i, c_i)\}_{i=1}^{N}$, where $q_i \in \mathbb{R}^{3 \times W \times H}$ is an input image, $y_i \in \mathbb{R}^{3 \times W \times H}$ its encoded target image (\eg, a segmentation map, depth map, \etc) 
and $c_i$ a deterministically selected, fixed context for each query $q_i$.
For a query image $q_i$, a VICL model $f_\theta(\cdot, \cdot)$ receives a context pair
$c_i = (c_{\mathrm{in}}, c_{\mathrm{out}}) \in \mathbb{R}^{2\times3\times W\times H}$ based on which it needs to infer the task and apply it to $q_i$ to produce $\hat{y}_i = f_\theta(q_i, c_i)$.
The goal is to predict $\hat{y}_i$ as close to $y_i$ as possible, which is assessed with task-specific metrics.
The process for selecting the context $c_i$ for a query $q_i$ is task-dependent, \ie, for tasks with semantic classes (\eg, semantic segmentation, object detection), the context is chosen via a class-aware greedy policy, to make sure the context contains as many classes in the query as possible, all remaining tasks follow fixed-seed random selection.

\textbf{Pre-processing for visual tasks}
As all tasks are formulated as image-to-image transformations to provide a unified interface for processing them with a VICL model $f_\theta$, tasks which are not naturally image-to-image tasks are encoded as RGB images.
These task-encodings are designed to preserve the structure of the original annotations to retain the nature of the task. 
Segmentation masks are rendered with a color palette where each class is assigned a fixed color, depth maps are rendered with a color mapping that transfers the continuous distance values to colors, and, for edge detection, black and white images are used with white indicating the presence of an edge. 
The encoding for keypoint detection renders red circle markers as indication for the presence of a keypoint on a black image.
For object detection the bounding boxes are rectangles filled with the value $1$ in the red channel, and, in case of overlapping boxes that indicate different instances, we use the green and blue channels to encode them, allowing the representation of up to three instances at the same location.


\subsection{Vision tasks and imaging domains in the \benchmarkname~toolkit}
Our \benchmarkname~toolkit comprises 14 datasets, 6 different domains and 12 tasks.
We group the tasks into three categories based on their prediction objective.
\pertext{Perception \& Localization} covers tasks that require spatial understanding of images, namely, \emph{semantic segmentation}, \emph{edge detection}, \emph{depth estimation}, \emph{keypoint-} and \emph{object detection}.
\restext{Restoration \& Enhancement} includes tasks in which degraded or incomplete visual content has to be recovered, \ie, \emph{inpainting}, \emph{denoising}, \emph{colorization} and \emph{artifact removal}. \mantext{Manipulation \& Transformation} contains tasks where the model has to infer and apply a systematic modification to the input, covering \emph{modality transfer}, \emph{geometric transformations} and \emph{object removal}.
These categories allow us to evaluate the adaptive visual reasoning capabilities across the different models.
\Cref{fig:benchmark_overview} shows the encoding for each task on an exemplary dataset.
Further, the benchmark covers six imaging domains to evaluate generalization beyond natural images. 

\textit{Medical \& biomedical} imaging includes scans of diseases and anatomy -- challenging out-of-domain images for VICL models.
We include Duke~\cite{srinivasan2014fully}, retinal optical coherence tomographies with diseased regions and layer annotations, and IXI~\cite{IXIDataset}, brain magnetic resonance images in two modalities.

\textit{Underwater \& aquatic} environments introduce low-light and color distortions through optical properties of water.
Datasets in this domain are ATLANTIS~\cite{erfani2022atlantis}, for water body segmentation, SUIM~\cite{islam2020semantic}, for underwater segmentation and SeaThru~\cite{akkaynak2019sea}, a collection of underwater scenes with depth maps.

\textit{Indoor \& structural} environments, represented by NYU Depth V2~\cite{silberman2012indoor}, a RGB-Depth dataset with densely annotated indoor scenes and DeepCrack~\cite{liu2019deepcrack}, for crack segmentation. This domain challenges VICL models on geometry-driven prediction and detection of fine structures.

\textit{Arts, aesthetics and photography} covers images with artistic character, leading to unusual texture and color patterns, and scenic photography images.
It includes DRAM~\cite{cohen2022semantic}, a stylized painting dataset and CDD~\cite{guo2024onerestore}, an image restoration dataset with diverse degradations.

\textit{Object- and human-centric scenes} are centered on people or distinguished objects, including low-light, infrared and thermal imagery.
We include
OpenThermalPose~\cite{kuzdeuov2024openthermalpose,kuzdeuov2025openthermalpose2},
a thermal human pose dataset, LLVIP~\cite{jia2021llvip}, a low-light visible-infrared
dataset and RORD~\cite{sagong2022rord}, an object removal dataset.

\textit{Wildlife and animal-centric} imagery focuses on animals in nature and their variation in appearance and pose.
We use CaltechBirds~\cite{wah2011caltech}, a bird species-, and AP-10K~\cite{yu2021ap}, an animal pose dataset.







\section{Evaluation}
\label{sec:evaluation}

\subsection{Visual in-context models}
Next, we introduce the VICL models we stress test on~\benchmarkname, we mainly emphasize their differences.

\textbf{MAE-VQGAN}~\cite{bar2022visual} trains a ViT backbone~\cite{vit_visiontransformer} on an uncurated dataset of computer vision figures with masked image modeling~\cite{xie2022simmim,bao2021beit,he2022masked} to predict tokens from a visual codebook~\cite{esser2021taming}.


\textbf{Painter}~\cite{wang2023images} uses a ViT backbone and masked image modeling as well, but it is trained on a highly curated set of vision tasks that are put into a grid.
Further, it directly regresses RGB values for the masked out regions, circumventing the use of discrete visual tokens.

\textbf{LVM}~\cite{bai2024sequential} also mainly uses curated image data, but from 50 computer vision datasets, which are put into sequential form.
The approach uses a discrete visual codebook to shorten these sequences and then applies next token prediction as an objective~\cite{bengio2003neural,radford2019language} for a 7B parameter LLamA model~\cite{touvron2023llama}.

\textbf{DeLVM}~\cite{delvm} follows the training recipe of LVM on a set of five datasets and adds data augmentation and knowledge distillation~\cite{hintondistill} to obtain a small 300M parameter model.
In this benchmark we stress test their larger, more capable 1B parameter model.

\textbf{PromptDiffusion}~\cite{promptdiffusion} makes use of a diffusion model~\cite{sohl2015deep}, namely a pre-trained stable diffusion model~\cite{rombach2022high} for prediction. It builds upon ControlNet~\cite{zhang2023adding} to condition the generation on the context and is trained on enriched, synthetically generated data~\cite{brooks2023instructpix2pix}. We prompt it purely with images and not with text, like all other models, as we are interested in vision-based adaptation capabilities.

\textbf{SD-VICL}~\cite{oorloff2025stable} also uses stable diffusion but it is the only training-free approach.
It expands on methods from style-transfer~\cite{alaluf2024cross} to condition the image generation on the visual analogy through reformulating self-attention at inference (as in self-attention cloning~\cite{gu2024analogist} or visual token matching~\cite{kim2023universal}).

\textbf{Copy target} is a naive baseline, that takes a target image from the context set as output.

\subsection{Quantitative comparison}
Next, we show the results for these six VICL models on \benchmarkname, which includes evaluation on $12$ tasks across $14$ datasets in a total of $106$ task-dataset pairings.
All computation is done on NVIDIA A100 GPUs in the one-shot setting introduced in~\Cref{sec:preliminaries} with one context pair defining the task.

\subsubsection{Perception and localization tasks}
\begin{figure}[t]
    \centering
    \includegraphics[width=0.9\textwidth, trim={0 0 0 51cm},clip]{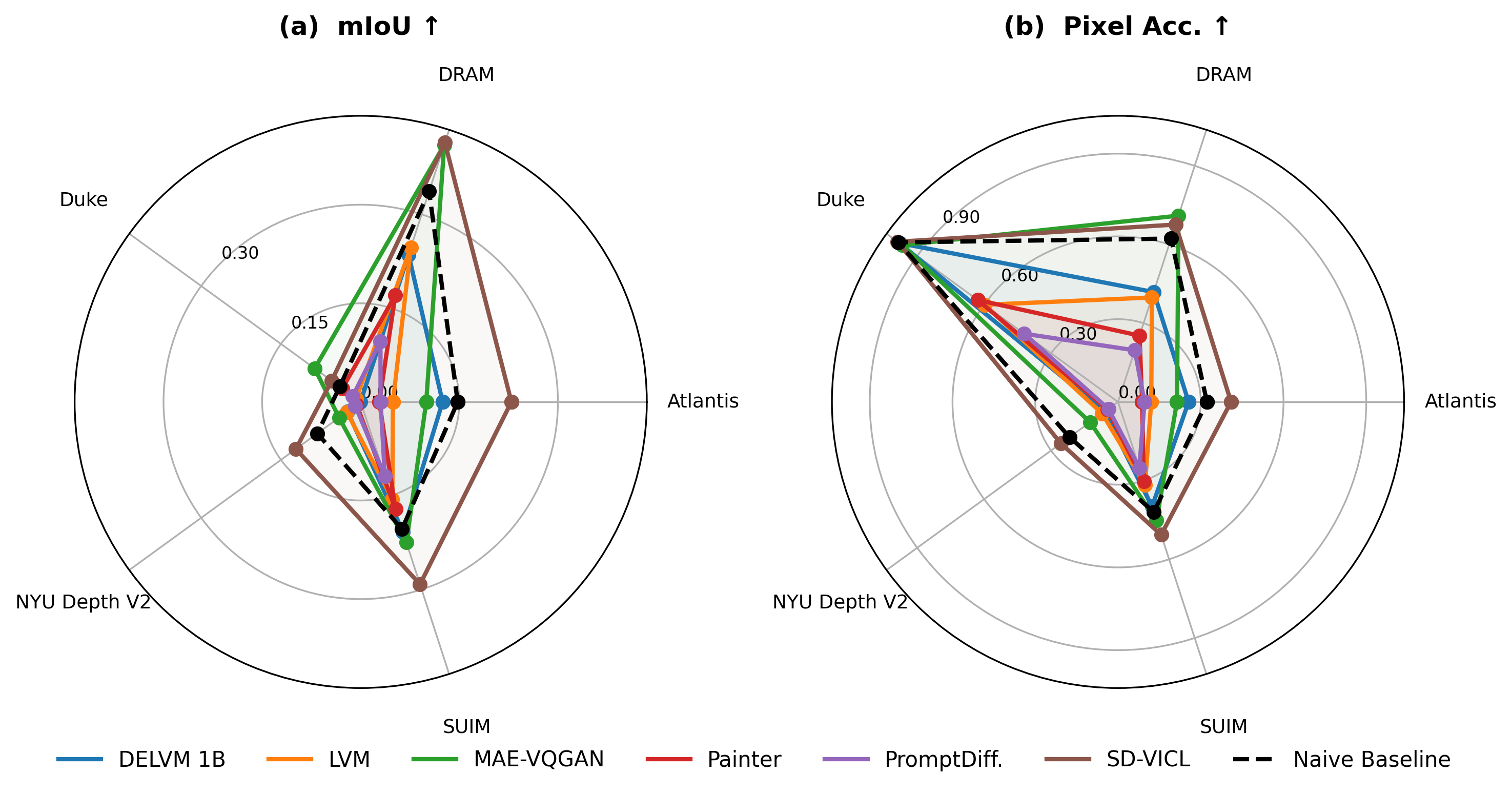}

    \begin{subfigure}[t]{0.325\textwidth}
        \centering
        \includegraphics[width=0.9\textwidth, trim={0 3.2cm 52.5cm 3.3cm},clip]{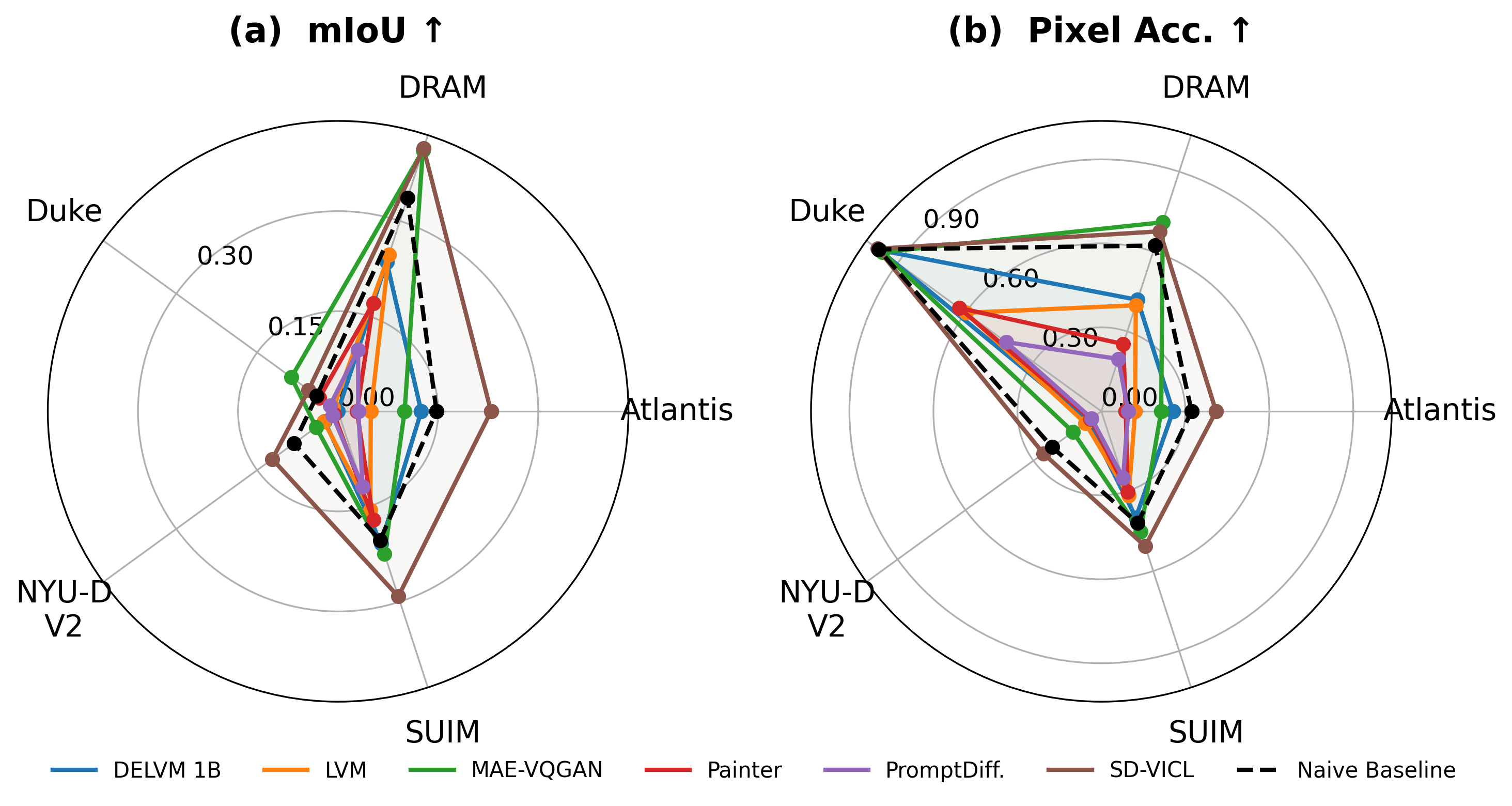}
        \caption{Semantic segmentation (IoU $\uparrow$)}
        \label{fig:perceptual_localization_spiderplot_a}
    \end{subfigure}%
    ~
    \begin{subfigure}[t]{0.325\textwidth}
        \centering
        \includegraphics[width=0.9\textwidth, trim={0 3.2cm 52.5cm 3.3cm},clip]{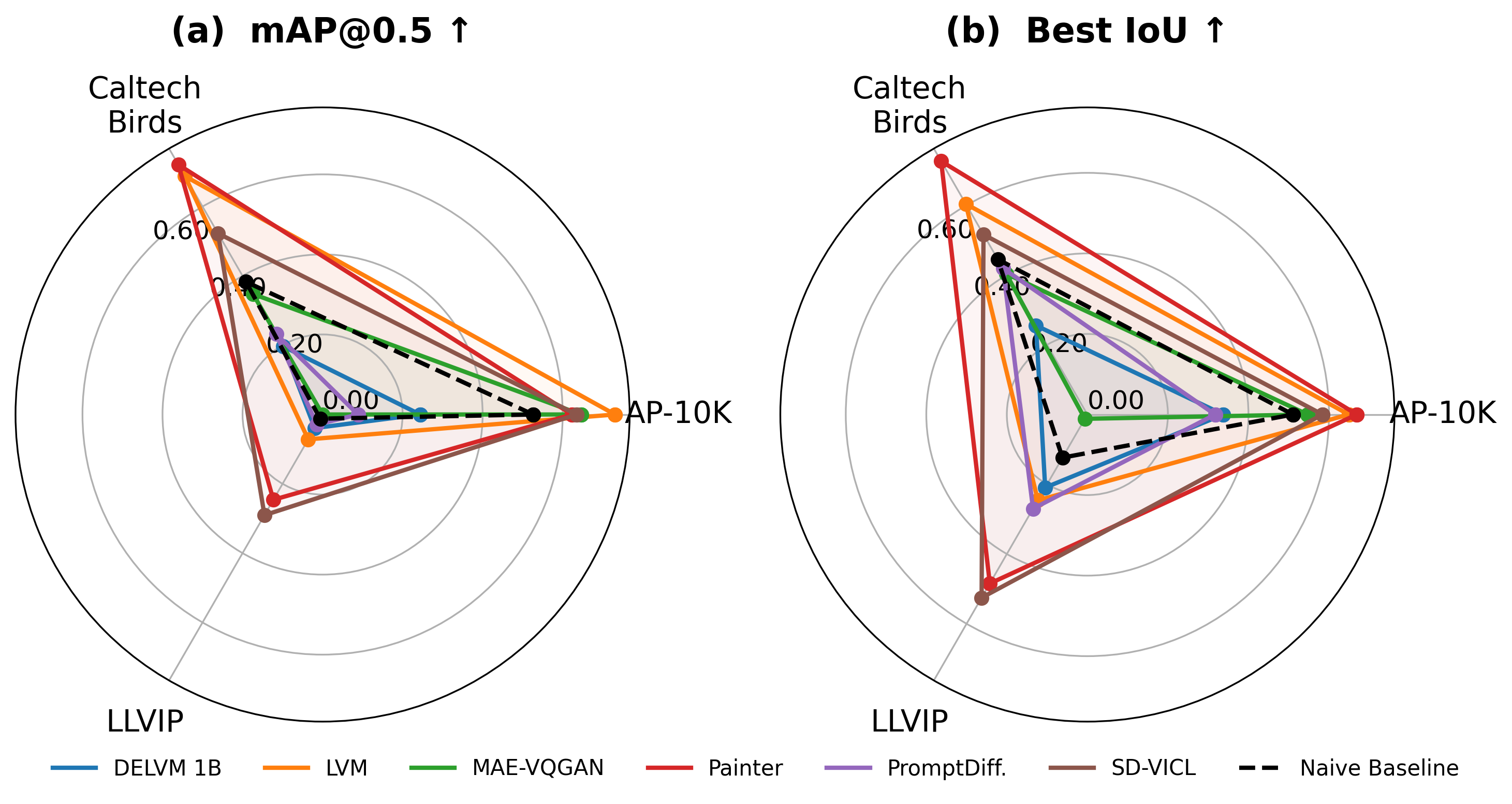}
        \caption{Object detection (mAP$@50\%~\uparrow$)}
        \label{fig:perceptual_localization_spiderplot_b}
    \end{subfigure}%
    ~
    \begin{subfigure}[t]{0.325\textwidth}
        \centering
        \includegraphics[width=0.9\textwidth, trim={0 57cm 52.5cm 3.3cm},clip]{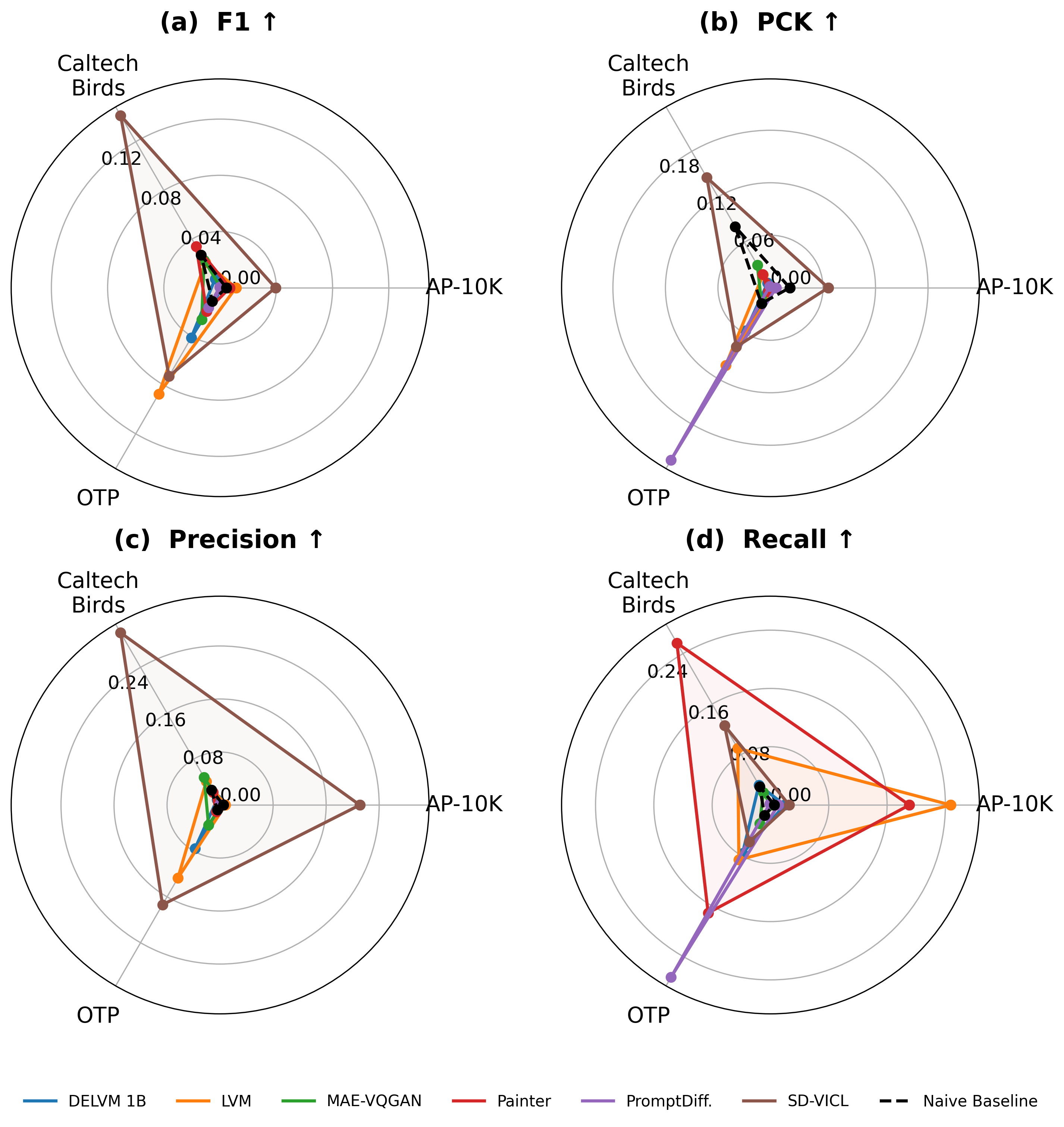}
        \caption{Keypoint detection (F1-Score $\uparrow$)}
        \label{fig:perceptual_localization_spiderplot_c}
    \end{subfigure}%
    \caption{Quantitative results for semantic segmentation in color-aware mIoU (for DUKE foreground class IoU is shown), object detection in mAP$@50\%$ and keypoint detection in F1-Score.}
    \label{fig:perceptual_localization_spiderplot}
\end{figure}

We present our results either in numerical form through tables, or in task-specific spider-plots showing results on different datasets.
We start with~\Cref{fig:perceptual_localization_spiderplot}, the dashed lines always indicate the Copy target baseline.
For~\Cref{fig:perceptual_localization_spiderplot_a}, we see the results for semantic segmentation on five datasets.
The segmentation performance as measured in color-aware IoU for VICL models is generally low, only MAE-VQGAN and SD-VICL surpass the baseline, whereas the latter consistently exceeds it.
One important factor for the low scores is that apart from these two models, no other model is able to follow the segmentation class-colors as defined in the given context examples, failing to shift away from their color-maps from pre-training.
On the detection task, in~\Cref{fig:perceptual_localization_spiderplot_b}, three models are able to solve the task consistently better than the baseline: Painter, LVM and SD-VICL.
The lowest mAP scores are achieved for the LLVIP dataset, which requires the detection of multiple instances in images, making the task more challenging, while the other two datasets always contain one instance to localize.
Keypoint detection is not well addressed by any model with low F1-Scores throughout. 

Concluding the results of the perception and localization tasks are edge detection and depth estimation in~\Cref{tab:edge_table} and~\Cref{tab:depth_table}, respectively.
For edge detection we see a similar pattern as before, where SD-VICL leads, followed by MAE-VQGAN.
Interestingly, this is one of the few tasks, where Prompt Diffusion works quite well, which can be attributed to its set of pre-training tasks including edge detection.
The results on depth estimation indicate, that none of the models is able to coherently solve the task with RMSE values close or higher than Copy target, indicating that dense regression is a hard class of tasks in VICL.
Mostly, models tend to produce segmentation maps or depth maps with different depth color coding as defined in the context.
MAE-VQGAN and SD-VICL match the colors of the depth maps from the context, but are not able to correctly predict the depth values in the scenes.

{\setlength{\tabcolsep}{2pt}

\begin{table}[b]
\centering
\begin{minipage}{0.3\textwidth}

\centering
\caption{Edge detection results on DUKE and DeepCrack.}
\label{tab:edge_table}
\scriptsize
\begin{tabular}{lccc|ccc}
\toprule
Method            & \multicolumn{3}{c}{DeepCrack} & \multicolumn{3}{c}{DUKE} \\
\cmidrule(lr){2-4} \cmidrule(lr){5-7}
                  & Prec. & Rec. & F1 & Prec. & Rec. & F1 \\
\midrule
Copy target       & 0.05 & 0.05 & 0.04 & 0.05 & 0.06 & 0.06 \\
MAE-VQGAN         & 0.35 & 0.29 & 0.29 & 0.01 & 0.03 & 0.02 \\
DeLVM             & 0.03 & 0.01 & 0.00 & 0.00 & 0.00 & 0.00 \\
LVM               & 0.16 & 0.13 & 0.08 & 0.00 & 0.14 & 0.01 \\
Painter           & 0.02 & 0.04 & 0.02 & 0.02 & \textbf{0.73} & 0.03 \\
PromptDiffusion   & 0.20 & 0.29 & 0.20 & 0.01 & 0.32 & 0.01 \\
SD-VICL           & \textbf{0.72} & \textbf{0.69} & \textbf{0.62} & \textbf{0.55} & 0.54 & \textbf{0.54} \\
\bottomrule
\end{tabular}
\end{minipage}
\hfill
\begin{minipage}{0.19\textwidth}
\centering
\caption{Depth estimation (RMSE) on NYU and Seathru.}
\label{tab:depth_table}
\scriptsize
\begin{tabular}{lc|cc}
\toprule
Method            & NYU-D v2 & Sea-thru \\
\midrule
Copy target       & 1.60 & \textbf{1.64} \\
MAE-VQGAN         & 1.75 & 2.16 \\
DeLVM             & 3.10 & 2.87 \\
LVM               & 2.73 & 4.34 \\
Painter           & 2.70 & 3.58 \\
PromptDiffusion   & 2.76 & 2.16 \\
SD-VICL           & \textbf{1.48} & 2.28 \\
\bottomrule
\end{tabular}
\end{minipage}
\hfill
\begin{minipage}{0.3\textwidth}
\centering
\caption{Object removal results on RORD-mini in PSNR, SSIM and LPIPS.}
\label{tab:object_removal_table}
\scriptsize
\begin{tabular}{lccc}
\toprule
Method            & PSNR & SSIM & LPIPS \\
\midrule
Copy target       & 9.58  & 0.22 & 0.77 \\
MAE-VQGAN         & 14.23 & 0.38 & 0.70 \\
DeLVM             & 14.62 & 0.41 & 0.60 \\
LVM               & \textbf{15.84} & 0.42 & \textbf{0.57} \\
Painter           & 13.51 & \textbf{0.49} & 0.58 \\
PromptDiffusion   & 10.68 & 0.34 & 0.75 \\
SD-VICL           & 12.19 & 0.36 & 0.64 \\
\bottomrule
\end{tabular}
\end{minipage}
\end{table}
}

\subsubsection{Image restoration and enhancement tasks}
\begin{figure}[t!]
    \centering
    \includegraphics[width=0.9\textwidth, trim={0 0 0 51cm},clip]{images/quantitative/spiderplots/segmentation_datasets.png}
    
    \begin{subfigure}[t]{0.32\textwidth}
        \centering
        \includegraphics[width=0.89\textwidth, trim={0 3.2cm 105cm 3.3cm},clip]{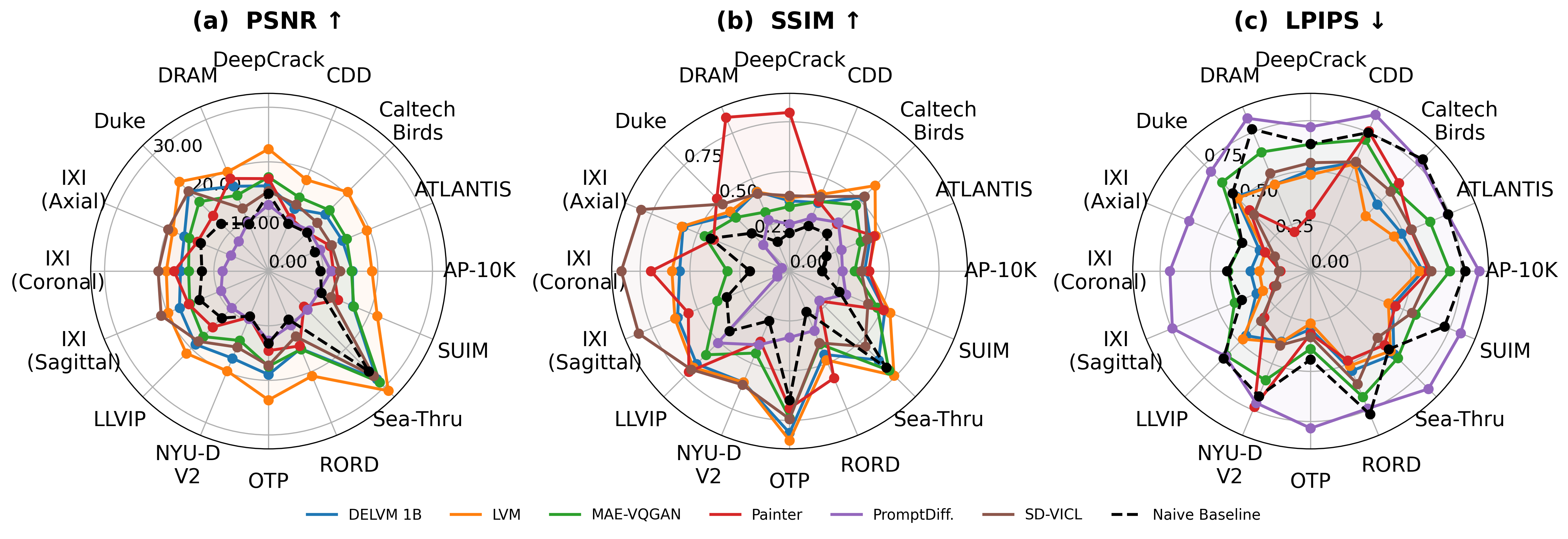}
        \caption{Inpainting (PSNR $\uparrow$)}
        \label{fig:image_restoration_enhancement_spiderplot_a}
    \end{subfigure}
    ~
    \begin{subfigure}[t]{0.32\textwidth}
        \centering
        \includegraphics[width=0.89\textwidth, trim={0 3.2cm 105cm 3.3cm},clip]{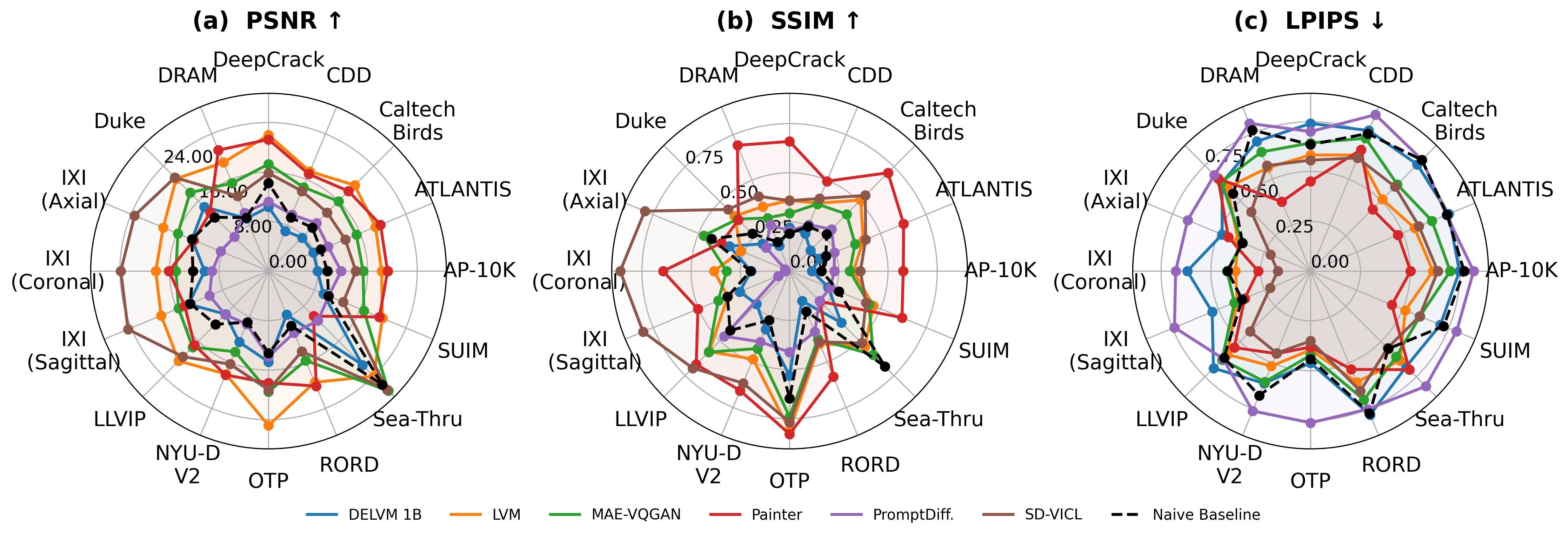}
        \caption{Denoising (PSNR $\uparrow$)}
        \label{fig:image_restoration_enhancement_spiderplot_b}
    \end{subfigure}
    ~
    \begin{subfigure}[t]{0.32\textwidth}
        \centering
        \includegraphics[width=0.89\textwidth, trim={0 3.2cm 105cm 3.3cm},clip]{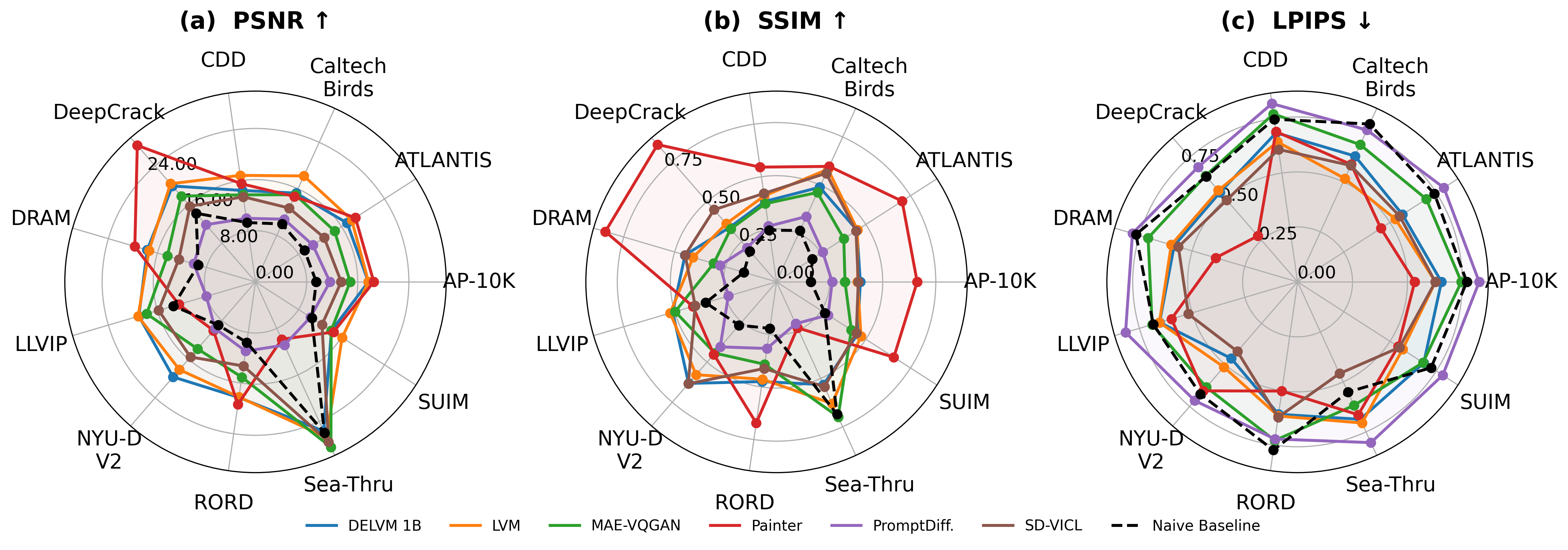}
        \caption{Colorization (PSNR $\uparrow$)}
        \label{fig:image_restoration_enhancement_spiderplot_c}
    \end{subfigure}

    \begin{subfigure}[t]{0.33\textwidth}
        \centering
        \includegraphics[width=0.89\textwidth, trim={105cm 3.2cm 0 3.3cm},clip]{images/quantitative/spiderplots/colorization_datasets.png}
        \caption{Colorization (LPIPS $\downarrow$)}
        \label{fig:image_restoration_enhancement_spiderplot_d}
    \end{subfigure}
    ~ 
    \begin{subfigure}[t]{0.33\textwidth}
        \centering
        \includegraphics[width=0.89\textwidth, trim={0 3.2cm 105cm 3.3cm},clip]{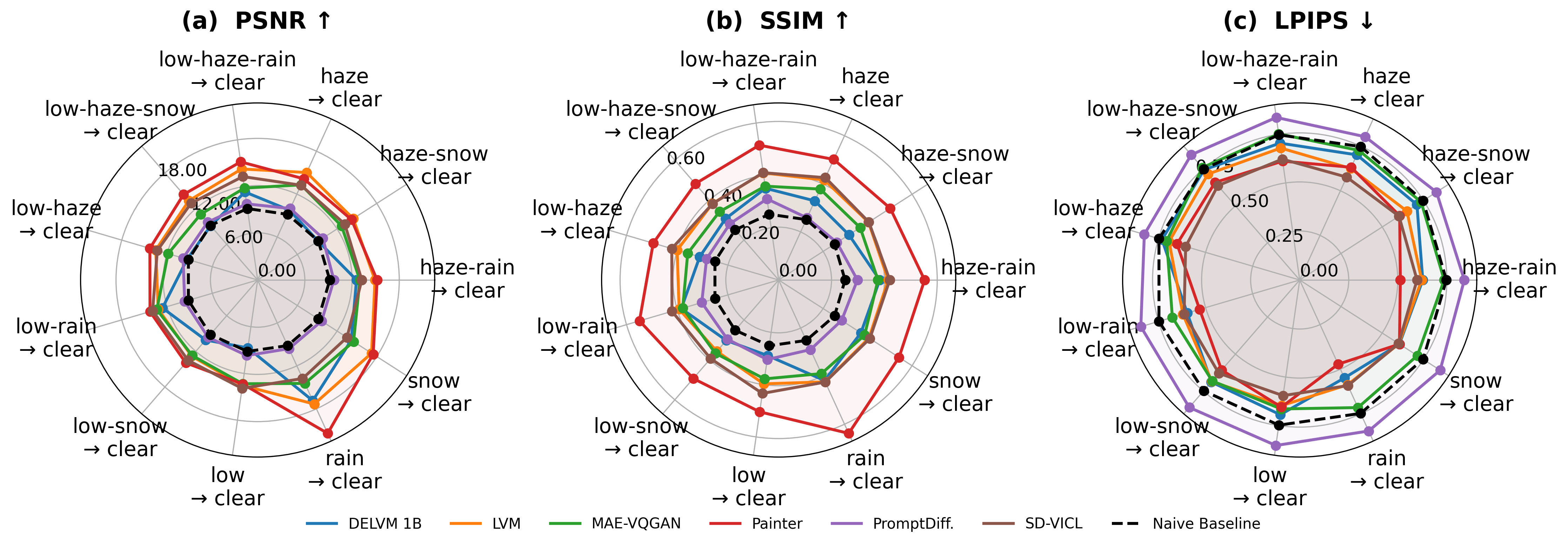}
        \caption{Artifact removal (PSNR $\uparrow$)}
        \label{fig:image_restoration_enhancement_spiderplot_e}
    \end{subfigure}
    
    \caption{Quantitative results for image restoration tasks. We report PSNR for inpainting, denoising and artifact removal, for colorization PSNR and LPIPS. Artifact removal is done on the CDD dataset.}
    \label{fig:image_restoration_enhancement_spiderplot}
\end{figure}

In~\Cref{fig:image_restoration_enhancement_spiderplot}, we consider tasks where a clean image needs to be restored from a corrupted image.
Inpainting (\Cref{fig:image_restoration_enhancement_spiderplot_a}) is done best by LVM, with the exception of the IXI dataset, where SD-VICL yields the highest PSNR.
This is unsurprising, as LVM is trained on next-token prediction, which naturally aligns with predicting masked out regions.
Similarly, MAE-VQGAN was pre-trained on masked image modeling but produces lower results, which can be attributed to the dated codebook used by the model, limiting the generated level of detail.
Interestingly, Painter, also trained with a masking objective, is not able to coherently solve the task, just passing along the masked query.

The front runners for image denoising (\Cref{fig:image_restoration_enhancement_spiderplot_b}) are Painter, SD-VICL and LVM.
SD-VICL is the best model on the medical domain, where it even removes the original image noise in the vitreous body on the retina scans of DUKE.
This may be due to the use of a quantized codebook~\cite{rombach2022high}, which does not capture noise patterns well~\cite{reiss2025visual}.
LVM also uses such a codebook while Painter does not, which aligns with the PSNR scores.
While Painter performs well overall for denoising, on which it was pretrained, its results deteriorate in the medical domain and on Seathru .
This is due to the low brightness in images of these datasets, where Painter incorrectly infers the low-light enhancement task from the context, a task from its pre-training, leading to bright rather than denoised predictions.

This pattern can also be seen on colorization (\Cref{fig:image_restoration_enhancement_spiderplot_c}).
There, LVM and DeLVM perform most robust in terms of PSNR.
Yet, DeLVM and Painter merely predict the grayscale query image as `colorized image', indicating, that in conjunction to PSNR, a perceptual metric like LPIPS (\Cref{fig:image_restoration_enhancement_spiderplot_d}) needs to be considered for this task.
Still, passing along the query image leads to perceptually sound images, which lead to low LPIPS scores in the case of Painter as opposed to, \eg, LVM and SD-VICL, where the codebooks limit the generated level of detail, but which meaningfully colorize images.

Lastly, in~\Cref{fig:image_restoration_enhancement_spiderplot_e}, we evaluate the models on artifact removal, \ie, transferring snowy, rainy, hazy and low-light, to clear images.
Painter, which was pre-trained on image de-raining, performs well on rain $\to$ clear, while SD-VICL and LVM mostly match it's performance, even with their limitation to not reconstruct fine details.
Prompt Diffusion performs worst producing no coherent image content.


\subsubsection{Image manipulation and transformation tasks}
As the last task group, we consider image manipulations and transformations.
First, we start with the task of object removal, in~\Cref{tab:object_removal_table}, where models have to remove objects that are marked with red bounding boxes.
While none of the models shine on the task, the only models that start to address it are LVM, which sometimes removes highly dynamic objects, like a moving person, which might be due to its pretraining on image sequences from video, and, SD-VICL, which consistently removes the red boxes and the area within it, but tends to produce artifact riddled reconstructions of the scene.

\begin{figure}
    \centering
    \includegraphics[width=0.9\textwidth, trim={0 0 0 51cm},clip]{images/quantitative/spiderplots/segmentation_datasets.png}
    
    \begin{subfigure}[t]{0.32\textwidth}
        \centering
        \includegraphics[width=0.9\textwidth, trim={0 3.2cm 105cm 3.3cm},clip]{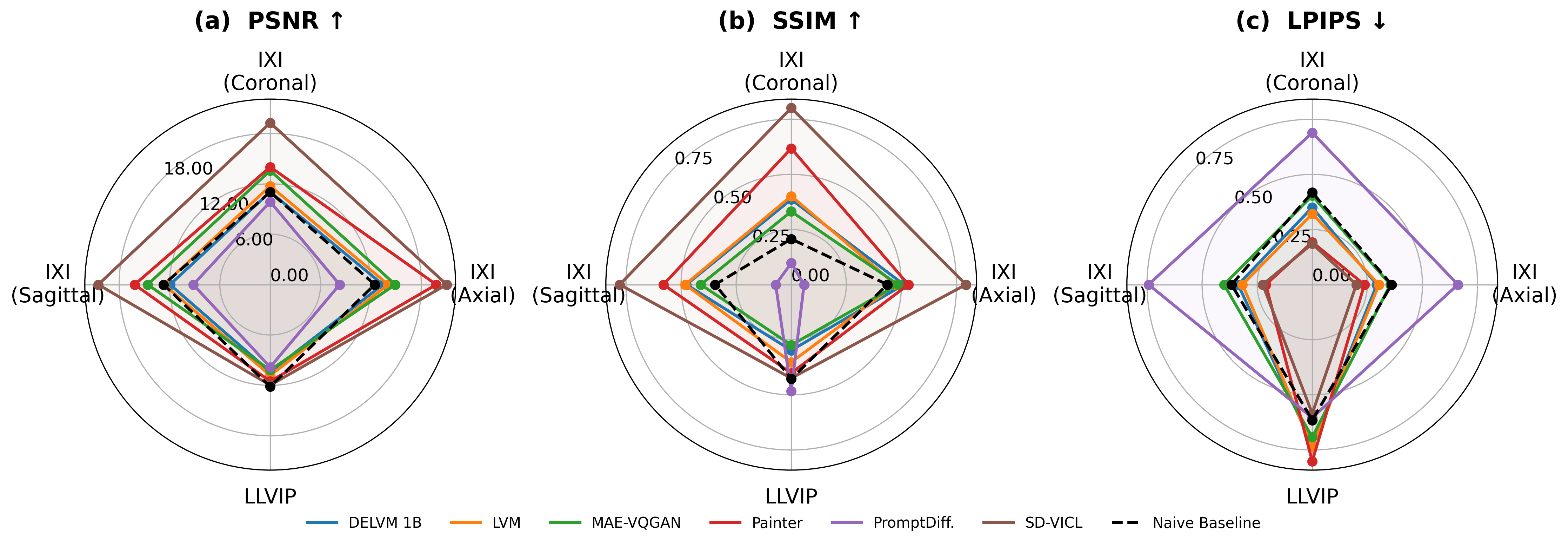}
        \caption{Modality transfer (PSNR $\uparrow$)}
    \label{fig:manipulation_transformation_spiderplot_a}
    \end{subfigure}
    ~
    \begin{subfigure}[t]{0.32\textwidth}
        \centering
        \includegraphics[width=0.9\textwidth, trim={52.5cm 3.2cm 52.5cm 3.3cm},clip]{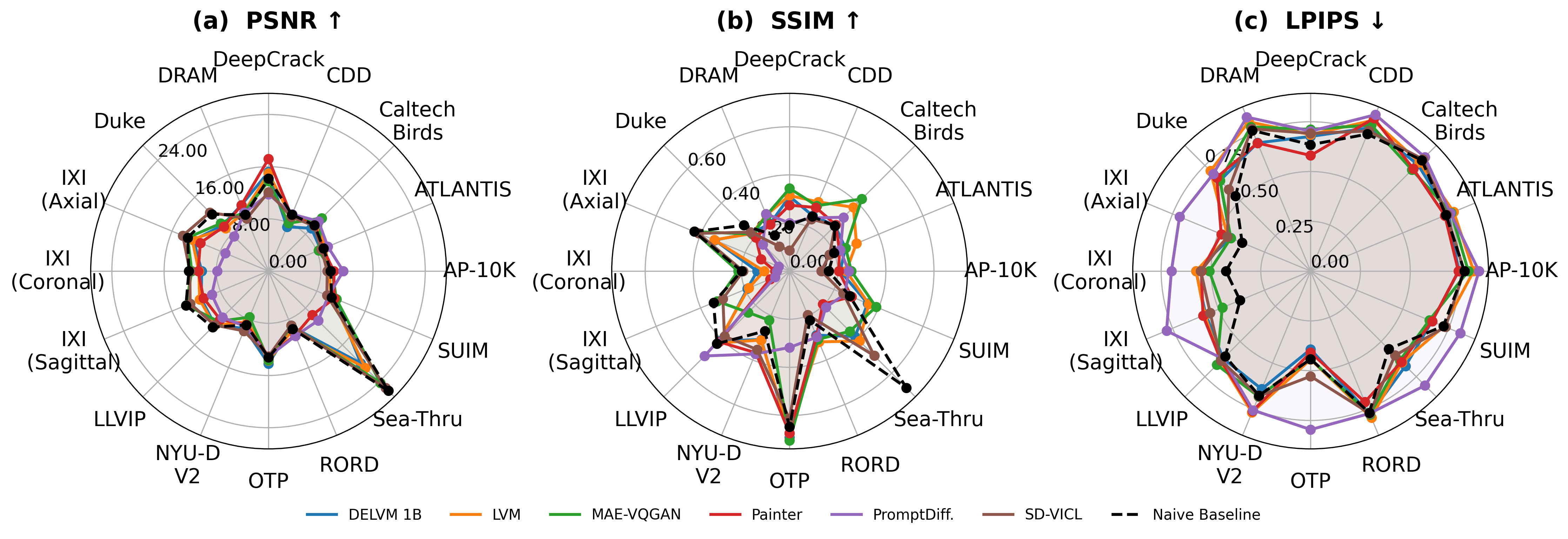}
        \caption{Rotation (SSIM $\uparrow$)}
    \label{fig:manipulation_transformation_spiderplot_b}
    \end{subfigure}
    ~
    \begin{subfigure}[t]{0.32\textwidth}
        \centering
        \includegraphics[width=0.9\textwidth, trim={52.5cm 3.2cm 52.5cm 3.3cm},clip]{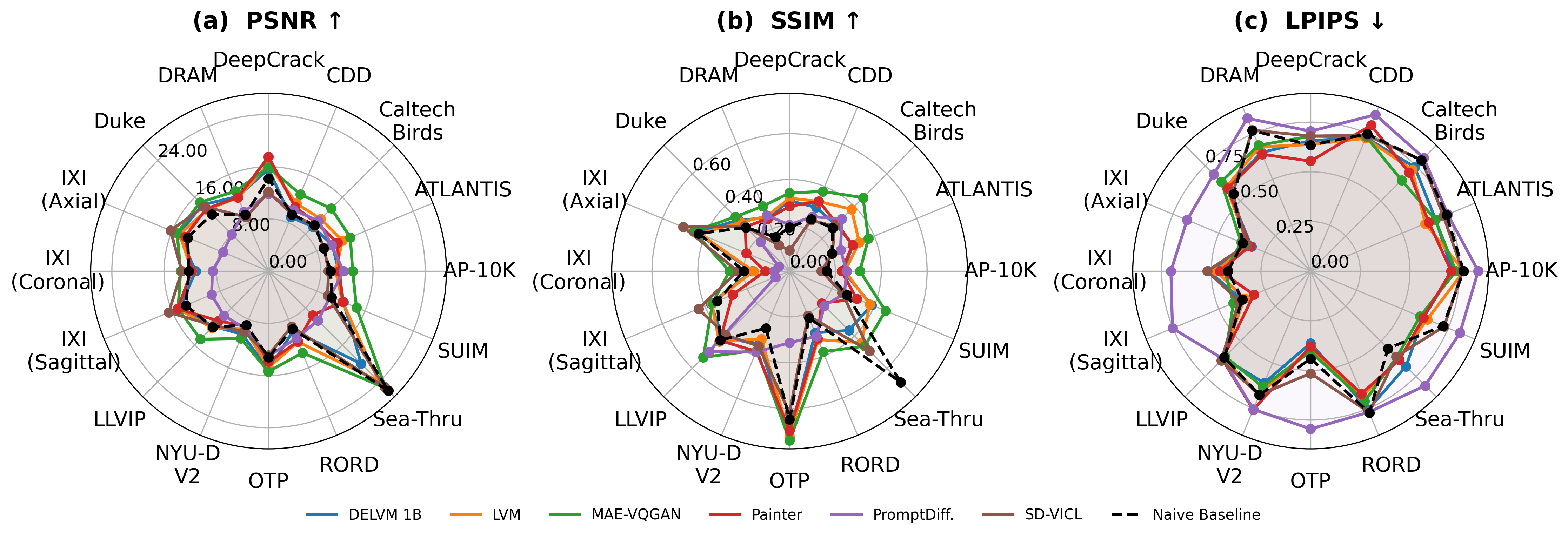}
        \caption{Flip (SSIM $\uparrow$)}
    \label{fig:manipulation_transformation_spiderplot_c}
    \end{subfigure}    
    
    \caption{Quantitative results for modality transfer in PSNR, and rotation and flipping in SSIM.}
    \label{fig:manipulation_transformation_spiderplot}
\end{figure}

In~\Cref{fig:manipulation_transformation_spiderplot_a}, we show the performance of VICL models on transferring medical imaging modalities from T2 weighted to PD MRI images on the IXI dataset, and, RGB images to infrared on LLVIP.
Throughout the MRI transfer experiments, SD-VICL works best with a substantial margin in PSNR.
Due to the homogeneous target domain, the AdaIN~\cite{huang2017arbitrary} mechanism used in SD-VICL is particularly helpful.
On LLVIP, all models cluster around the baseline, showing no meaningful task adaptation.

In our final task-specific evaluation in~\Cref{fig:manipulation_transformation_spiderplot_b} and~\Cref{fig:manipulation_transformation_spiderplot_c}, we test, whether models can cope with simple geometric transformations.
Specifically, we look into whether VICL models can solve the task of rotating an image by 90$^\circ$ or flipping it horizontally.
We see, that no model is able to accomplish this, as performance is mostly on par or worse than the baseline in SSIM, which is consistent with insights from Vision-Language models~\cite{anis2025limitations}.
There is just one notable exception, which is MAE-VQGAN, which exceeds the baseline consistently on the flip task, falling short only once.





\begin{wraptable}{R}{8cm}
\setlength{\tabcolsep}{4.5pt}
    \centering
    \scriptsize
    \begin{tabular}{rp{0.003\textwidth}!{\color{white}|}p{0.003\textwidth}!{\color{white}|}p{0.003\textwidth}!{\color{white}|}p{0.003\textwidth}!{\color{white}|}p{0.003\textwidth}!{\color{white}|}p{0.003\textwidth}!{\color{white}|}p{0.003\textwidth}!{\color{white}|}p{0.003\textwidth}!{\color{white}|}p{0.003\textwidth}!{\color{white}|}p{0.003\textwidth}!{\color{white}|}p{0.003\textwidth}!{\color{white}|}p{0.003\textwidth}!{\color{white}|}p{0.003\textwidth}!{\color{white}|||}p{0.003\textwidth}}
         & \rotatebox{75}{\tiny Sem. seg.} & \rotatebox{75}{\tiny Edge det.} & \rotatebox{75}{\tiny Depth est.} & \rotatebox{75}{\tiny Obj. det.} & \rotatebox{75}{\tiny Keyp. det.} & \rotatebox{75}{\tiny Inpaint} & \rotatebox{75}{\tiny Denoise} & \rotatebox{75}{\tiny Colorize} & \rotatebox{75}{\tiny Artif. rem.} & \rotatebox{75}{\tiny Mod. trans.} & \rotatebox{75}{\tiny Rot.} & \rotatebox{75}{\tiny Flip} & \rotatebox{75}{\tiny Obj. rem.} & \rotatebox{75}{\tiny Total} \\
        SD-VICL & \1 & \1 & \2 & \2 & \1 & \2 & \1 & \3 & \2 & \1 & \4 & \6 & \5 & \1 \\
        & & & & & & & & & & & & & & \\[-2.2mm]
        LVM & \5 & \5 & \7 & \3 & \2 & \1 & \3 & \2 & \3 & \3 & \4 & \2 & \1 & \2 \\
        & & & & & & & & & & & & & & \\[-2.2mm]
        Painter & \6 & \6 & \6 & \1 & \4 & \3 & \2 & \1 & \1 & \2 & \6 & \4 & \2 & \3 \\
        & & & & & & & & & & & & & & \\[-2.2mm]
        MAE-VQGAN & \2 & \2 & \3 & \5 & \3 & \5 & \4 & \5 & \4 & \4 & \2 & \1 & \4 & \4 \\
        & & & & & & & & & & & & & & \\[-2.2mm]
        DeLVM & \4 & \7 & \5 & \6 & \6 & \3 & \6 & \3 & \5 & \4 & \3 & \3 & \3 & \5 \\
        & & & & & & & & & & & & & & \\[-2.2mm]
        Copy target & \3 & \4 & \1 & \4 & \5 & \6 & \5 & \6 & \7 & \6 & \1 & \5 & \7 & \6 \\
        & & & & & & & & & & & & & & \\[-2.2mm]
        Prompt Diffusion & \7 & \3 & \4 & \7 & \7 & \7 & \7 & \7 & \6 & \7 & \7 & \7 & \6 & \7 \\
        & & & & & & & & & & & & & & \\[-2.2mm]
    \end{tabular}
    \caption{Ranking of the six VICL models and baseline across all tasks and datasets, as well as average total rank.}
    \label{tab:vicl_ranking}
\end{wraptable}

\subsubsection{Ranking visual in-context learners}
To get a clearer picture of which model performs best and adapts to the most datasets and tasks we cover in~\benchmarkname, we compute a ranking in~\Cref{tab:vicl_ranking}, which is based on all datasets in each task.
Then we take the average of the ranks for a final ranking (rightmost column).
We see a clear winning model, which is the training-free SD-VICL approach, ahead of LVM, Painter, MAE-VQGAN and DeLVM.
Prompt Diffusion ranks below the Copy baseline, indicating no real capabilities for adaptive task solving.

\subsection{Qualitative comparison}

\begin{figure}
    \tiny
    $\phantom{\rotatebox{90}{\tiny T}}$
    \hfill
    \begin{tabular}{p{0.04\textwidth}p{0.04\textwidth}p{0.04\textwidth}p{0.04\textwidth}p{0.04\textwidth}p{0.04\textwidth}p{0.04\textwidth}p{0.04\textwidth}p{0.04\textwidth}p{0.04\textwidth}p{0.04\textwidth}p{0.04\textwidth}p{0.04\textwidth}p{0.04\textwidth}}
         SUIM & DeepC. & NYUDv2 & SeaThru & LLVIP & OTP & AP10K & DUKE & DRAM & CDD & IXI C. & Atlantis & CuB &RORD 
    \end{tabular}

    \centering

    \rotatebox{90}{\tiny $\phantom{...}$Target$\phantom{....}$SD-VICL$\phantom{...}$Painter$\phantom{...}$MAE-VQ.$\phantom{..}$DeLVM$\phantom{.....}$LVM$\phantom{....}$PromptD.$\phantom{....}$Query}
    \hfill
    \includegraphics[width=0.9825\textwidth, trim={5.5cm 0 0 2cm}, clip]{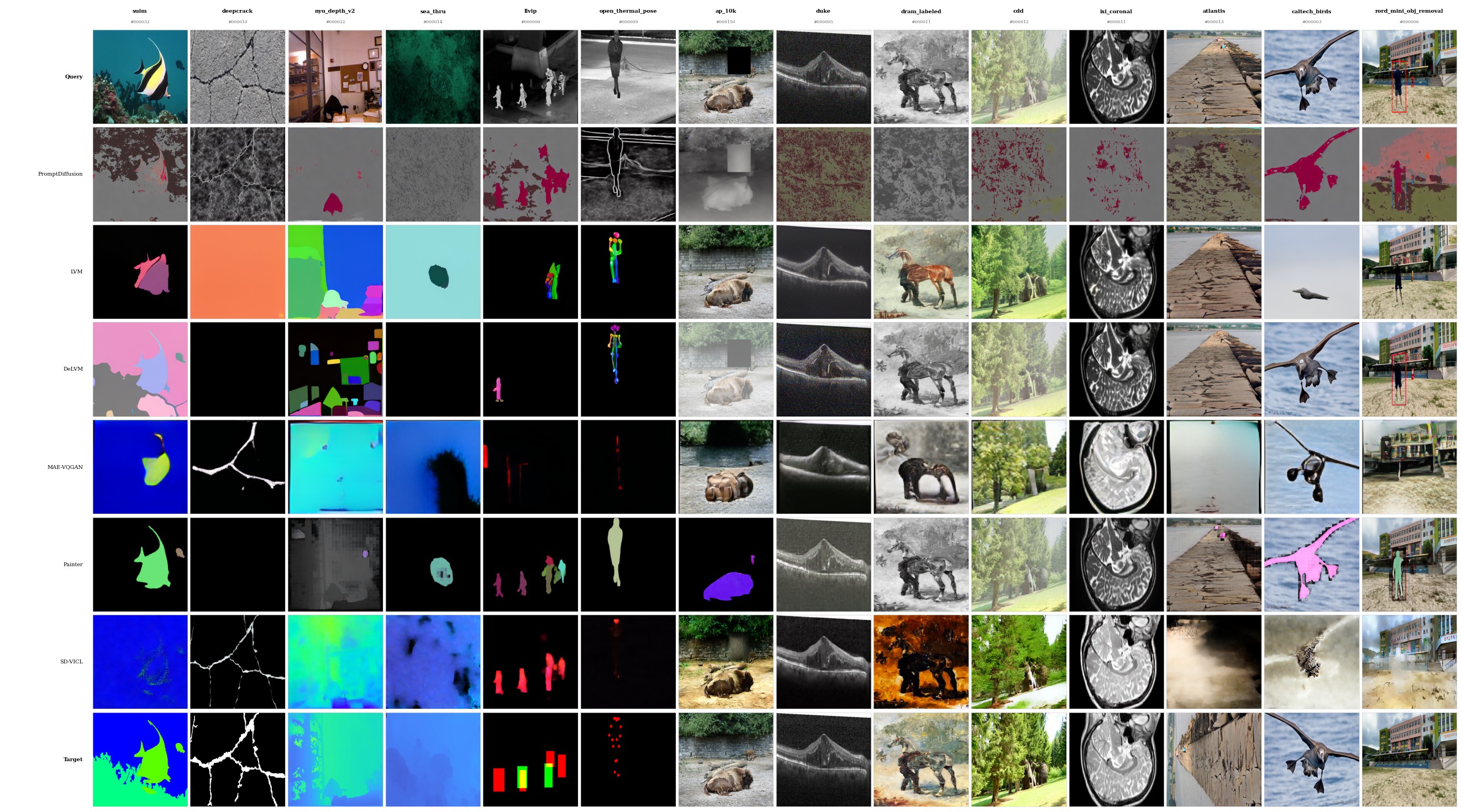}

    \tiny
    $\phantom{\rotatebox{90}{\tiny T}}$
    \hfill
    \begin{tabular}{p{0.04\textwidth}p{0.04\textwidth}p{0.04\textwidth}p{0.04\textwidth}p{0.04\textwidth}p{0.04\textwidth}p{0.04\textwidth}p{0.04\textwidth}p{0.04\textwidth}p{0.04\textwidth}p{0.04\textwidth}p{0.04\textwidth}p{0.04\textwidth}p{0.04\textwidth}}
         S.Seg. & Edge & Depth & Depth & Obj.det. & Kp.det. & Inpaint & Denoise & Colorize & Art.rem. & M.trans. & Rotate & Flip & Obj.rem.
    \end{tabular}
    
    \caption{Qualitative results for subset of 14 task-dataset pairs in~\benchmarkname, showing skills and failures of models. Top row shows the dataset- and bottom row task abbreviations (best viewed zoomed in).}
    \label{fig:qualitative}
\end{figure}

In~\Cref{fig:qualitative}, we show the predictions of the six models, which highlights failure cases, such as predicting task encodings from pretraining, handing through query images and failing to adapt to new tasks as well as notable skills, \eg, SD-VICL's capabilities on edge detection and modality transfer, MAE-VQGAN on flipping an image or successful object removal by both (rightmost column).

\section{Discussion and conclusion}
\label{sec:discussion}

\noindent\textbf{Most models are multi-task models, not in-context learners.}
With our experiments, it becomes evident, that most VICL models are heavily limited by their pre-training tasks.
Their predictions are biased towards tasks seen in training and prompting them with new context sets still leads to outputs aligned with seen tasks.
This shows that rather than being adaptive models, they seem to learn a \emph{switch} based on the context that merely steers the model to an appropriate task from pre-training~\cite{yin2020meta}.
The most notable exceptions are SD-VICL and MAE-VQGAN, which draw contextual predictions.

{\setlength{\tabcolsep}{2pt}
\begin{table}[b!]
    \centering
    \scriptsize
    \begin{tabular}{cccccccccc}
        $C_{in}$ & $C_{out}$ & $Q$ & & & SD-VICL Pred. & LVM Pred. & &  &Ground-truth \\
        \includegraphics[width=0.125\textwidth]{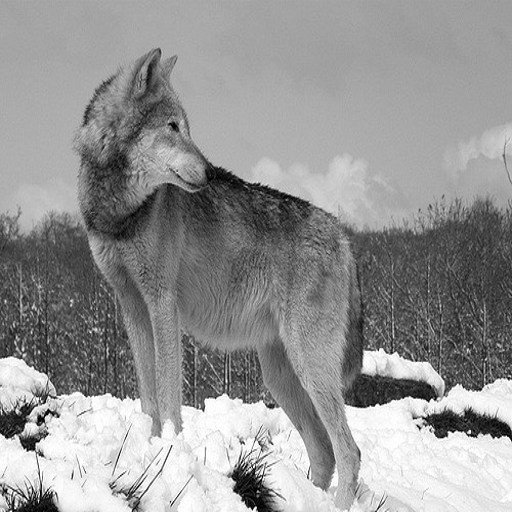}&\includegraphics[width=0.125\textwidth]{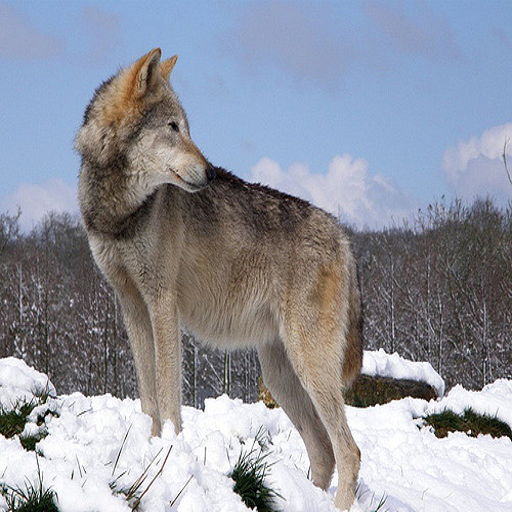}&\includegraphics[width=0.125\textwidth]{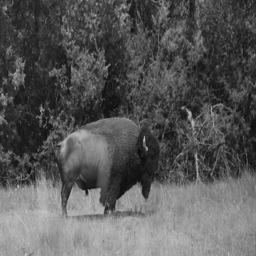}& & & \includegraphics[width=0.125\textwidth]{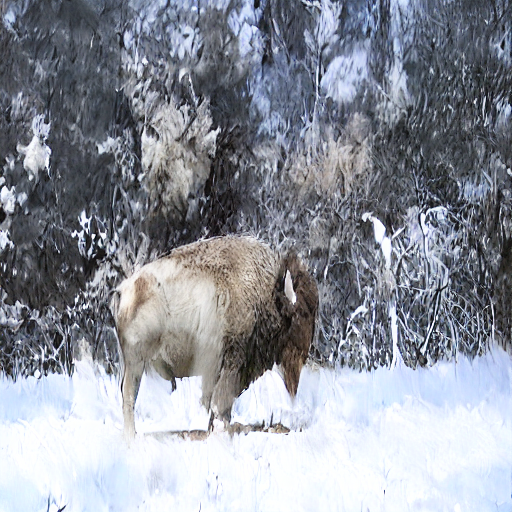}&\includegraphics[width=0.125\textwidth]{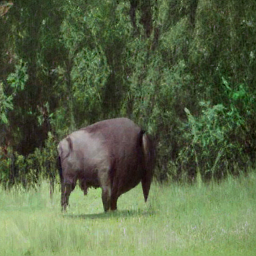} & & &\includegraphics[width=0.125\textwidth]{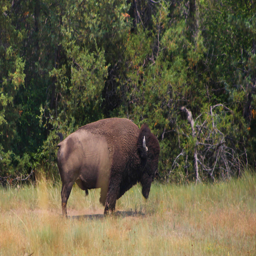} \\ [-3.2mm]
        &&&&&{\tiny LPIPS $0.669$} &{\tiny LPIPS $\textbf{0.638}$}&&&

    \end{tabular}
    \caption{Example of the ill-posedness in VICL showing the need for context sensitive metrics.}
    \label{tab:colorize}
\end{table}}

\noindent\textbf{Correspondence is key.}
Contextual prediction, which is fundamental for VICL, requires models to find relevant correspondences between the query image and the context.
This is possible by hard-wiring such correspondences into the prediction process, as, \eg, SD-VICL does with its altered self-attention mechanism, or it requires that models have an incentive to learn them.
MAE-VQGAN is an example where correspondences need to be learned due to the training data and objective.
It is trained on \emph{unstructured} data from figures of computer vision papers.
To solve masked image modeling on this data, MAE-VQGAN needs to learn correspondences within these diverse figures.
Models like LVM, DeLVM, Painter and Prompt Diffusion have no incentive to learn correspondences, as the context is always provided in a \emph{structured} way in training, \ie, through regular grids, visual sentences or ControlNet, leading them to take the shortcut of learning a \emph{switch} based on the image statistics in the task-specific contexts of the pre-training tasks, degenerating them to multi-task models.
Research on Visual in-Context Classification~\cite{chan2022data,bratulic2025unlocking} also points to the central role of skewed data distributions in the contexts in training as important factor to encourage learning correspondences.

\noindent\textbf{Inherently contextual vision tasks.}
To a computer vision researcher, it is clear what is meant by a task like \emph{image colorization}: sensibly fill in a grayscale image with colors that are plausible. Yet, this notion is vastly different from the contextual colorization task in VICL, where the context defines the task to carry out on a query image.
Take a context of a wolf in a snowy forest, and a grayscale image of a bison in a forest as query image, as in~\Cref{tab:colorize}.
The ground-truth and metrics suggest that LVM's prediction is more accurate, although, based on the context, one can sensibly make the case that SD-VICL's prediction is far more fitting.
This exemplifies the vastly different view on what a task is in the VICL setting, \ie, inherently contextual, and profoundly ill-posed.
To address this limitation, in evaluation context-sensitive metrics or ground-truths that vary with different contexts could be considered in the future.
Altering reference-based perceptual metrics~\cite{fu2023dreamsim}, \eg, through using context sensitive image representations~\cite{born2026context}, or designing human evaluation protocols, could be promising directions for better measuring adaptive capabilities of VICL models.

\begin{ack}
This work was supported by funding from the pilot program Core-Informatics of the Helmholtz Association (HGF) and by the joint research school “HIDSS4Health – Helmholtz Information and Data Science School for Health. The authors gratefully acknowledge the computing time provided on the high-performance computer HoreKa by the National High-Performance Computing Center at KIT (NHR@KIT). This center is jointly supported by the Federal Ministry of Education and Research and the Ministry of Science, Research and the Arts of Baden-Württemberg, as part of the National High-Performance Computing (NHR) joint funding program (\url{https://www.nhr-verein.de/en/our-partners}). HoreKa is partly funded by the German Research Foundation (DFG). The authors acknowledge support by the state of Baden-Württemberg through bwHPC. This work was performed with the help of the Large Scale Data Facility at the Karlsruhe Institute of Technology funded by the Ministry of Science, Research and the Arts Baden-Württemberg and by the Federal Ministry of Education and Research.
\end{ack}


\bibliography{neurips_2026}
\bibliographystyle{splncs04}

\clearpage

\appendix

\section{Technical Appendices and Supplementary Material}

\subsection{Limitations}
\label{sec:limitations}
Next, we discuss some limitations of our work and give insight into our assessment of them.

While extensive, \benchmarkname~contains some datasets which are rather small, particularly in the medical domain.
While this is a notoriously data-scarce imaging domain, reflecting real-world settings, additional larger medical datasets would be interesting to explore.
Yet, of course, there has to be a limit, which led to the medical domain being represented by two datasets that are on the smaller side.

As we discuss in~\Cref{sec:discussion}, existing metrics, even perceptual metrics like LPIPS not always reflect the task adaptive performance faithfully.
This is especially the case for models that can pixel-wise hand-through the query, which leads to good results on some metric-task combinations, even though the model did not really address the task.
Similarly, some models can segment images, but their segmentation follows the learned color map from pre-training rather than the colors provided in the context.
By reporting results in mIoU with Hungarian matching instead of color-aware mIoU, such models would achieve better segmentation performance.
Yet, we argue, that for evaluation of task adaptation, \ie, whether a model predicts contextually, models need to base their predictions on the context and not on their color palette from pre-training.

In~\Cref{tab:vicl_ranking}, we aggregate results into rankings. Inherently, these rankings are biased by factors such as metrics used for the tasks, the number of datasets in the different domains, or, the selection of vision tasks in the benchmark.
To address these effects, task taxonomies could be explored to attribute the same weight to distinct task categories.
As stated in~\Cref{sec:discussion}, we take a more \emph{task instance} based view, considering each context set as defining a own task, which avoids the notion of task categories.
Yet, of course, there are highly correlated and less correlated task instances, which affects the ranking. 

Our idea in the \benchmarkname~toolkit is to use task-encodings which are as simple as possible.
The models we benchmark often define such tasks encodings differently, and with their task encodings the results of these models would likely be better.
As we want to measure task adaptiveness we take the view that models should be able to adapt to new task encodings, which is the premise of VICL.
Yet, this may lead to some models having better aligned task encodings that they were pre-trained on than other models, an advantage for the former.
As one of our insights is that most models that are trained on structured inputs tend to collapse to multi-task models, in practice, this effect does not take shape.

\begin{figure}[h]
    \centering
    \includegraphics[width=0.32\textwidth]{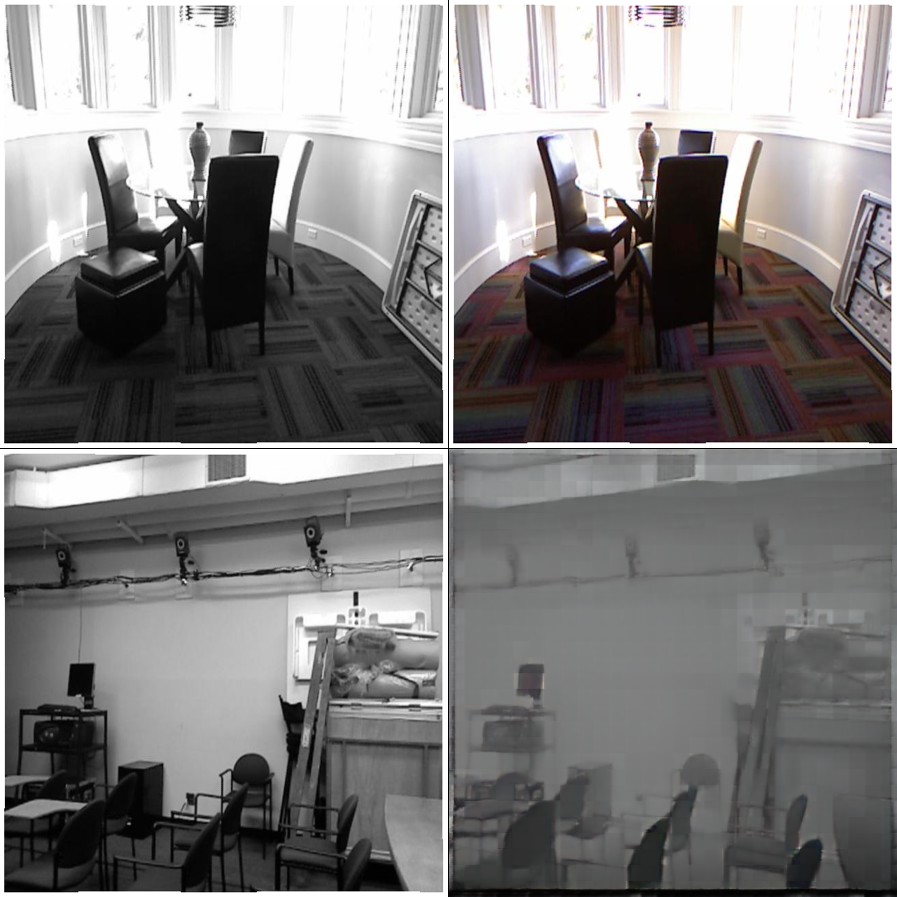}\hfill
    \includegraphics[width=0.32\textwidth]{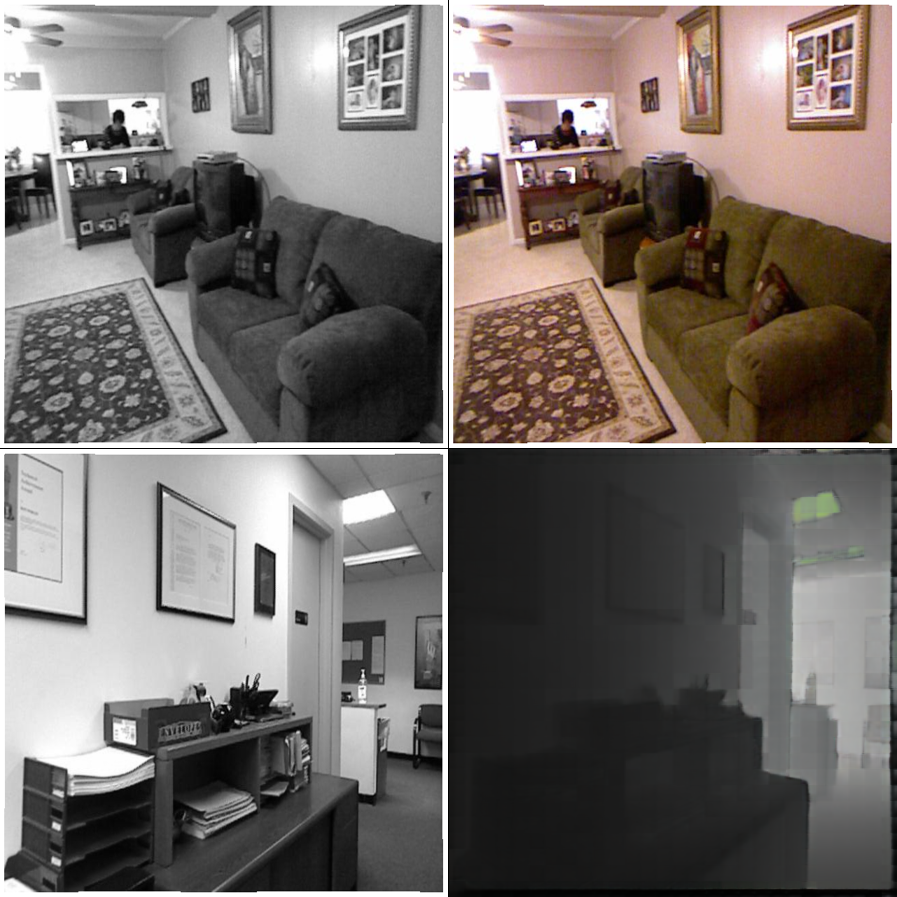}\hfill
    \includegraphics[width=0.32\textwidth]{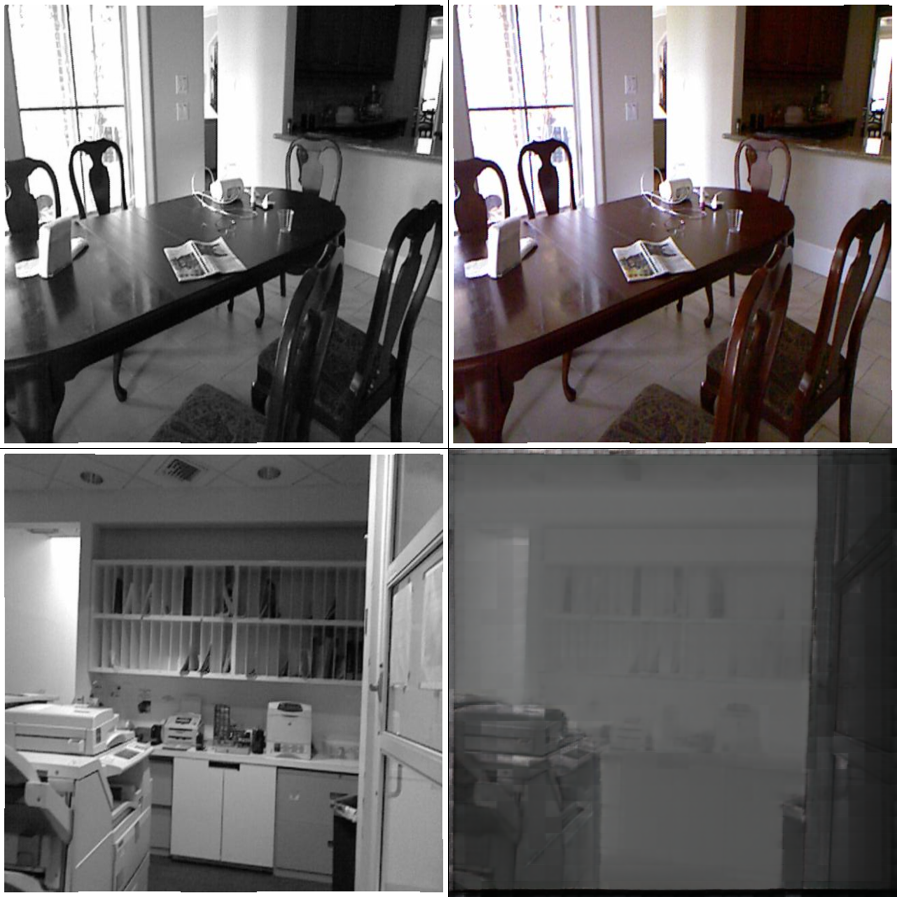}
    \caption{Qualitative examples for Painter on NYU-Depth V2, prompted with the colorization task.}
    \label{fig:painter_depth}
\end{figure}

Similarly, there are overlaps between training tasks of the VICL models and our tasks in \benchmarkname.
This is unavoidable, as there is no standardized set of training and evaluation tasks in literature, and the different works make use of many different publicly available image datasets.
Even though, some models may have been pre-trained on datasets we use for evaluation, through the different task encodings, trained models generally still fail to adapt.
A particularly noteworthy example is Painter on NYU-Depth V2, a dataset on which it was trained to estimate depth maps color coded in gray.
In~\Cref{fig:painter_depth}, rather than adapting to the tasks of colorization on NYU-Depth V2 as defined in the context (top row of image grids), Painter resorts to always produce gray depth maps (bottom right quadrant), as the pre-training bias is that strong, hindering adaptation.
This is also a hint that a model seems to not necessarily require task outputs in the context to learn a \emph{switch} as we reason in~\Cref{sec:discussion}, it can also be learned solely based on dataset statistics of the input images.

A shortcoming in this work -- which we deem to be out of scope given our extensive experiments -- is considering inference compute as part of the evaluation metrics.
Qualifying the performance metrics with compute cost gives credit to efficient models and incentivizes research into these models.
Given the vastly different inference costs of the models this is an important factor to explore in the future.

\subsection{Societal impact}
\label{sec:societal_impact}
As we build a benchmark for evaluating visual in-context learning models and do not introduce a novel trained model, we do not see concerns that might be of particularly negative societal impact.

Our toolkit benchmarks previously released and openly accessible deep learning models, and inherits the properties of these models in terms of bias and fairness with respect to the datasets and tasks they were trained on.
We do not specifically explore these biases in our one-shot experiments.
Generative models such as the openly accessible stable diffusion checkpoints used by some VICL approaches can be fine-tuned for misuse to generate deep fakes and illegal content~\cite{Internet_Watch_Foundation}.

Due to the nature of VICL -- quick adaptation to new tasks -- misusing the models is easier for a layperson as merely a malicious image analogy (an image pair and a query image) needs to be supplied and no re-training is involved.
Yet, this also goes the other way, VICL models can be key enablers for addressing a vision task of interest to a layperson in a data-efficient way, \eg, a researcher without programming skills and no budget for data annotation, a private citizen or small businesses.
Our research assesses to what extent a broad diversity of productive vision tasks can be addressed by current VICL models.

\subsection{Benchmark datasets}
In~\Cref{tab:benchmark_dataset_summary}, we provide an overview of the datasets we use in \benchmarkname, indicating the abbreviated domain name and the tasks for which the datasets provide annotations for.
Aside to these tasks, we evaluate datasets where possible on the tasks which do not specifically require annotations, such as denoising, colorization, \etc.

\begin{table}[h]
\centering
\footnotesize
\setlength{\tabcolsep}{0.75pt}
\caption{Summary of the datasets included in \benchmarkname{}. Sample counts refer to the benchmark subsets after dataset-specific filtering and preprocessing. For Duke, the two numbers correspond to the cyst segmentation and the layer edge-detection subsets, respectively.}
\label{tab:benchmark_dataset_summary}
\begin{tabular}{l p{4.7cm} p{5cm} l}
\toprule
\textbf{Dataset} & \textbf{Task(s)} & \textbf{Benchmark subset / remarks} & \textbf{Samples} \\
\midrule
\multicolumn{4}{c}{\textbf{Medical \& Biomedical Imaging}} \\
Duke            & Semantic segmentation, edge detection & Filtered scans with fluid cysts/boundaries & 78 / 110 \\
IXI             & Modality transfer & Preprocessed into 2D slices & 578 \\
\midrule
\multicolumn{4}{c}{\textbf{Underwater \& Aquatic environments}} \\
ATLANTIS        & Sem. segmentation & Waterbody segmentation dataset & 5,195 \\
SeaThru         & Image restoration & Subset with paired depth info; low-light & 1,205 \\
SUIM            & Semantic segmentation & Underwater segmentation benchmark subset & 1,635 \\
\midrule
\multicolumn{4}{c}{\textbf{Indoor \& Structural Environments}} \\
NYU Depth V2    & Depth, sem. seg. & Standard labeled benchmark subset & 1449 \\
DeepCrack       & Edge detection & Edge targets derived from crack masks & 537 \\
\midrule
\multicolumn{4}{c}{\textbf{Arts, Aesthetics \& Photography}} \\
CDD             & Image restoration & Benchmark uses the test split & 200 \\
DRAM            & Semantic segmentation & Only the labeled subset is used & 718 \\
\midrule
\multicolumn{4}{c}{\textbf{Object- and Human-Centric Scenes}} \\
OpenThermalPose & Keypoint, detection & Filtered to single-instance images & 1,950 \\
RORD            & Object removal & Benchmark-specific validation subset & 343 \\
LLVIP           & Detection, Modality transfer & Filtered to max 3 objects; every 25th frame & 608 \\
\midrule
\multicolumn{4}{c}{\textbf{Wildlife and Animal-Centric Imagery}} \\
AP-10K          & Keypoint/obj. detection & Filtered to remove multi-object images & 5,961 \\
CaltechBirds    & Keypoint/obj. detection & Randomly filtered to 30 images per class & 6,000 \\
\bottomrule
\end{tabular}
\end{table}

Next, we describe dataset-specific filtering and which tasks each dataset is evaluated on in~\benchmarkname.

\textbf{ATLANTIS}~\cite{erfani2022atlantis} We use the entire dataset as is.

Tasks in \benchmarkname: Semantic segmentation, colorization, denoising, inpainting, rotation, flip.\\

\textbf{Caltech Birds}~\cite{wah2011caltech} To have a manageable size of the dataset for benchmarking, we filter the dataset and obtain $30$ images for each of the $200$ bird classes by random selection.

Tasks in \benchmarkname: Object detection, keypoint detection (based on landmarks on birds), colorization, denoising, inpainting, rotation, flip.\\

\textbf{CDD}~\cite{guo2024onerestore} Due to the large size of the original dataset and the many different artifact types, we utilize $200$ images from the official test split for each of the artifact removal tasks.

Tasks in \benchmarkname: Artifact removal (haze $\to$ clear, haze+rain $\to$ clear, haze+snow $\to$ clear, low+haze+rain $\to$ clear, low+haze+snow $\to$ clear, low+haze $\to$ clear, low+rain $\to$ clear, low $\to$ clear, rain $\to$ clear, snow $\to$ clear, low+snow $\to$ clear), colorization, denoising, inpainting, rotation, flip.\\

\textbf{DeepCrack}~\cite{liu2019deepcrack} We use the all images in the dataset.

Tasks in \benchmarkname: Edge detection, colorization, denoising, inpainting, rotation, flip.\\

\textbf{DRAM}~\cite{cohen2022semantic} We use all images with associated segmentation annotations.

Tasks in \benchmarkname: Semantic segmentation, colorization, denoising, inpainting, rotation, flip.\\

\textbf{DUKE} For the edge task, we only use images in DUKE, which have annotated retinal layers and render the topmost and lowest layers in our encoding. For the segmentation task, we only use images which have fluid / cyst annotations.

Tasks in \benchmarkname: Edge detection, semantic segmentation, denoising, inpainting, rotation, flip.\\

\textbf{IXI}~\cite{IXIDataset} For each patient 3D volume, the most prominent slice is selected by means of highest intensity among pixels in 2D slices.
We repeat this process for the volumes from the T2 weighted and PD MRI modalities in the three dimensions, \ie, sagittal, lateral, coronal, yielding the three dataset  variants in the main paper.

Tasks in \benchmarkname~(for sagittal, lateral, coronal) : Modality transfer (T2 weighted MRI to PD), denoising, inpainting, rotation, flip.\\

\textbf{LLVIP}~\cite{jia2021llvip} In this dataset, to obtain a manageable size, we use every $25^{th}$ frame based on the provided video data.
Further, filter out images, where the bounding boxes of more than three persons overlap, in order to simplify the task encoding for multi-object detection, where overlapping boxes are represented in different color channels of the RGB image.

Tasks in \benchmarkname: Object detection, modality transfer (RGB to infrared), colorization, denoising, inpainting, rotation, flip.\\

\textbf{NYU Depth V2}~\cite{silberman2012indoor} We use the whole dataset, for segmentation, use the $40$ classes provided.

Tasks in \benchmarkname: Depth estimation, semantic segmentation, colorization, denoising, inpainting, rotation, flip.\\

\textbf{OpenThermalPose}~\cite{kuzdeuov2024openthermalpose,kuzdeuov2025openthermalpose2} We filter the dataset to only contain images with one human, which eases the process of encoding keypoints based on human poses, as they are always associated to the single human in the scene.

Tasks in \benchmarkname: Keypoint detection (from human poses), denoising, inpainting, rotation, flip.\\

\textbf{RORD mini}~\cite{sagong2022rord} As RORD is a very large dataset, we use the provided RORD mini subset (the val-343 split) to have a manageable size.

Tasks in \benchmarkname: Object removal (based on red bounding boxes), colorization, denoising, inpainting, rotation, flip.\\

\textbf{SeaThru}~\cite{akkaynak2019sea} We use all images in the dataset with the associated depth maps.

Tasks in \benchmarkname: Depth estimation, colorization, denoising, inpainting, rotation, flip.\\

\textbf{SUIM}~\cite{islam2020semantic} We use all images in the dataset with their associated segmentation annotations.

Tasks in \benchmarkname: Semantic segmentation, colorization, denoising, inpainting, rotation, flip.\\

\textbf{AP-10K}~\cite{yu2021ap} As for OpenThermalPose, we filter out images with multiple instances and images with less than 7 keypoints annotated which indicates an occlusion.

Tasks in \benchmarkname: Object detection, keypoint detection (based on animal poses), colorization, denoising, inpainting, rotation, flip.

\subsection{Benchmark tasks}
Here, we describe selection criteria, encoding and post-processing for each task in the \benchmarkname~toolkit.
 
{
\setlength{\leftmargini}{1em}
\paragraph{Segmentation}
\begin{itemize}
    \item Selection criteria: Semantic segmentation is included as a dense pixel-wise
    classification task in the benchmark. It requires structured understanding of scene content by assigning a discrete semantic class to every pixel. This directly tests whether VICL models can infer region-level semantic mappings from a single context example. It covers a broad range of domains in the benchmark, including artistic paintings (DRAM~\cite{cohen2022semantic}), underwater and aquatic environment scenes (SUIM~\cite{islam2020semantic}, ATLANTIS~\cite{erfani2022atlantis}), medical imaging (DUKE~\cite{srinivasan2014fully}) and indoor scenes (NYU Depth V2~\cite{silberman2012indoor}), making it a
    central task for evaluating cross-domain adaptation.
    \item Encoding process: Raw integer-valued masks in $H\times W$ are converted into 3-channel image encodings. In our benchmark, we use a color palette that maps each class index to a color. We use the pre-defined color map of the ADE20K dataset~\cite{zhou2017scene}. Each class index is mapped to a predefined RGB value from this palette. From this color list, we select $C$ visually distinct colors by using a greedy spread sampling strategy. We start from the darkest color, while each subsequent color is chosen to maximize its distance to the already selected colors in RGB space. This resulting mapping is then used to map it to the 3-channel target in our benchmark. See~\Cref{tab:segmentation} for an example visual analogy.
    \item Post-processing: The model's prediction is decoded into class indices by assigning each pixel in the prediction to the nearest color from the palette among the colors shown in the context examples presented to the model. The resulting discrete label map is resized with nearest-neighbor interpolation to target resolution and is then compared against the ground truth mask using \textit{mIoU} (this way of evaluating can be referred to as color-aware mIoU~\cite{bar2022visual}).
\end{itemize}

\begin{table}[h]
\centering
\normalsize
\setlength{\tabcolsep}{0pt}
\scalebox{0.9}{%
\begin{tabular}{cccc}
    \makebox[0.25\linewidth]{\centering $C_{in}$} &
    \makebox[0.25\linewidth]{\centering $C_{out}$} &
    \makebox[0.25\linewidth]{\centering $Q$} &
    \makebox[0.25\linewidth]{\centering Ground-truth} \\
    \multicolumn{4}{c}{
        \includegraphics[width=\linewidth]{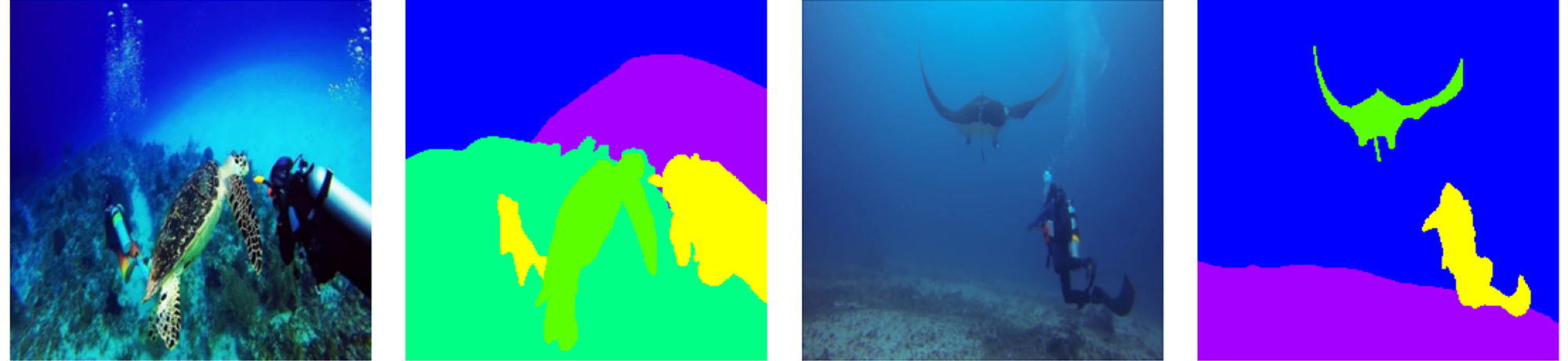}
    }
\end{tabular}%
}
\caption{Example visual analogy in the SUIM dataset for semantic segmentation encoding in \benchmarkname.}
\label{tab:segmentation}
\end{table}

\clearpage
\paragraph{Depth Estimation}
\begin{itemize}
    \item Selection criteria: Depth estimation is included as a representative continuous dense regression task. Unlike categorical dense tasks, it requires the model to infer a per-pixel quantity from the input image. This task evaluates whether VICL models can adapt not only to discrete classes but to continuous values for spatial prediction. Within the benchmark, it is instantiated on NYU Depth V2~\cite{silberman2012indoor}, which captures cluttered indoor scenes, and on SeaThru~\cite{akkaynak2019sea} to introduce underwater and low-light conditions with scattering effects to test robustness under domain-specific visual distortions.
    \item Encoding process: Raw target depth maps are stored as numeric arrays and dynamically transformed into 3-channel image representations. Continuous depth values are quantized into 256 bins and each of the bins is mapped to an RGB value using a Turbo color map\footnote{\url{https://research.google/blog/turbo-an-improved-rainbow-colormap-for-visualization/}}. This converts the scalar depth map into a 3-channel image while preserving a reversible mapping to an approximate depth range. $NaN$ values are mapped directly to the invalid value $-1.0$. See~\Cref{tab:depth} for an example visual analogy.
    \item Post-processing: Predicted RGB images are decoded by finding the nearest color for each pixel in the Turbo color map lookup table. The corresponding bin index is then mapped back to a depth value (bin center). The reconstructed depth map is then compared to the depth ground truth.
\end{itemize}

\begin{table}[h]
\centering
\normalsize
\setlength{\tabcolsep}{0pt}
\scalebox{0.9}{%
\begin{tabular}{cccc}
    \makebox[0.25\linewidth]{\centering $C_{in}$} &
    \makebox[0.25\linewidth]{\centering $C_{out}$} &
    \makebox[0.25\linewidth]{\centering $Q$} &
    \makebox[0.25\linewidth]{\centering Ground-truth} \\
    \multicolumn{4}{c}{
        \includegraphics[width=\linewidth]{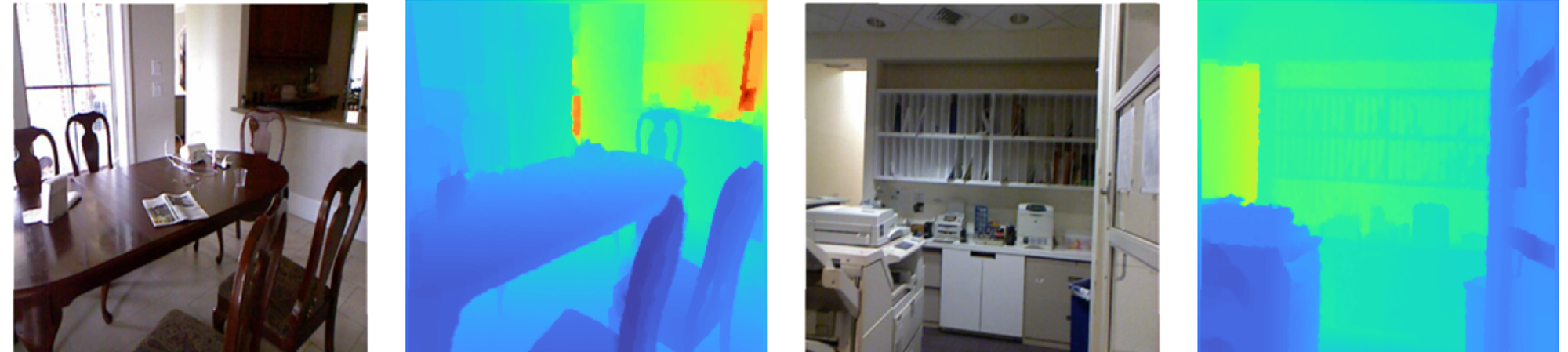}
    }
\end{tabular}%
}
\caption{Example visual analogy in the NYU-D V2 dataset for depth estimation encoding in \benchmarkname.}
\label{tab:depth}
\end{table}

\paragraph{Edge Detection}
\begin{itemize}
    \item Selection criteria: Edge detection is included as a sparse binary dense prediction task that emphasizes fine geometric structures and thin boundaries. It evaluates whether VICL models can achieve high spatial precision. We evaluate edge detection on structurally demanding datasets such as DeepCrack~\cite{liu2019deepcrack} and Duke~\cite{srinivasan2014fully}, to cover the structural and the medical domain.
    \item Encoding process: Edge annotations are represented as binary image masks with background encoded as $0$ and edge pixels as $255$. In order to get a 3-channel target image, the grayscale mask is represented as a 3-channel image by duplicating the same edge mask across channels. See~\Cref{tab:edge_detection} for an example visual analogy.
    \item Post-processing: The predicted edge images are converted into edge maps by using intensity-based thresholding. The decoded edge map is then compared with the ground truth edge annotation. 
\end{itemize}

\begin{table}[h]
\centering
\normalsize
\setlength{\tabcolsep}{0pt}
\scalebox{0.9}{%
\begin{tabular}{cccc}
    \makebox[0.25\linewidth]{\centering $C_{in}$} &
    \makebox[0.25\linewidth]{\centering $C_{out}$} &
    \makebox[0.25\linewidth]{\centering $Q$} &
    \makebox[0.25\linewidth]{\centering Ground-truth} \\
    \multicolumn{4}{c}{
        \includegraphics[width=\linewidth]{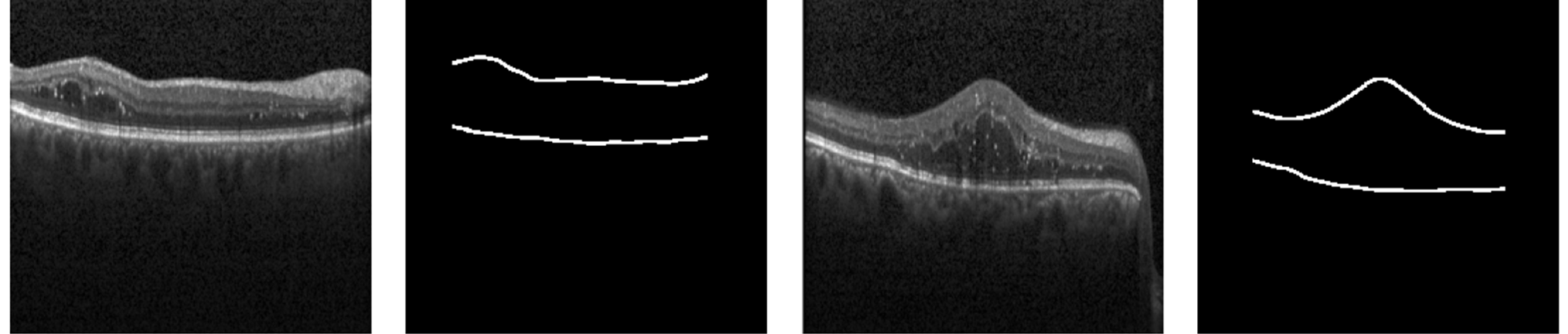}
    }
\end{tabular}%
}
\caption{Example visual analogy in the DUKE dataset for edge detection encoding in \benchmarkname.}
\label{tab:edge_detection}
\end{table}

\paragraph{Artifact Removal}
\begin{itemize}
    \item Selection criteria: Artifact removal is included as a compound degradation restoration
    task. Unlike denoising, which applies a single controlled noise type, it targets real-world degradations that occur in combination
    rather than in isolation. The task is instantiated on CDD~\cite{guo2024onerestore}, a dataset covering 11 degradation types across composite corruption settings. We evaluate heterogeneous VICL models under diverse degradation combinations to assess their robustness and capabilities to remove artifacts.
    \item Encoding process: The degraded image serves as the model input, while the corresponding clean restored image is retained as the target for evaluation. See~\Cref{tab:artifact_removal} for an example visual analogy.
    \item Post-processing: The predicted output is resized to the spatial resolution of the target and is evaluated using standard image quality metrics PSNR, SSIM and LPIPS.
\end{itemize}

\begin{table}[h]
\centering
\normalsize
\setlength{\tabcolsep}{0pt}
\scalebox{0.9}{%
\begin{tabular}{cccc}
    \makebox[0.25\linewidth]{\centering $C_{in}$} &
    \makebox[0.25\linewidth]{\centering $C_{out}$} &
    \makebox[0.25\linewidth]{\centering $Q$} &
    \makebox[0.25\linewidth]{\centering Ground-truth} \\
    \multicolumn{4}{c}{
        \includegraphics[width=\linewidth]{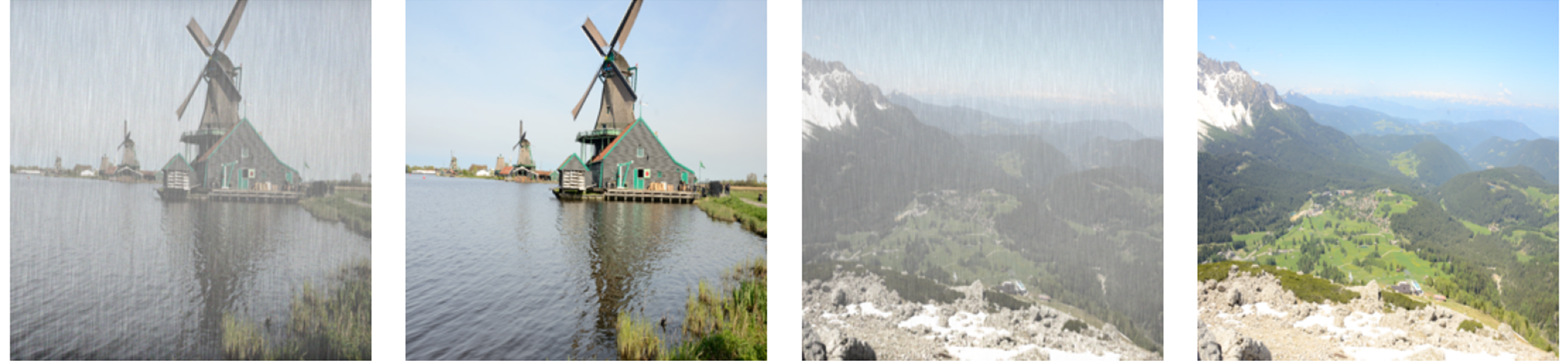}
    }
\end{tabular}%
}
\caption{Example visual analogy in the CDD dataset for artifact removal encoding in \benchmarkname.}
\label{tab:artifact_removal}
\end{table}

\paragraph{Object Removal}
\begin{itemize}
    \item Selection criteria: Object removal is included as a spatially conditioned image generation task that requires the model to remove a designated object while preserving scene structure. It is the most demanding manipulation task in the benchmark, as it requires inferring from the bounding box that the enclosed region corresponds to an object to be removed, suppressing it and synthesizing plausible background content to fill the resulting gap. The task is instantiated on a subset of RORD~\cite{sagong2022rord}, a large-scale real-world dataset with paired before-and-after images and precise annotations. 
    \item Encoding process: The query is the original RGB image annotated with a red bounding box indicating the object region to be removed, while its corresponding target is the same image with the specific object removed. This preserves the full scene while providing an explicit spatial cue for the removal target. The same construction of the bounding box surrounding the target is applied to $C_{in}$. See~\Cref{tab:object_removal} for an example visual analogy.
    \item Post-processing: The predicted output is resized to the spatial resolution of the target and is evaluated using standard image quality metrics PSNR, SSIM and LPIPS.
\end{itemize}
\begin{table}[h]
\centering
\normalsize
\setlength{\tabcolsep}{0pt}
\scalebox{0.9}{%
\begin{tabular}{cccc}
    \makebox[0.25\linewidth]{\centering $C_{in}$} &
    \makebox[0.25\linewidth]{\centering $C_{out}$} &
    \makebox[0.25\linewidth]{\centering $Q$} &
    \makebox[0.25\linewidth]{\centering Ground-truth} \\
    \multicolumn{4}{c}{
        \includegraphics[width=\linewidth]{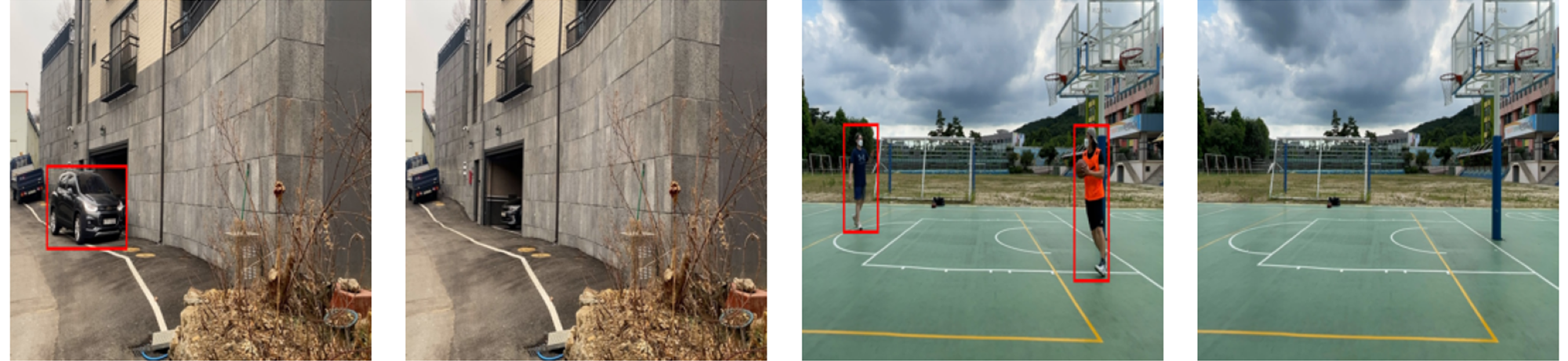}
    }
\end{tabular}%
}
\caption{Example visual analogy in the RORD dataset for object removal encoding in \benchmarkname.}
\label{tab:object_removal}
\end{table}

\paragraph{Keypoint Detection}
\begin{itemize}
    \item Selection criteria: Keypoint detection is included as a sparse localization task in the benchmark. It requires predicting a small set of precise point locations rather than dense per-pixel labels, which tests a different form of structured spatial inference. It is instantiated across both human-centric (OpenThermalPose~\cite{kuzdeuov2024openthermalpose,kuzdeuov2025openthermalpose2}) and animal-centric (CaltechBirds~\cite{wah2011caltech}, AP-10K~\cite{yu2021ap}) datasets, which cover humans and different animals with substantial pose- and species variation. This breadth makes it a strong test case of whether VICL models can adapt to fine-grained localization in different domains from a single context example.
    \item Encoding process: Keypoint annotations are originally provided as sparse coordinate locations and are encoded as 3-channel target images to be compatible for prompting VICL approaches. Visible keypoints are clipped to the valid image region and are rendered as filled red circles with a fixed radius of 5 pixels on a black background. All keypoints use the same red marker instead of category-specific colors, which keeps the target representation simple and avoids any ambiguities from overlapping visual encodings. The original coordinates are still retained for postprocessing. See~\Cref{tab:keypoint_detection} for an example visual analogy.
    \item Post-processing: The predicted keypoint prediction is decoded using a redness score map which is calculated by $S = \text{clip}(R - \max(G,B), 0, 1)$. The score map is thresholded to obtain a binary keypoint mask. For point-level evaluation, connected components are extracted and each component is reduced to the pixel with the highest redness score. Predicted points are then matched to visible ground truth keypoints based on spatial proximity using a threshold $\tau = 2r$, where $r$ is the radius of the rendered ground truth keypoints.
\end{itemize}

\begin{table}[h]
\centering
\normalsize
\setlength{\tabcolsep}{0pt}
\scalebox{0.9}{%
\begin{tabular}{cccc}
    \makebox[0.25\linewidth]{\centering $C_{in}$} &
    \makebox[0.25\linewidth]{\centering $C_{out}$} &
    \makebox[0.25\linewidth]{\centering $Q$} &
    \makebox[0.25\linewidth]{\centering Ground-truth} \\
    \multicolumn{4}{c}{
        \includegraphics[width=\linewidth]{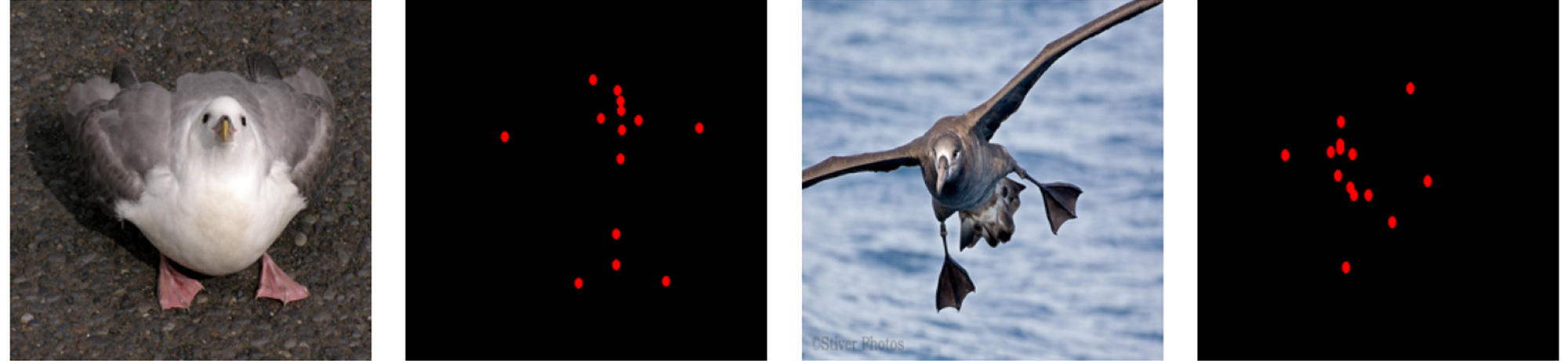}
    }
\end{tabular}%
}
\caption{Example visual analogy in Caltech Birds dataset for keypoint detection encoding in \benchmarkname.}
\label{tab:keypoint_detection}
\end{table}

\paragraph{Modality Transfer}
\begin{itemize}
    \item Selection criteria: Modality transfer is included as it focuses on cross-modality image-to-image mappings within a unified framework, requiring VICL models to infer an appearance transformation rule rather than a fixed pattern from a single context pair. The task is instantiated on infrared-to-RGB transfer (LLVIP~\cite{jia2021llvip}) and T2-weighted to proton density (PD) MRI transfer (IXI~\cite{IXIDataset}), covering object- and human-centric as well as medical imaging domains.
    \item Encoding process: Each sample consists of an input-output pair, which is derived from a source and a corresponding target domain. See~\Cref{tab:modality_transfer} for an example visual analogy.
    \item Post-processing: The post-processing stage is restricted to quantitative evaluation of the generated outputs. Model predictions are compared against the corresponding ground-truth target images using PSNR, SSIM and LPIPS for evaluation.
\end{itemize}
\begin{table}[h]
\centering
\normalsize
\setlength{\tabcolsep}{0pt}
\scalebox{0.9}{%
\begin{tabular}{cccc}
    \makebox[0.25\linewidth]{\centering $C_{in}$} &
    \makebox[0.25\linewidth]{\centering $C_{out}$} &
    \makebox[0.25\linewidth]{\centering $Q$} &
    \makebox[0.25\linewidth]{\centering Ground-truth} \\
    \multicolumn{4}{c}{
        \includegraphics[width=\linewidth]{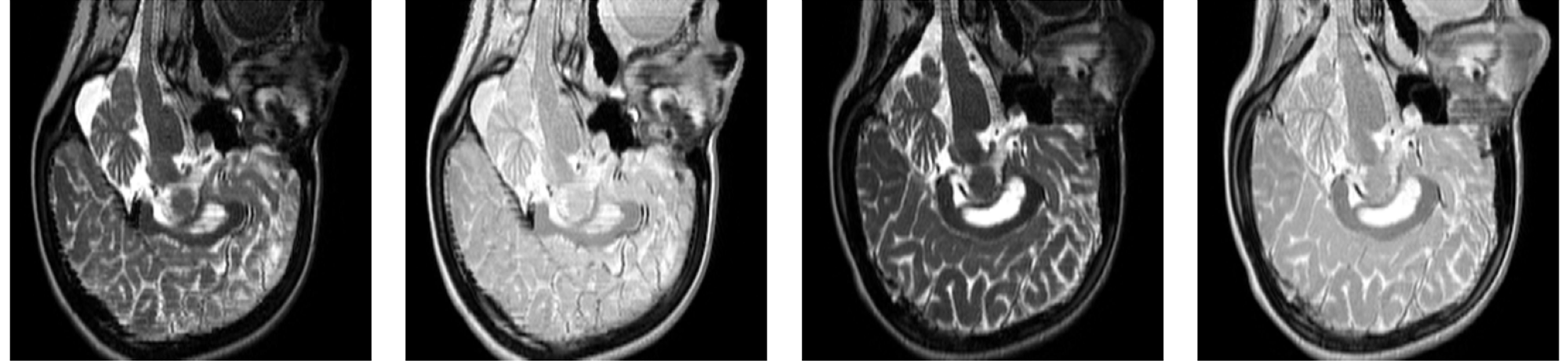}
    }
\end{tabular}%
}
\caption{Example visual analogy in the IXI dataset for modality transfer encoding in \benchmarkname.}
\label{tab:modality_transfer}
\end{table}

\paragraph{Object Detection}
\begin{itemize}
    \item Selection criteria: Object detection is included as a structured spatial localization task which requires models to identify and localize a single- or multiple objects. The task is instantiated on LLVIP~\cite{jia2021llvip}, which introduces pedestrian detection under low-light and infrared conditions, AP-10K~\cite{yu2021ap} and CaltechBirds~\cite{wah2011caltech}, which both provide animal-centric annotations. 
    \item Encoding process: Bounding box annotations are rendered into 3-channel image targets. Each object is represented as a bounding box drawn as a filled colored rectangle on black background. To preserve overlapping objects, boxes are assigned to separate RGB channels: red by default, green if they overlap existing red boxes and blue if they overlap both red and green boxes. Samples with more than three overlapping boxes are excluded. See~\Cref{tab:object_detection} for an example visual analogy.
    \item Post-processing: Predicted RGB outputs are thresholded separately per channel and connected-component analysis is used to extract object regions. Each connected component is then converted into a bounding box, aligned to the image axis, with using its mean intensity as a confidence score for mAP ranking. Boxes are matched to ground truth using IoU and evaluated with mAP@0.5.
\end{itemize}

\begin{table}[h]
\centering
\normalsize
\setlength{\tabcolsep}{0pt}

\begin{tabular}{cccc}
    \makebox[0.25\linewidth]{\centering $C_{in}$} &
    \makebox[0.25\linewidth]{\centering $C_{out}$} &
    \makebox[0.25\linewidth]{\centering $Q$} &
    \makebox[0.25\linewidth]{\centering Ground-truth} \\
    \multicolumn{4}{c}{
        \includegraphics[width=\linewidth]{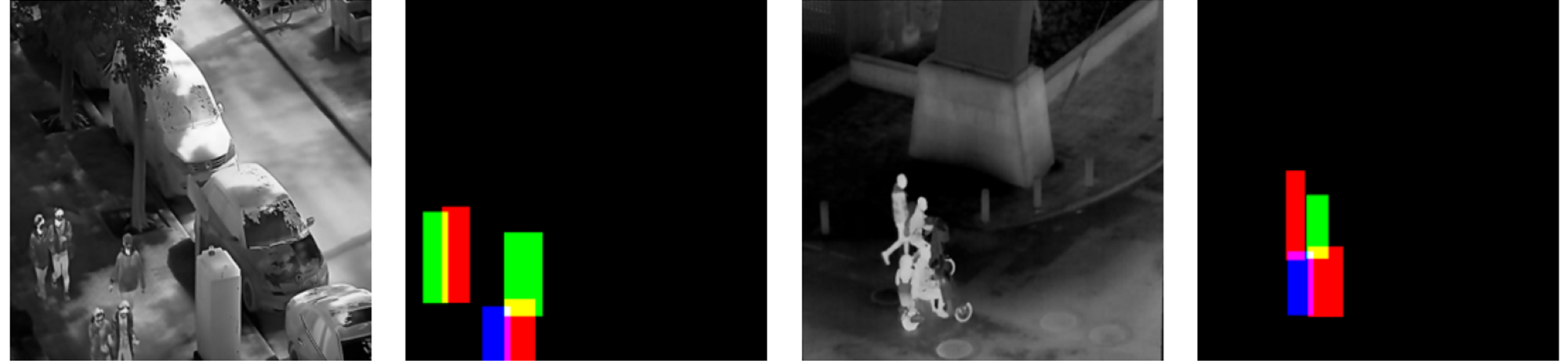}
    }
\end{tabular}

\caption{Example visual analogy in the LLVIP dataset for object detection encoding in \benchmarkname.}
\label{tab:object_detection}
\end{table}

\paragraph{Inpainting}
\begin{itemize}
    \item Selection criteria: Inpainting is included as the representative region-level reconstruction
    task. It is used to test whether VICL models can
    complete missing image regions by inferring plausible content from surrounding visual context and whether they can infer the reconstruction pattern demonstrated in the prompt. The task is applied across all datasets and domains in the benchmark.
    \item Encoding process: For each sample, a binary mask is generated and applied to the original RGB image. The masked pixels are replaced by a fixed fill value (black, \ie, pixel value~$0$) while unmasked regions remain unchanged. The resulting masked image serves as the query input, while the original uncorrupted RGB image is retained as the target for evaluation. The same binary mask is applied to $C_{in}$. See~\Cref{tab:inpainting} for an example visual analogy.
    \item Post-processing: The predicted output is resized to the spatial resolution of the target. Evaluation is executed by using the standard image quality metrics PSNR, SSIM and LPIPS.
\end{itemize}

\begin{table}[h]
\centering
\normalsize
\setlength{\tabcolsep}{0pt}
\scalebox{0.9}{%
\begin{tabular}{cccc}
    \makebox[0.25\linewidth]{\centering $C_{in}$} &
    \makebox[0.25\linewidth]{\centering $C_{out}$} &
    \makebox[0.25\linewidth]{\centering $Q$} &
    \makebox[0.25\linewidth]{\centering Ground-truth} \\
    \multicolumn{4}{c}{
        \includegraphics[width=\linewidth]{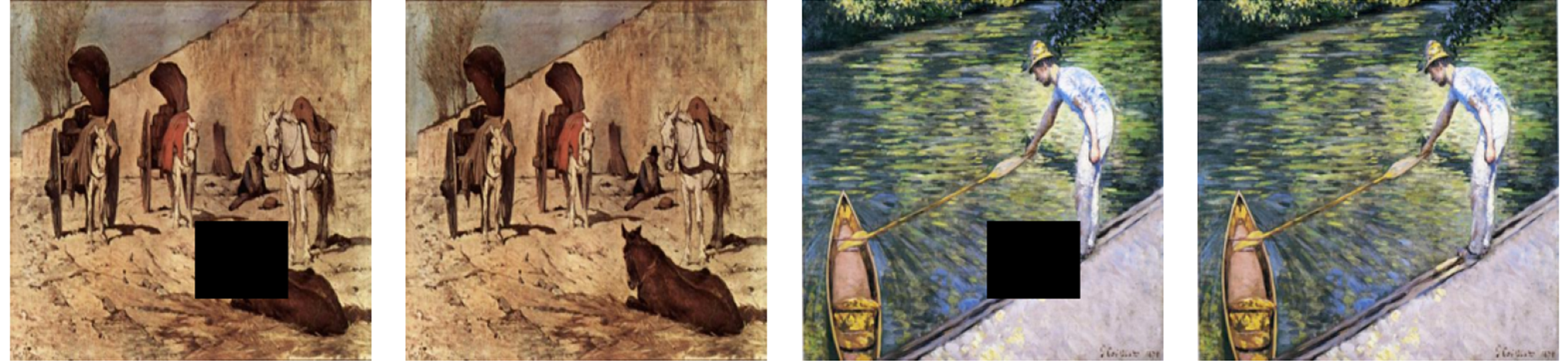}
    }
\end{tabular}%
}
\caption{Example visual analogy in the DRAM dataset for inpainting encoding in \benchmarkname.}
\label{tab:inpainting}
\end{table}

\clearpage
\paragraph{Denoising}
\begin{itemize}
    \item Selection criteria: Denoising is included as a pixel-level signal recovery task. Because the corruption is applied using a fixed, deterministically generated noise configuration, it provides a fully controlled and reproducible low-level restoration challenge. It tests whether VICL models can infer the noise-removal objective from a noisy-to-clean context pair and apply it faithfully to an unseen query.
    \item Encoding process: Additive Gaussian noise with a fixed standard deviation of $\sigma=0.15$
    is applied to the original clean RGB image to produce the noisy model input. The original image is retained as the ground-truth target for evaluation. The same noise construction is also applied to $C_{in}$. The noise realization
    for each sample is generated deterministically from the sample index, ensuring reproducibility while producing non-identical corruptions across samples. See~\Cref{tab:denoising} for an example visual analogy.
    \item Post-processing: The predicted output is resized to the spatial resolution of the target and is evaluated using standard image quality metrics PSNR, SSIM and LPIPS.
\end{itemize}
\begin{table}[h]
\centering
\normalsize
\setlength{\tabcolsep}{0pt}
\scalebox{0.9}{%
\begin{tabular}{cccc}
    \makebox[0.25\linewidth]{\centering $C_{in}$} &
    \makebox[0.25\linewidth]{\centering $C_{out}$} &
    \makebox[0.25\linewidth]{\centering $Q$} &
    \makebox[0.25\linewidth]{\centering Ground-truth} \\
    \multicolumn{4}{c}{
        \includegraphics[width=\linewidth]{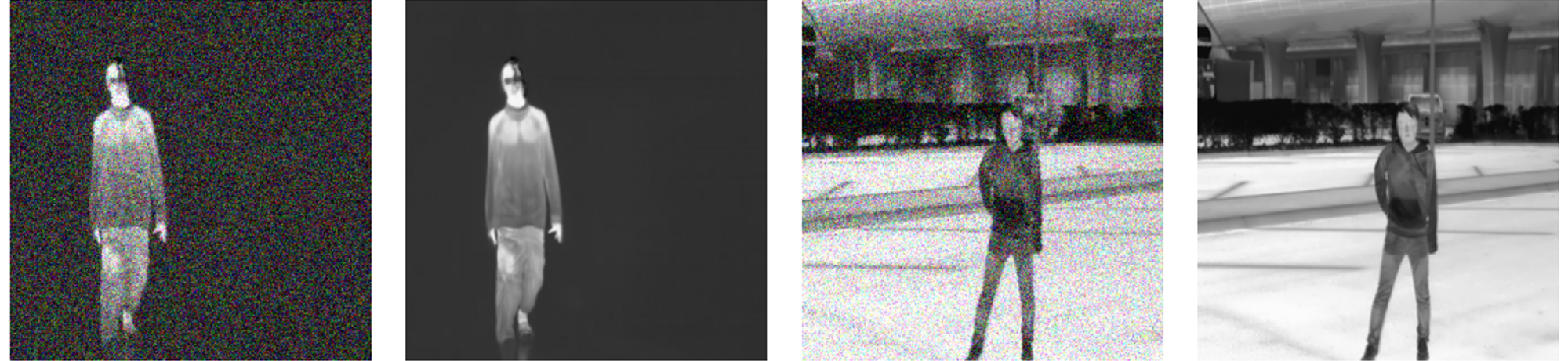}
    }
\end{tabular}%
}
\caption{Example visual analogy in the OTP dataset for denoising encoding in \benchmarkname.}
\label{tab:denoising}
\end{table}

\paragraph{Image Colorization}
\begin{itemize}
    \item Selection criteria: Colorization is included as a perceptual inference task within the Image Restoration. Unlike denoising or inpainting, it requires the model to infer plausible color information from a grayscale input. This task evaluates whether VICL models can learn an appearance mapping from a single context example rather than simply recovering a known corruption. 
    \item Encoding process: The original RGB image is converted to a grayscale representation using the standard luminance transform $\text{GRAY} = 0.299\,R + 0.587\,G + 0.114\,B$. The resulting single channel is repeated across three channels so that the model receives a 3-channel input. For evaluation, the original RGB image is retained as the target. See~\Cref{tab:colorization} for an example visual analogy.
    \item Post-processing: The predicted output is resized to the spatial resolution of the target and is evaluated using standard image quality metrics PSNR, SSIM and LPIPS.
\end{itemize}
\begin{table}[h]
\centering
\normalsize
\setlength{\tabcolsep}{0pt}
\scalebox{0.9}{%
\begin{tabular}{cccc}
    \makebox[0.25\linewidth]{\centering $C_{in}$} &
    \makebox[0.25\linewidth]{\centering $C_{out}$} &
    \makebox[0.25\linewidth]{\centering $Q$} &
    \makebox[0.25\linewidth]{\centering Ground-truth} \\
    \multicolumn{4}{c}{
        \includegraphics[width=\linewidth]{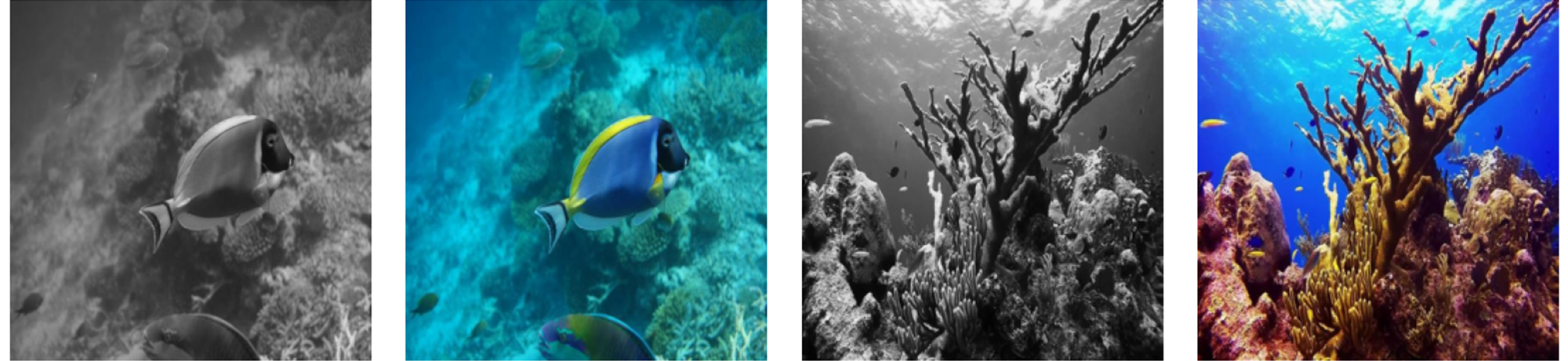}
    }
\end{tabular}%
}
\caption{Example visual analogy in the SUIM dataset for image colorization encoding in \benchmarkname.}
\label{tab:colorization}
\end{table}

\paragraph{Geometric Transformations}
\begin{itemize}
    \item Selection criteria: With geometric tasks, we probe, whether VICL models can handle simple transformations like rotating or flipping an image, where the task does not involve altering the image content but geometrically transforming it.
    \item Encoding process: The original RGB image serves as the model input. The target is obtained
    by applying the respective geometric transformation directly to that image. We include a $90^{\circ}$-rotated- and a horizontally flipped version. See~\Cref{tab:geometric_transform} for an example visual analogy for rotation.
    \item Post-processing: The predicted output is resized to the spatial resolution of the ground-truth target. This is followed by using the shared image-to-image protocol using PSNR, SSIM and LPIPS for evaluation.
\end{itemize}
\begin{table}[h]
\centering
\normalsize
\setlength{\tabcolsep}{0pt}
\scalebox{0.9}{%
\begin{tabular}{cccc}
    \makebox[0.25\linewidth]{\centering $C_{in}$} &
    \makebox[0.25\linewidth]{\centering $C_{out}$} &
    \makebox[0.25\linewidth]{\centering $Q$} &
    \makebox[0.25\linewidth]{\centering Ground-truth} \\
    \multicolumn{4}{c}{
        \includegraphics[width=\linewidth]{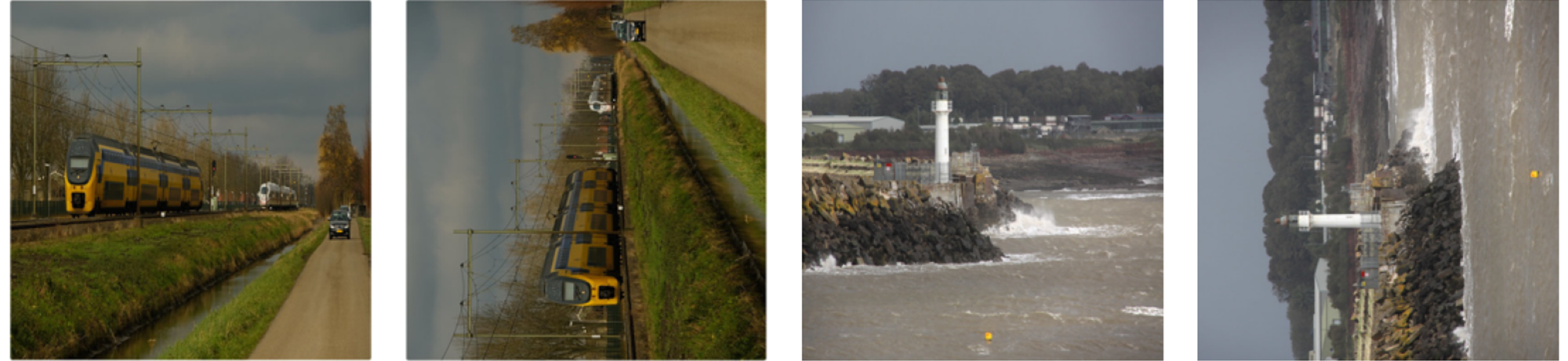}
    }
\end{tabular}%
}
\caption{Example visual analogy in the ATLANTIS dataset for geometric transformations encoding of Rotation with $90^{\circ}$ in \benchmarkname.}
\label{tab:geometric_transform}
\end{table}
}

\subsection{Benchmark implementation pipelines}
Next, we briefly describe how we have considered the notions of extensibility and reproducibility in~\benchmarkname.
For details on implementation, we refer to our source code\footnote{\url{https://github.com/Simael/visual-in-context-benchmark}}.
Our code is structured into two stages, the first for dataset download and pre-processing, the second for model inference.

\textbf{Stage 1: Offline dataset preparation.}
In the first stage, the datasets are downloaded automatically and further converted into a canonical format and put into a unified folder structure.
In this pre-processing, each dataset is stored in a separate folder with subfolders for different annotation types.
During this process, each image is assigned an ID.
We then deterministically pre-compute for each image ID a list of context samples for later prompt construction and store it into a CSV file.
This ensures, reproducible runs with fixed context sets for each query image.
We extract a list of context samples to enable future exploration beyond one-shot, towards few-shot prompting experiments.
New datasets and tasks can be added to the benchmark by following the steps described in the source code, making extensibility a core principle of the \benchmarkname~toolkit.

\textbf{Stage 2: Online inference and evaluation.}
In the second stage, the prepared data can be loaded by specifying input and output folders based on the previously described folder structure.
Further, a model can be specified, which determines how the context set and query image is provided to the model at inference time.
Then, the CSV file for prompt construction is loaded and inference is done for each entry in the file using the specified contexts.
For less disk space requirements, some task targets are computed at runtime (\eg, inpainting, denoising, \etc).
The predictions of the specified model are then decoded and evaluated with a set of task-specific metrics.
The results are persistently logged into CSV files.
Additionally, it can be specified to log qualitative results during inference.
Here, the \benchmarkname~toolkit can also be extended towards including new models and prompt strategies.

A small note: In the source code of our \benchmarkname~toolkit, we also provide train / validation / test splits ($25\% / 25 \% / 50 \%$), even though in the experiments of the main paper we do not train models, and therefore just evaluate on the test split.
We do this with extensibility in mind, \eg, to enable future development of visual in-context learners through hold-tasks-out cross-validation or to allow for development of test-time fine-tuning methods tailored for VICL~\cite{xie2025test}.

\subsection{Rank aggregation}
To compare models across tasks with different metrics and value ranges, we report a rank-aggregated benchmark score in \Cref{tab:vicl_ranking}. We compute this ranking by aggregating performance the level of tasks to avoid averaging different metrics and value ranges, such as mIoU, PSNR, LPIPS or RMSE.

For each task $t \in T$, we first compute the mean score of each model $m$ for every metric $k$ across all datasets $d \in D_t$ that belong to this task:
$\bar{\mu}_{m,k}^{(t)} = \frac{1}{|D_t|} \sum_{d \in D_t} \mu_{m,d,k}.$
This gives each dataset the same weight within a task, independent of the number of samples it contains. We then rank the models separately for each metric. For error metrics, lower values receive better ranks, while for performance metrics, higher values receive better ranks. This yields metric-specific ranks $r_{m,k}^{(t)}$, where rank 1 corresponds to the best model.

The final rank for a task is obtained by averaging over all metrics $K_t$ used for this task:
$\bar{r}_{m}^{(t)} = \frac{1}{|K_t|} \sum_{k \in K_t} r_{m,k}^{(t)}$.
The overall benchmark rank is then computed by averaging these task-level ranks over all tasks:
\[
\bar{r}_{m}
=
\frac{1}{|T|}
\sum_{t \in T}
\bar{r}_{m}^{(t)}.
\]
Lower values therefore indicate stronger overall performance across the benchmark.

This rank aggregation avoids comparing raw metric values across incompatible scales and instead measures how consistently a model performs relative to the other methods. It also prevents individual datasets or metrics from dominating the final ranking. Additionally, in \Cref{tab:vicl_ranking}, we show the task-level ranks $\bar{r}_{m}^{(t)}$ to indicate where each model performs well or poorly before computing the final average shown on the far right.
Such rankings on multi-task benchmarks are very useful to get a clearer view on how models compare to each other, especially on such a diverse evaluation as ours, yet,  naturally nuance can be lost~\cite{zhang2024inherent} when viewing them in isolation.


\subsection{Compute utilized for all experiments}
\label{sec:compute_used}
All computation in this benchmarking effort was carried out on NVIDIA A100 GPUs.
Considering all experiments, we utilized computation for a wallclock time of $160.875$ days with a total energy consumption of $16,805.07$ Megajoule ($4,668.08$ Kilowatt-hours).
The average memory utilized was $8.31$ GB although this varies with the models tested and the batch size which was used in inference: MAE-VQGAN ($12.05$ GB, batch size $20$), Painter ($6.39$ GB, batch size $4$), DeLVM ($5.20$ GB), LVM ($15.40$ GB), PromptDiffusion ($10.23$ GB), SD-VICL ($4.93$ GB).
\Cref{tab:compute} summarizes the wall-clock time per model.
While this includes re-runs and is mainly an indication of the compute we used to collect the results in this paper, it can be seen, that SD-VICL comes with significantly increased inference time as compared to the other models with more than half the total wall-clock time.

\begin{table}[h!]
    \centering
    \caption{Computation statistics for carrying out the experiments as in the paper. These numbers do not indicate the precise efficiency of each model as this includes some computation jobs that failed, or which had to be re-run, rather it is an indication of how much compute was expended per model.}
    \begin{tabular}{rrr}
        \toprule
         Method & Total Memory (GB) & Total Wallclock time (h:m:s)\\
         \midrule
         MAE-VQGAN & $843.49$ & $12:06:12$\\
         Painter &  $767.20$  & $21:36:58$\\
         DeLVM & $1,216.98$ & $258:31:52$\\
         LVM & $3,603.21$& $350:41:20$ \\
         PromptDiffusion & $2,372.44$& $995:48:19$ \\
         SD-VICL & $1,113.55$ & $2,163:23:17$ \\
         \bottomrule
    \end{tabular}
    \label{tab:compute}
\end{table}

\subsection{Model architecture overview}
\label{sec:modelarchitectures}
In~\Cref{tab:model_parameters}, we compare the architectures of the VICL models that we compare in our experiments.

\begin{table}[h]
\caption{Overview of the compared VICL approaches and their main architectural properties.\\
$^\ast$ Painter parameter count refers to the reported ViT-L backbone setting. \\
$^{\ast\ast}$ MAE-VQGAN parameter count refers to the ViT-L-based inpainting model. \\
$^{\ast\ast\ast}$ Combined parameters of frozen Stable Diffusion and trainable ControlNet.}
\label{tab:model_parameters}

\centering
\begin{tabular}{l l r c c}
\toprule
Model & Backbone & Parameters & Codebook (size) & Resolution \\
\midrule
Painter~\cite{wang2023images} & ViT-L~\cite{vit_visiontransformer} & 307M$^\ast$ & -- & $448 \times 448$ \\
MAE-VQGAN~\cite{bar2022visual} & ViT-MAE~\cite{he2022masked} & 307M$^{\ast\ast}$ & VQGAN (1024) & $224 \times 224$ \\
LVM~\cite{bai2024sequential} & LLaMA~\cite{touvron2023llama} & 7B & VQGAN (8192) & $256 \times 256$ \\
DeLVM~\cite{delvm} & LLaMA~\cite{touvron2023llama} & 1B & VQGAN (8192) & $256 \times 256$ \\
PromptDiffusion~\cite{promptdiffusion} & Stable Diffusion~\cite{rombach2022high} & 1.2B$^{\ast\ast\ast}$& VAE-KL & $512 \times 512$ \\
SD-VICL~\cite{oorloff2025stable} & Stable Diffusion~\cite{rombach2022high} &  860M & VAE-KL & $512 \times 512$ \\
\bottomrule
\end{tabular}

\label{tab:compared_models}
\end{table}

\subsection{Licenses of datasets and used codebases}
In~\Cref{tab:dataset_metainformation} and~\Cref{tab:code_metainformation}, we provide meta information on datasets and code we utilize in~\benchmarkname.

\begin{table}[h]
    \centering
    \caption{Meta information on the datasets we utilize in \benchmarkname.}
    \label{tab:dataset_metainformation}
    \begin{tabular}{rll}
    \toprule
         Dataset name & License & Link\\
         \midrule
         DUKE & -- & \tiny \url{https://people.duke.edu/~sf59/Chiu_BOE_2014_dataset.htm}\\
         IXI & \href{https://creativecommons.org/licenses/by-sa/3.0/legalcode}{CC BY-SA 3.0} & \tiny \url{https://brain-development.org/ixi-dataset/} \\
         ATLANTIS & \href{https://github.com/smhassanerfani/atlantis#dataset-description}{Custom}& \tiny \url{https://github.com/smhassanerfani/atlantis} \\
         SUIM & \href{https://github.com/xahidbuffon/SUIM/blob/master/LICENSE}{	
MIT License}& \tiny \url{https://irvlab.cs.umn.edu/resources/suim-dataset} \\
         Sea-thru & \href{https://creativecommons.org/licenses/by-nc-sa/4.0/}{CC BY-NC-SA 4.0} & \tiny \url{https://www.kaggle.com/datasets/colorlabeilat/seathru-dataset}\\
         NYU Depth v2 & Citation & \tiny \url{https://cs.nyu.edu/~fergus/datasets/nyu_depth_v2.html}\\
         DeepCrack & \href{https://github.com/qinnzou/DeepCrack#copy-right}{Custom} & \tiny \url{https://github.com/qinnzou/DeepCrack} \\
         DRAM & \href{https://github.com/Nadavc220/SemanticSegmentationInArtPaintings#terms-of-use}{Citation} & \tiny \url{https://nadavc220.github.io/ssiap/} \\
         CDD & -- & \tiny \url{https://github.com/gy65896/OneRestore}\\
         OpenThermalPose & \href{https://github.com/IS2AI/OpenThermalPose/blob/main/LICENSE}{MIT License} & \tiny \url{https://github.com/IS2AI/OpenThermalPose} \\
         LLVIP & \href{https://github.com/bupt-ai-cz/LLVIP/blob/main/Term%20of%20Use%20and%20License.md}{Custom License} & \tiny \url{https://github.com/bupt-ai-cz/LLVIP}\\
         RORD & \href{https://github.com/IIDA-AILab/RORD/blob/main/LICENSE}{MIT License} & \tiny \url{https://github.com/IIDA-AILab/RORD}\\
         Caltech Birds& Citation & \tiny \url{https://www.vision.caltech.edu/datasets/cub_200_2011/} \\
         AP-10K& \href{https://github.com/AlexTheBad/AP-10K/blob/main/LICENSE}{CC BY 4.0} & \tiny \url{https://github.com/AlexTheBad/AP-10K} \\
    \bottomrule
    \end{tabular}
\end{table}

\begin{table}[h]
    \centering
    \caption{Meta information on the code we utilize in \benchmarkname.}
    \label{tab:code_metainformation}
    \begin{tabular}{rll}
    \toprule
         Code & License & Link\\
         \midrule
         MAE-VQGAN~\cite{bar2022visual} & -- & \tiny \url{https://github.com/amirbar/visual_prompting}\\
         Painter~\cite{wang2023images} & \href{https://github.com/baaivision/Painter/blob/main/LICENSE}{MIT License} &  \tiny \url{https://github.com/baaivision/Painter/tree/main/Painter}\\
         DeLVM~\cite{delvm} & \href{https://github.com/ggjy/DeLVM/tree/main#license}{MIT License} &  \tiny \url{https://github.com/ggjy/DeLVM/tree/main}\\
         LVM~\cite{bai2024sequential} & \href{https://github.com/ytongbai/LVM/blob/main/LICENSE}{Apache License 2.0}&  \tiny \url{https://github.com/ytongbai/LVM}\\
         PromptDiffusion~\cite{promptdiffusion} & \href{https://github.com/Zhendong-Wang/Prompt-Diffusion/blob/main/LICENSE}{Apache License 2.0} & \tiny \url{https://github.com/Zhendong-Wang/Prompt-Diffusion} \\
         SD-VICL~\cite{oorloff2025stable} re-impl. & \href{https://github.com/garibida/cross-image-attention/blob/main/LICENSE}{MIT License} & \tiny \url{https://github.com/garibida/cross-image-attention/tree/main} \\
        based on~\cite{alaluf2024cross}&& \\
    \bottomrule
    \end{tabular}
\end{table}

We would like to acknowledge the following python packages in particular, as we make substantial use of them in the \benchmarkname~toolkit: Pytorch~\cite{NEURIPS2019_9015}, Numpy~\cite{harris2020array}, Pandas~\cite{McKinney.2010}, nibabel~\cite{markiewicz_2026_18944560}, Torchvision~\cite{torchvision2016}, scikit-learn~\cite{scikit-learn}, SciPy~\cite{2020SciPy-NMeth}, OpenCV~\cite{opencv_library}, matplotlib~\cite{Hunter:2007} and h5py~\cite{collette_python_hdf5_2014}.

\clearpage
\subsection{Numerical results as tables}
\label{sec:numericalresults}
In this section, we provide the numerical results of our experiments.
The mean and standard deviation were computed along each prediction in the dataset-task setting of a model, \eg, in~\Cref{tab:segmentation_mean_iou}, on the ATLANTIS dataset, for all predictions of DeLVM, the mIoU across the classes is calculated and for these values mean and standard deviation are reported.  

\subsubsection{Perception \& Localization}
\begin{table}[h]
\centering
\footnotesize
\renewcommand{\arraystretch}{0.88}
\setlength{\tabcolsep}{4pt}
\caption{Segmentation results by dataset (mIoU $\uparrow$). Mean $\pm$ standard deviation.}
\label{tab:segmentation_mean_iou}
\resizebox{\textwidth}{!}{%
\begin{tabular}{lccccc}
\toprule
\textbf{Model} & \textbf{ATLANTIS} & \textbf{DRAM} & \textbf{Duke} & \textbf{NYU-D V2} & \textbf{SUIM} \\
\midrule
DELVM 1B & 0.125 $\pm$ 0.133 & 0.236 $\pm$ 0.211 & 0.000 $\pm$ 0.002 & 0.025 $\pm$ 0.028 & 0.208 $\pm$ 0.173 \\
LVM & 0.049 $\pm$ 0.072 & 0.247 $\pm$ 0.232 & 0.009 $\pm$ 0.032 & 0.026 $\pm$ 0.029 & 0.156 $\pm$ 0.152 \\
MAE-VQGAN & 0.100 $\pm$ 0.106 & 0.411 $\pm$ 0.161 & \textbf{0.087 $\pm$ 0.132} & 0.041 $\pm$ 0.033 & 0.225 $\pm$ 0.152 \\
Painter & 0.028 $\pm$ 0.064 & 0.171 $\pm$ 0.223 & 0.035 $\pm$ 0.058 & 0.008 $\pm$ 0.013 & 0.171 $\pm$ 0.183 \\
PromptDiff. & 0.030 $\pm$ 0.050 & 0.097 $\pm$ 0.105 & 0.015 $\pm$ 0.023 & 0.010 $\pm$ 0.015 & 0.119 $\pm$ 0.094 \\
SD-VICL & \textbf{0.229 $\pm$ 0.151} & \textbf{0.415 $\pm$ 0.177} & 0.055 $\pm$ 0.108 & \textbf{0.122 $\pm$ 0.069} & \textbf{0.292 $\pm$ 0.192} \\
\midrule
Copy target & 0.148 $\pm$ 0.101 & 0.337 $\pm$ 0.116 & 0.039 $\pm$ 0.087 & 0.082 $\pm$ 0.041 & 0.203 $\pm$ 0.127 \\
\bottomrule
\end{tabular}
}
\end{table}

\begin{table}[h]
\centering
\footnotesize
\renewcommand{\arraystretch}{0.88}
\setlength{\tabcolsep}{3.5pt}

\begin{minipage}[t]{0.47\textwidth}
\centering
\caption{Depth estimation results by dataset (RMSE $\downarrow$). Mean $\pm$ standard deviation.}
\label{tab:depth_rmse}
\resizebox{\textwidth}{!}{%
\begin{tabular}{lcc}
\toprule
\textbf{Model} & \textbf{NYU-D V2} & \textbf{Sea-Thru} \\
\midrule
DELVM 1B & 3.096 $\pm$ 0.907 & 2.872 $\pm$ 2.665 \\
LVM & 2.727 $\pm$ 0.884 & 4.341 $\pm$ 4.311 \\
MAE-VQGAN & 1.747 $\pm$ 0.767 & 2.158 $\pm$ 1.646 \\
Painter & 2.704 $\pm$ 0.658 & 3.583 $\pm$ 1.793 \\
PromptDiff. & 2.760 $\pm$ 0.782 & \textbf{2.157 $\pm$ 2.023} \\
SD-VICL & \textbf{1.478 $\pm$ 0.689} & 2.283 $\pm$ 1.486 \\
\midrule
Copy target & 1.596 $\pm$ 0.712 & 1.637 $\pm$ 1.243 \\
\bottomrule
\end{tabular}
}
\end{minipage}
\hfill
\begin{minipage}[t]{0.47\textwidth}
\centering
\caption{Edge detection results by dataset (Edge F1 $\uparrow$). Mean $\pm$ standard deviation.}
\label{tab:edge_detection_edge_f1}
\resizebox{\textwidth}{!}{%
\begin{tabular}{lcc}
\toprule
\textbf{Model} & \textbf{DeepCrack} & \textbf{Duke} \\
\midrule
DELVM 1B & 0.003 $\pm$ 0.028 & 0.000 $\pm$ 0.000 \\
LVM & 0.082 $\pm$ 0.171 & 0.008 $\pm$ 0.020 \\
MAE-VQGAN & 0.289 $\pm$ 0.211 & 0.019 $\pm$ 0.029 \\
Painter & 0.019 $\pm$ 0.116 & 0.030 $\pm$ 0.019 \\
PromptDiff. & 0.195 $\pm$ 0.141 & 0.011 $\pm$ 0.007 \\
SD-VICL & \textbf{0.625 $\pm$ 0.225} & \textbf{0.538 $\pm$ 0.120} \\
\midrule
Copy target & 0.037 $\pm$ 0.052 & 0.056 $\pm$ 0.075 \\
\bottomrule
\end{tabular}
}
\end{minipage}

\end{table}

\begin{table}[h]
\centering
\footnotesize
\renewcommand{\arraystretch}{0.88}
\setlength{\tabcolsep}{3.5pt}

\begin{minipage}[t]{0.47\textwidth}
\vspace{0pt}
\centering
\caption{Object detection results by dataset (mAP@0.5 $\uparrow$). Mean $\pm$ standard deviation.}
\label{tab:object_detection_map50}
\resizebox{\textwidth}{!}{%
\begin{tabular}{lccc}
\toprule
\textbf{Model} & \textbf{AP-10K} & \textbf{Caltech Birds} & \textbf{LLVIP} \\
\midrule
DELVM 1B & 0.244 $\pm$ 0.382 & 0.197 $\pm$ 0.367 & 0.039 $\pm$ 0.109 \\
LVM & \textbf{0.731 $\pm$ 0.438} & 0.688 $\pm$ 0.455 & 0.072 $\pm$ 0.180 \\
MAE-VQGAN & 0.646 $\pm$ 0.478 & 0.348 $\pm$ 0.476 & 0.000 $\pm$ 0.000 \\
Painter & 0.624 $\pm$ 0.448 & \textbf{0.721 $\pm$ 0.435} & 0.246 $\pm$ 0.314 \\
PromptDiff. & 0.089 $\pm$ 0.244 & 0.232 $\pm$ 0.407 & 0.029 $\pm$ 0.111 \\
SD-VICL & 0.635 $\pm$ 0.480 & 0.523 $\pm$ 0.497 & \textbf{0.290 $\pm$ 0.351} \\
\midrule
Copy target & 0.526 $\pm$ 0.499 & 0.383 $\pm$ 0.486 & 0.012 $\pm$ 0.058 \\
\bottomrule
\end{tabular}
}
\end{minipage}
\hfill
\begin{minipage}[t]{0.47\textwidth}
\vspace{0pt}
\centering
\caption{Keypoint detection results by dataset (F1 $\uparrow$). Mean $\pm$ standard deviation.}
\label{tab:keypoint_detection_kp_f1}
\resizebox{\textwidth}{!}{%
\begin{tabular}{lccc}
\toprule
\textbf{Model} & \textbf{AP-10K} & \textbf{Caltech Birds} & \textbf{OTP} \\
\midrule
DELVM 1B & 0.002 $\pm$ 0.011 & 0.007 $\pm$ 0.031 & 0.041 $\pm$ 0.059 \\
LVM & 0.011 $\pm$ 0.023 & 0.021 $\pm$ 0.045 & \textbf{0.087 $\pm$ 0.073} \\
MAE-VQGAN & 0.000 $\pm$ 0.006 & 0.023 $\pm$ 0.042 & 0.026 $\pm$ 0.038 \\
Painter & 0.007 $\pm$ 0.027 & 0.034 $\pm$ 0.062 & 0.019 $\pm$ 0.049 \\
PromptDiff. & 0.001 $\pm$ 0.007 & 0.001 $\pm$ 0.009 & 0.016 $\pm$ 0.022 \\
SD-VICL & \textbf{0.039 $\pm$ 0.063} & \textbf{0.142 $\pm$ 0.088} & 0.073 $\pm$ 0.116 \\
\midrule
Copy target & 0.004 $\pm$ 0.011 & 0.027 $\pm$ 0.035 & 0.011 $\pm$ 0.021 \\
\bottomrule
\end{tabular}
}
\end{minipage}
\end{table}

\subsubsection{Image Restoration and Enhancement}
\begin{table}[h!]
\centering
\footnotesize
\renewcommand{\arraystretch}{0.88}
\setlength{\tabcolsep}{4pt}
\begin{minipage}[t]{0.31\textwidth}
\vspace{0pt}
\centering
\caption{Artifact removal results by dataset (PSNR $\uparrow$). Mean $\pm$ standard deviation.}
\label{tab:artifact_removal_psnr}
\begin{tabular}{lc}
\toprule
\textbf{Model} & \textbf{CDD} \\
\midrule
DELVM 1B & 11.112 $\pm$ 4.608 \\
LVM & 14.502 $\pm$ 2.560 \\
MAE-VQGAN & 12.985 $\pm$ 2.099 \\
Painter & \textbf{15.313 $\pm$ 3.663} \\
PromptDiff. & 9.904 $\pm$ 1.437 \\
SD-VICL & 13.470 $\pm$ 2.454 \\
\midrule
Copy target & 9.378 $\pm$ 1.623 \\
\bottomrule
\end{tabular}
\end{minipage}
\hfill
\begin{minipage}[t]{0.31\textwidth}
\vspace{0pt}
\centering
\caption{Artifact removal results by dataset (SSIM $\uparrow$). Mean $\pm$ standard deviation.}
\label{tab:artifact_removal_ssim}
\begin{tabular}{lc}
\toprule
\textbf{Model} & \textbf{CDD} \\
\midrule
DELVM 1B & 0.327 $\pm$ 0.188 \\
LVM & 0.386 $\pm$ 0.160 \\
MAE-VQGAN & 0.358 $\pm$ 0.157 \\
Painter & \textbf{0.517 $\pm$ 0.148} \\
PromptDiff. & 0.272 $\pm$ 0.118 \\
SD-VICL & 0.398 $\pm$ 0.130 \\
\midrule
Copy target & 0.246 $\pm$ 0.129 \\
\bottomrule
\end{tabular}
\end{minipage}
\hfill
\begin{minipage}[t]{0.31\textwidth}
\centering
\caption{Artifact removal results by dataset (LPIPS $\downarrow$). Mean $\pm$ standard deviation.}
\label{tab:artifact_removal_lpips}
\begin{tabular}{lc}
\toprule
\textbf{Model} & \textbf{CDD} \\
\midrule
DELVM 1B & 0.676 $\pm$ 0.126 \\
LVM & 0.654 $\pm$ 0.080 \\
MAE-VQGAN & 0.721 $\pm$ 0.082 \\
Painter & \textbf{0.590 $\pm$ 0.120} \\
PromptDiff. & 0.842 $\pm$ 0.069 \\
SD-VICL & 0.611 $\pm$ 0.099 \\
\midrule
Copy target & 0.747 $\pm$ 0.081 \\
\bottomrule
\end{tabular}
\end{minipage}

\end{table}

\begin{table}[h]
\centering
\footnotesize
\renewcommand{\arraystretch}{0.88}
\setlength{\tabcolsep}{4pt}
\caption{Denoising results by dataset (PSNR $\uparrow$) . Mean $\pm$ standard deviation.}
\label{tab:denoising_psnr_p1}
\resizebox{\textwidth}{!}{%
\begin{tabular}{lcccccccc}
\toprule
\textbf{Model} & \textbf{AP-10K} & \textbf{ATLANTIS} & \textbf{Caltech Birds} & \textbf{CDD} & \textbf{DeepCrack} & \textbf{DRAM} & \textbf{Duke} & \textbf{IXI (Axial)} \\
\midrule
DELVM 1B & 7.908 $\pm$ 2.974 & 7.831 $\pm$ 3.481 & 7.649 $\pm$ 3.363 & 7.012 $\pm$ 3.318 & 10.350 $\pm$ 4.249 & 9.226 $\pm$ 4.522 & 14.669 $\pm$ 6.548 & 13.381 $\pm$ 3.335 \\
LVM & 18.252 $\pm$ 2.858 & 18.693 $\pm$ 3.079 & \textbf{19.636 $\pm$ 3.276} & \textbf{17.426 $\pm$ 2.790} & \textbf{21.984 $\pm$ 3.596} & 19.006 $\pm$ 2.588 & 21.093 $\pm$ 0.895 & 18.405 $\pm$ 0.890 \\
MAE-VQGAN & 15.273 $\pm$ 1.776 & 15.337 $\pm$ 1.853 & 15.953 $\pm$ 1.984 & 14.679 $\pm$ 1.806 & 17.284 $\pm$ 1.852 & 15.382 $\pm$ 1.723 & 17.918 $\pm$ 1.117 & 15.837 $\pm$ 0.943 \\
Painter & \textbf{19.213 $\pm$ 3.471} & \textbf{19.560 $\pm$ 4.061} & 18.266 $\pm$ 4.707 & 16.960 $\pm$ 4.976 & 21.269 $\pm$ 2.302 & \textbf{21.154 $\pm$ 2.973} & 13.368 $\pm$ 1.597 & 13.072 $\pm$ 1.264 \\
PromptDiff. & 11.725 $\pm$ 1.999 & 10.403 $\pm$ 1.672 & 10.959 $\pm$ 2.131 & 9.857 $\pm$ 1.518 & 11.233 $\pm$ 2.936 & 10.179 $\pm$ 1.529 & 7.818 $\pm$ 0.482 & 8.381 $\pm$ 0.412 \\
SD-VICL & 14.103 $\pm$ 2.650 & 13.389 $\pm$ 2.672 & 13.385 $\pm$ 3.313 & 14.012 $\pm$ 2.537 & 15.830 $\pm$ 3.519 & 13.128 $\pm$ 3.105 & \textbf{21.384 $\pm$ 2.454} & \textbf{23.416 $\pm$ 2.387} \\
\midrule
Copy target & 9.509 $\pm$ 1.575 & 9.117 $\pm$ 1.627 & 9.958 $\pm$ 2.123 & 9.378 $\pm$ 1.630 & 14.206 $\pm$ 2.747 & 9.347 $\pm$ 1.761 & 12.301 $\pm$ 2.110 & 13.417 $\pm$ 1.120 \\
\bottomrule
\end{tabular}
}
\end{table}

\begin{table}[h]
\centering
\footnotesize
\renewcommand{\arraystretch}{0.88}
\setlength{\tabcolsep}{4pt}
\caption{Denoising results by dataset (PSNR $\uparrow$). Mean $\pm$ standard deviation.}
\label{tab:denoising_psnr_p2}
\resizebox{\textwidth}{!}{%
\begin{tabular}{lcccccccc}
\toprule
\textbf{Model} & \textbf{IXI (Coronal)} & \textbf{IXI (Sagittal)} & \textbf{LLVIP} & \textbf{NYU-D V2} & \textbf{OTP} & \textbf{RORD} & \textbf{Sea-Thru} & \textbf{SUIM} \\
\midrule
DELVM 1B & 10.315 $\pm$ 2.947 & 13.966 $\pm$ 4.203 & 9.842 $\pm$ 2.898 & 12.405 $\pm$ 4.872 & 14.659 $\pm$ 5.876 & 7.623 $\pm$ 3.003 & 21.427 $\pm$ 5.142 & 9.581 $\pm$ 4.219 \\
LVM & 18.215 $\pm$ 0.694 & 18.772 $\pm$ 0.895 & \textbf{20.528 $\pm$ 2.389} & 18.002 $\pm$ 2.518 & \textbf{24.943 $\pm$ 2.384} & 19.442 $\pm$ 3.281 & 24.013 $\pm$ 1.390 & \textbf{19.896 $\pm$ 3.495} \\
MAE-VQGAN & 14.958 $\pm$ 0.687 & 15.730 $\pm$ 0.823 & 17.357 $\pm$ 1.375 & 14.098 $\pm$ 1.274 & 19.487 $\pm$ 2.116 & 15.621 $\pm$ 1.741 & \textbf{27.346 $\pm$ 2.979} & 16.658 $\pm$ 2.152 \\
Painter & 16.091 $\pm$ 0.742 & 14.648 $\pm$ 0.704 & 16.900 $\pm$ 2.171 & \textbf{18.157 $\pm$ 4.152} & 18.048 $\pm$ 1.798 & \textbf{20.097 $\pm$ 3.882} & 10.275 $\pm$ 2.066 & 19.374 $\pm$ 4.685 \\
PromptDiff. & 9.118 $\pm$ 0.386 & 10.321 $\pm$ 0.247 & 9.926 $\pm$ 1.269 & 9.267 $\pm$ 1.493 & 13.821 $\pm$ 2.185 & 10.783 $\pm$ 1.768 & 11.295 $\pm$ 0.394 & 10.511 $\pm$ 1.643 \\
SD-VICL & \textbf{23.851 $\pm$ 1.813} & \textbf{24.560 $\pm$ 1.903} & 19.552 $\pm$ 3.291 & 16.266 $\pm$ 3.587 & 19.171 $\pm$ 6.509 & 14.048 $\pm$ 2.981 & 27.050 $\pm$ 2.855 & 13.062 $\pm$ 2.510 \\
\midrule
Copy target & 12.199 $\pm$ 0.674 & 13.692 $\pm$ 0.825 & 12.150 $\pm$ 2.102 & 8.930 $\pm$ 1.420 & 13.224 $\pm$ 4.048 & 9.576 $\pm$ 1.834 & 25.947 $\pm$ 2.798 & 10.467 $\pm$ 1.964 \\
\bottomrule
\end{tabular}
}
\end{table}

\begin{table}[h!]
\centering
\footnotesize
\renewcommand{\arraystretch}{0.88}
\setlength{\tabcolsep}{4pt}
\caption{Denoising results by dataset (SSIM $\uparrow$) . Mean $\pm$ standard deviation.}
\label{tab:denoising_ssim_p1}
\resizebox{\textwidth}{!}{%
\begin{tabular}{lcccccccc}
\toprule
\textbf{Model} & \textbf{AP-10K} & \textbf{ATLANTIS} & \textbf{Caltech Birds} & \textbf{CDD} & \textbf{DeepCrack} & \textbf{DRAM} & \textbf{Duke} & \textbf{IXI (Axial)} \\
\midrule
DELVM 1B & 0.114 $\pm$ 0.127 & 0.167 $\pm$ 0.177 & 0.149 $\pm$ 0.207 & 0.207 $\pm$ 0.190 & 0.222 $\pm$ 0.191 & 0.139 $\pm$ 0.125 & 0.196 $\pm$ 0.092 & 0.331 $\pm$ 0.113 \\
LVM & 0.324 $\pm$ 0.163 & 0.405 $\pm$ 0.187 & 0.511 $\pm$ 0.201 & 0.374 $\pm$ 0.174 & 0.360 $\pm$ 0.212 & 0.355 $\pm$ 0.132 & 0.396 $\pm$ 0.059 & 0.267 $\pm$ 0.054 \\
MAE-VQGAN & 0.306 $\pm$ 0.135 & 0.360 $\pm$ 0.158 & 0.410 $\pm$ 0.169 & 0.369 $\pm$ 0.157 & 0.294 $\pm$ 0.161 & 0.292 $\pm$ 0.104 & 0.374 $\pm$ 0.069 & 0.472 $\pm$ 0.045 \\
Painter & \textbf{0.577 $\pm$ 0.131} & \textbf{0.627 $\pm$ 0.128} & \textbf{0.707 $\pm$ 0.160} & \textbf{0.495 $\pm$ 0.153} & \textbf{0.659 $\pm$ 0.130} & \textbf{0.692 $\pm$ 0.073} & 0.368 $\pm$ 0.040 & 0.375 $\pm$ 0.039 \\
PromptDiff. & 0.227 $\pm$ 0.094 & 0.246 $\pm$ 0.120 & 0.301 $\pm$ 0.146 & 0.255 $\pm$ 0.118 & 0.205 $\pm$ 0.108 & 0.255 $\pm$ 0.090 & 0.170 $\pm$ 0.018 & 0.026 $\pm$ 0.021 \\
SD-VICL & 0.360 $\pm$ 0.116 & 0.418 $\pm$ 0.131 & 0.544 $\pm$ 0.140 & 0.399 $\pm$ 0.131 & 0.358 $\pm$ 0.142 & 0.412 $\pm$ 0.098 & \textbf{0.443 $\pm$ 0.054} & \textbf{0.796 $\pm$ 0.044} \\
\midrule
Copy target & 0.162 $\pm$ 0.082 & 0.200 $\pm$ 0.112 & 0.266 $\pm$ 0.136 & 0.246 $\pm$ 0.129 & 0.191 $\pm$ 0.120 & 0.160 $\pm$ 0.064 & 0.269 $\pm$ 0.055 & 0.429 $\pm$ 0.052 \\
\bottomrule
\end{tabular}
}
\end{table}

\begin{table}[h!]
\centering
\footnotesize
\renewcommand{\arraystretch}{0.88}
\setlength{\tabcolsep}{4pt}
\caption{Denoising results by dataset (SSIM $\uparrow$) . Mean $\pm$ standard deviation.}
\label{tab:denoising_ssim_p2}
\resizebox{\textwidth}{!}{%
\begin{tabular}{lcccccccc}
\toprule
\textbf{Model} & \textbf{IXI (Coronal)} & \textbf{IXI (Sagittal)} & \textbf{LLVIP} & \textbf{NYU-D V2} & \textbf{OTP} & \textbf{RORD} & \textbf{Sea-Thru} & \textbf{SUIM} \\
\midrule
DELVM 1B & 0.196 $\pm$ 0.081 & 0.275 $\pm$ 0.056 & 0.237 $\pm$ 0.173 & 0.319 $\pm$ 0.084 & 0.531 $\pm$ 0.399 & 0.164 $\pm$ 0.185 & 0.371 $\pm$ 0.167 & 0.212 $\pm$ 0.205 \\
LVM & 0.384 $\pm$ 0.038 & 0.339 $\pm$ 0.042 & 0.578 $\pm$ 0.147 & 0.484 $\pm$ 0.132 & 0.809 $\pm$ 0.136 & 0.399 $\pm$ 0.233 & 0.542 $\pm$ 0.064 & 0.460 $\pm$ 0.207 \\
MAE-VQGAN & 0.320 $\pm$ 0.045 & 0.392 $\pm$ 0.040 & 0.582 $\pm$ 0.112 & 0.426 $\pm$ 0.090 & 0.741 $\pm$ 0.172 & 0.375 $\pm$ 0.205 & \textbf{0.601 $\pm$ 0.121} & 0.437 $\pm$ 0.175 \\
Painter & 0.643 $\pm$ 0.041 & 0.504 $\pm$ 0.032 & 0.670 $\pm$ 0.097 & \textbf{0.657 $\pm$ 0.129} & \textbf{0.828 $\pm$ 0.046} & \textbf{0.580 $\pm$ 0.170} & 0.218 $\pm$ 0.054 & \textbf{0.618 $\pm$ 0.170} \\
PromptDiff. & 0.020 $\pm$ 0.032 & 0.063 $\pm$ 0.013 & 0.469 $\pm$ 0.103 & 0.389 $\pm$ 0.104 & 0.412 $\pm$ 0.097 & 0.334 $\pm$ 0.184 & 0.215 $\pm$ 0.043 & 0.224 $\pm$ 0.095 \\
SD-VICL & \textbf{0.860 $\pm$ 0.029} & \textbf{0.806 $\pm$ 0.041} & \textbf{0.697 $\pm$ 0.122} & 0.616 $\pm$ 0.131 & 0.768 $\pm$ 0.221 & 0.389 $\pm$ 0.175 & 0.517 $\pm$ 0.110 & 0.420 $\pm$ 0.131 \\
\midrule
Copy target & 0.199 $\pm$ 0.042 & 0.342 $\pm$ 0.039 & 0.427 $\pm$ 0.134 & 0.270 $\pm$ 0.068 & 0.647 $\pm$ 0.240 & 0.221 $\pm$ 0.139 & 0.686 $\pm$ 0.088 & 0.270 $\pm$ 0.143 \\
\bottomrule
\end{tabular}
}
\end{table}

\begin{table}[h!]
\centering
\footnotesize
\renewcommand{\arraystretch}{0.88}
\setlength{\tabcolsep}{4pt}
\caption{Denoising results by dataset (LPIPS $\downarrow$) . Mean $\pm$ standard deviation.}
\label{tab:denoising_lpips_p1}
\resizebox{\textwidth}{!}{%
\begin{tabular}{lcccccccc}
\toprule
\textbf{Model} & \textbf{AP-10K} & \textbf{ATLANTIS} & \textbf{Caltech Birds} & \textbf{CDD} & \textbf{DeepCrack} & \textbf{DRAM} & \textbf{Duke} & \textbf{IXI (Axial)} \\
\midrule
DELVM 1B & 0.747 $\pm$ 0.087 & 0.749 $\pm$ 0.090 & 0.757 $\pm$ 0.100 & 0.764 $\pm$ 0.093 & 0.741 $\pm$ 0.147 & 0.705 $\pm$ 0.105 & 0.628 $\pm$ 0.071 & 0.480 $\pm$ 0.044 \\
LVM & 0.614 $\pm$ 0.056 & 0.564 $\pm$ 0.069 & 0.510 $\pm$ 0.093 & 0.631 $\pm$ 0.056 & 0.583 $\pm$ 0.064 & 0.562 $\pm$ 0.042 & 0.593 $\pm$ 0.022 & 0.393 $\pm$ 0.036 \\
MAE-VQGAN & 0.701 $\pm$ 0.054 & 0.659 $\pm$ 0.065 & 0.615 $\pm$ 0.060 & 0.722 $\pm$ 0.074 & 0.643 $\pm$ 0.061 & 0.647 $\pm$ 0.040 & 0.633 $\pm$ 0.022 & 0.371 $\pm$ 0.033 \\
Painter & \textbf{0.502 $\pm$ 0.095} & \textbf{0.475 $\pm$ 0.101} & \textbf{0.438 $\pm$ 0.124} & 0.659 $\pm$ 0.139 & \textbf{0.451 $\pm$ 0.070} & \textbf{0.377 $\pm$ 0.067} & 0.657 $\pm$ 0.030 & 0.447 $\pm$ 0.021 \\
PromptDiff. & 0.818 $\pm$ 0.055 & 0.738 $\pm$ 0.060 & 0.790 $\pm$ 0.104 & 0.850 $\pm$ 0.069 & 0.701 $\pm$ 0.043 & 0.803 $\pm$ 0.078 & 0.682 $\pm$ 0.024 & 0.668 $\pm$ 0.012 \\
SD-VICL & 0.638 $\pm$ 0.081 & 0.589 $\pm$ 0.080 & 0.599 $\pm$ 0.124 & \textbf{0.615 $\pm$ 0.098} & 0.557 $\pm$ 0.077 & 0.573 $\pm$ 0.074 & \textbf{0.422 $\pm$ 0.017} & \textbf{0.218 $\pm$ 0.029} \\
\midrule
Copy target & 0.770 $\pm$ 0.059 & 0.740 $\pm$ 0.057 & 0.789 $\pm$ 0.084 & 0.747 $\pm$ 0.082 & 0.635 $\pm$ 0.057 & 0.766 $\pm$ 0.057 & 0.550 $\pm$ 0.050 & 0.370 $\pm$ 0.039 \\
\bottomrule
\end{tabular}
}
\end{table}

\begin{table}[h]
\centering
\footnotesize
\renewcommand{\arraystretch}{0.88}
\setlength{\tabcolsep}{4pt}
\caption{Denoising results by dataset (LPIPS $\downarrow$) . Mean $\pm$ standard deviation.}
\label{tab:denoising_lpips_p2}
\resizebox{\textwidth}{!}{%
\begin{tabular}{lcccccccc}
\toprule
\textbf{Model} & \textbf{IXI (Coronal)} & \textbf{IXI (Sagittal)} & \textbf{LLVIP} & \textbf{NYU-D V2} & \textbf{OTP} & \textbf{RORD} & \textbf{Sea-Thru} & \textbf{SUIM} \\
\midrule
DELVM 1B & 0.619 $\pm$ 0.103 & 0.535 $\pm$ 0.079 & 0.689 $\pm$ 0.076 & 0.610 $\pm$ 0.077 & 0.461 $\pm$ 0.208 & 0.781 $\pm$ 0.091 & 0.678 $\pm$ 0.072 & 0.705 $\pm$ 0.120 \\
LVM & 0.377 $\pm$ 0.034 & 0.399 $\pm$ 0.030 & 0.592 $\pm$ 0.057 & 0.516 $\pm$ 0.058 & 0.396 $\pm$ 0.067 & 0.603 $\pm$ 0.066 & 0.634 $\pm$ 0.043 & 0.514 $\pm$ 0.078 \\
MAE-VQGAN & 0.419 $\pm$ 0.024 & 0.414 $\pm$ 0.023 & 0.628 $\pm$ 0.048 & 0.601 $\pm$ 0.037 & 0.422 $\pm$ 0.132 & 0.699 $\pm$ 0.075 & 0.607 $\pm$ 0.059 & 0.594 $\pm$ 0.077 \\
Painter & 0.263 $\pm$ 0.023 & 0.358 $\pm$ 0.020 & 0.542 $\pm$ 0.037 & 0.448 $\pm$ 0.076 & 0.381 $\pm$ 0.040 & \textbf{0.534 $\pm$ 0.086} & 0.700 $\pm$ 0.049 & \textbf{0.441 $\pm$ 0.093} \\
PromptDiff. & 0.677 $\pm$ 0.017 & 0.739 $\pm$ 0.016 & 0.621 $\pm$ 0.035 & 0.760 $\pm$ 0.066 & 0.761 $\pm$ 0.052 & 0.750 $\pm$ 0.090 & 0.817 $\pm$ 0.025 & 0.793 $\pm$ 0.062 \\
SD-VICL & \textbf{0.165 $\pm$ 0.025} & \textbf{0.221 $\pm$ 0.027} & \textbf{0.429 $\pm$ 0.046} & \textbf{0.445 $\pm$ 0.066} & \textbf{0.351 $\pm$ 0.165} & 0.651 $\pm$ 0.096 & \textbf{0.546 $\pm$ 0.045} & 0.592 $\pm$ 0.108 \\
\midrule
Copy target & 0.416 $\pm$ 0.038 & 0.372 $\pm$ 0.032 & 0.615 $\pm$ 0.093 & 0.676 $\pm$ 0.043 & 0.441 $\pm$ 0.194 & 0.773 $\pm$ 0.071 & 0.551 $\pm$ 0.032 & 0.722 $\pm$ 0.093 \\
\bottomrule
\end{tabular}
}
\end{table}

\begin{table}[h!]
\centering
\footnotesize
\renewcommand{\arraystretch}{0.88}
\setlength{\tabcolsep}{4pt}
\caption{Inpainting results by dataset (PSNR $\uparrow$) . Mean $\pm$ standard deviation.}
\label{tab:inpainting_psnr_p1}
\resizebox{\textwidth}{!}{%
\begin{tabular}{lcccccccc}
\toprule
\textbf{Model} & \textbf{AP-10K} & \textbf{ATLANTIS} & \textbf{Caltech Birds} & \textbf{CDD} & \textbf{DeepCrack} & \textbf{DRAM} & \textbf{Duke} & \textbf{IXI (Axial)} \\
\midrule
DELVM 1B & 15.336 $\pm$ 3.290 & 14.630 $\pm$ 3.379 & 14.708 $\pm$ 4.140 & 12.364 $\pm$ 3.590 & 15.617 $\pm$ 2.047 & 16.814 $\pm$ 2.411 & 20.654 $\pm$ 2.885 & 16.690 $\pm$ 2.034 \\
LVM & \textbf{18.930 $\pm$ 2.882} & \textbf{19.428 $\pm$ 3.182} & \textbf{20.490 $\pm$ 3.060} & \textbf{18.099 $\pm$ 2.871} & \textbf{22.372 $\pm$ 3.947} & \textbf{19.643 $\pm$ 2.300} & \textbf{23.144 $\pm$ 1.167} & 18.980 $\pm$ 1.290 \\
MAE-VQGAN & 15.184 $\pm$ 1.966 & 15.465 $\pm$ 2.176 & 15.770 $\pm$ 2.217 & 14.634 $\pm$ 1.976 & 17.218 $\pm$ 2.662 & 15.033 $\pm$ 1.916 & 17.957 $\pm$ 1.158 & 15.771 $\pm$ 1.097 \\
Painter & 11.883 $\pm$ 5.117 & 12.164 $\pm$ 4.561 & 10.004 $\pm$ 4.631 & 10.505 $\pm$ 3.222 & 16.932 $\pm$ 1.609 & 18.359 $\pm$ 2.921 & 14.355 $\pm$ 1.327 & 14.048 $\pm$ 1.150 \\
PromptDiff. & 11.442 $\pm$ 1.802 & 10.197 $\pm$ 1.680 & 10.553 $\pm$ 2.041 & 9.907 $\pm$ 1.466 & 12.147 $\pm$ 3.218 & 9.754 $\pm$ 1.393 & 7.728 $\pm$ 0.492 & 7.498 $\pm$ 0.371 \\
SD-VICL & 13.059 $\pm$ 2.277 & 12.478 $\pm$ 2.363 & 12.549 $\pm$ 3.006 & 13.224 $\pm$ 2.312 & 14.622 $\pm$ 3.109 & 12.419 $\pm$ 2.795 & 20.721 $\pm$ 1.970 & \textbf{19.919 $\pm$ 1.696} \\
\midrule
Copy target & 9.509 $\pm$ 1.575 & 9.117 $\pm$ 1.627 & 9.958 $\pm$ 2.123 & 9.378 $\pm$ 1.630 & 14.206 $\pm$ 2.747 & 9.347 $\pm$ 1.761 & 12.301 $\pm$ 2.110 & 13.417 $\pm$ 1.120 \\
\bottomrule
\end{tabular}
}
\end{table}

\begin{table}[h!]
\centering
\footnotesize
\renewcommand{\arraystretch}{0.88}
\setlength{\tabcolsep}{4pt}
\caption{Inpainting results by dataset (PSNR $\uparrow$) . Mean $\pm$ standard deviation.}
\label{tab:inpainting_psnr_p2}
\resizebox{\textwidth}{!}{%
\begin{tabular}{lcccccccc}
\toprule
\textbf{Model} & \textbf{IXI (Coronal)} & \textbf{IXI (Sagittal)} & \textbf{LLVIP} & \textbf{NYU-D V2} & \textbf{OTP} & \textbf{RORD} & \textbf{Sea-Thru} & \textbf{SUIM} \\
\midrule
DELVM 1B & 16.122 $\pm$ 2.117 & 17.646 $\pm$ 1.791 & 18.958 $\pm$ 2.505 & 17.275 $\pm$ 2.649 & 18.955 $\pm$ 3.445 & 15.357 $\pm$ 3.064 & 27.992 $\pm$ 4.916 & 16.748 $\pm$ 3.404 \\
LVM & 18.716 $\pm$ 1.061 & 19.937 $\pm$ 0.988 & \textbf{21.292 $\pm$ 2.868} & \textbf{19.785 $\pm$ 2.579} & \textbf{23.601 $\pm$ 4.427} & \textbf{20.781 $\pm$ 4.259} & \textbf{31.000 $\pm$ 2.564} & \textbf{21.547 $\pm$ 4.429} \\
MAE-VQGAN & 14.628 $\pm$ 0.714 & 15.611 $\pm$ 0.903 & 16.962 $\pm$ 1.805 & 13.752 $\pm$ 1.386 & 17.439 $\pm$ 2.307 & 15.494 $\pm$ 2.147 & 28.766 $\pm$ 3.018 & 16.878 $\pm$ 2.665 \\
Painter & 17.278 $\pm$ 1.516 & 15.836 $\pm$ 1.293 & 14.509 $\pm$ 2.673 & 9.141 $\pm$ 1.559 & 14.566 $\pm$ 4.141 & 14.827 $\pm$ 3.937 & 9.161 $\pm$ 2.947 & 13.745 $\pm$ 5.656 \\
PromptDiff. & 8.421 $\pm$ 0.368 & 9.414 $\pm$ 0.241 & 9.538 $\pm$ 1.291 & 9.428 $\pm$ 1.561 & 12.682 $\pm$ 1.881 & 10.701 $\pm$ 1.645 & 10.022 $\pm$ 0.662 & 9.986 $\pm$ 1.520 \\
SD-VICL & \textbf{20.208 $\pm$ 1.722} & \textbf{21.270 $\pm$ 1.743} & 18.193 $\pm$ 2.769 & 15.068 $\pm$ 3.061 & 17.279 $\pm$ 5.678 & 12.948 $\pm$ 2.643 & 27.177 $\pm$ 3.039 & 12.363 $\pm$ 2.304 \\
\midrule
Copy target & 12.199 $\pm$ 0.674 & 13.692 $\pm$ 0.825 & 12.150 $\pm$ 2.102 & 8.930 $\pm$ 1.420 & 13.224 $\pm$ 4.048 & 9.576 $\pm$ 1.834 & 25.947 $\pm$ 2.798 & 10.467 $\pm$ 1.964 \\
\bottomrule
\end{tabular}
}
\end{table}

\begin{table}[h]
\centering
\footnotesize
\renewcommand{\arraystretch}{0.88}
\setlength{\tabcolsep}{4pt}
\caption{Inpainting results by dataset (SSIM $\uparrow$) . Mean $\pm$ standard deviation.}
\label{tab:inpainting_ssim_p1}
\resizebox{\textwidth}{!}{%
\begin{tabular}{lcccccccc}
\toprule
\textbf{Model} & \textbf{AP-10K} & \textbf{ATLANTIS} & \textbf{Caltech Birds} & \textbf{CDD} & \textbf{DeepCrack} & \textbf{DRAM} & \textbf{Duke} & \textbf{IXI (Axial)} \\
\midrule
DELVM 1B & 0.377 $\pm$ 0.172 & 0.438 $\pm$ 0.182 & 0.528 $\pm$ 0.231 & 0.372 $\pm$ 0.183 & 0.353 $\pm$ 0.192 & 0.433 $\pm$ 0.123 & 0.401 $\pm$ 0.067 & 0.579 $\pm$ 0.089 \\
LVM & 0.396 $\pm$ 0.169 & \textbf{0.466 $\pm$ 0.178} & \textbf{0.607 $\pm$ 0.182} & \textbf{0.417 $\pm$ 0.163} & 0.368 $\pm$ 0.205 & 0.430 $\pm$ 0.128 & 0.423 $\pm$ 0.066 & 0.586 $\pm$ 0.067 \\
MAE-VQGAN & 0.328 $\pm$ 0.154 & 0.387 $\pm$ 0.173 & 0.469 $\pm$ 0.196 & 0.379 $\pm$ 0.161 & 0.325 $\pm$ 0.196 & 0.321 $\pm$ 0.122 & 0.381 $\pm$ 0.067 & 0.462 $\pm$ 0.044 \\
Painter & \textbf{0.399 $\pm$ 0.301} & 0.459 $\pm$ 0.272 & 0.336 $\pm$ 0.328 & 0.380 $\pm$ 0.196 & \textbf{0.797 $\pm$ 0.118} & \textbf{0.836 $\pm$ 0.083} & \textbf{0.516 $\pm$ 0.051} & 0.413 $\pm$ 0.079 \\
PromptDiff. & 0.266 $\pm$ 0.120 & 0.282 $\pm$ 0.125 & 0.346 $\pm$ 0.139 & 0.289 $\pm$ 0.116 & 0.236 $\pm$ 0.139 & 0.279 $\pm$ 0.095 & 0.187 $\pm$ 0.028 & 0.042 $\pm$ 0.019 \\
SD-VICL & 0.360 $\pm$ 0.109 & 0.420 $\pm$ 0.120 & 0.532 $\pm$ 0.125 & 0.405 $\pm$ 0.127 & 0.379 $\pm$ 0.130 & 0.421 $\pm$ 0.097 & 0.476 $\pm$ 0.050 & \textbf{0.808 $\pm$ 0.039} \\
\midrule
Copy target & 0.162 $\pm$ 0.082 & 0.200 $\pm$ 0.112 & 0.266 $\pm$ 0.136 & 0.246 $\pm$ 0.129 & 0.191 $\pm$ 0.120 & 0.160 $\pm$ 0.064 & 0.269 $\pm$ 0.055 & 0.429 $\pm$ 0.052 \\
\bottomrule
\end{tabular}
}
\end{table}

\begin{table}[h]
\centering
\footnotesize
\renewcommand{\arraystretch}{0.88}
\setlength{\tabcolsep}{4pt}
\caption{Inpainting results by dataset (SSIM $\uparrow$) . Mean $\pm$ standard deviation.}
\label{tab:inpainting_ssim_p2}
\resizebox{\textwidth}{!}{%
\begin{tabular}{lcccccccc}
\toprule
\textbf{Model} & \textbf{IXI (Coronal)} & \textbf{IXI (Sagittal)} & \textbf{LLVIP} & \textbf{NYU-D V2} & \textbf{OTP} & \textbf{RORD} & \textbf{Sea-Thru} & \textbf{SUIM} \\
\midrule
DELVM 1B & 0.552 $\pm$ 0.134 & 0.608 $\pm$ 0.089 & 0.660 $\pm$ 0.136 & 0.604 $\pm$ 0.085 & 0.813 $\pm$ 0.135 & 0.454 $\pm$ 0.235 & 0.631 $\pm$ 0.187 & 0.518 $\pm$ 0.197 \\
LVM & 0.591 $\pm$ 0.048 & 0.623 $\pm$ 0.045 & 0.684 $\pm$ 0.140 & 0.604 $\pm$ 0.099 & \textbf{0.850 $\pm$ 0.122} & 0.483 $\pm$ 0.239 & \textbf{0.741 $\pm$ 0.077} & \textbf{0.546 $\pm$ 0.203} \\
MAE-VQGAN & 0.313 $\pm$ 0.044 & 0.393 $\pm$ 0.035 & 0.594 $\pm$ 0.133 & 0.446 $\pm$ 0.097 & 0.738 $\pm$ 0.168 & 0.396 $\pm$ 0.220 & 0.705 $\pm$ 0.095 & 0.476 $\pm$ 0.202 \\
Painter & 0.696 $\pm$ 0.056 & 0.550 $\pm$ 0.058 & \textbf{0.715 $\pm$ 0.125} & 0.382 $\pm$ 0.084 & 0.689 $\pm$ 0.262 & \textbf{0.582 $\pm$ 0.233} & 0.212 $\pm$ 0.070 & 0.511 $\pm$ 0.348 \\
PromptDiff. & 0.061 $\pm$ 0.015 & 0.067 $\pm$ 0.013 & 0.510 $\pm$ 0.104 & 0.401 $\pm$ 0.104 & 0.333 $\pm$ 0.075 & 0.323 $\pm$ 0.165 & 0.210 $\pm$ 0.040 & 0.306 $\pm$ 0.119 \\
SD-VICL & \textbf{0.845 $\pm$ 0.028} & \textbf{0.820 $\pm$ 0.037} & 0.699 $\pm$ 0.114 & \textbf{0.618 $\pm$ 0.127} & 0.743 $\pm$ 0.213 & 0.390 $\pm$ 0.159 & 0.536 $\pm$ 0.113 & 0.426 $\pm$ 0.116 \\
\midrule
Copy target & 0.199 $\pm$ 0.042 & 0.342 $\pm$ 0.039 & 0.427 $\pm$ 0.134 & 0.270 $\pm$ 0.068 & 0.647 $\pm$ 0.240 & 0.221 $\pm$ 0.139 & 0.686 $\pm$ 0.088 & 0.270 $\pm$ 0.143 \\
\bottomrule
\end{tabular}
}
\end{table}

\begin{table}[h]
\centering
\footnotesize
\renewcommand{\arraystretch}{0.88}
\setlength{\tabcolsep}{4pt}
\caption{Inpainting results by dataset (LPIPS $\downarrow$) . Mean $\pm$ standard deviation.}
\label{tab:inpainting_lpips_p1}
\resizebox{\textwidth}{!}{%
\begin{tabular}{lcccccccc}
\toprule
\textbf{Model} & \textbf{AP-10K} & \textbf{ATLANTIS} & \textbf{Caltech Birds} & \textbf{CDD} & \textbf{DeepCrack} & \textbf{DRAM} & \textbf{Duke} & \textbf{IXI (Axial)} \\
\midrule
DELVM 1B & 0.560 $\pm$ 0.087 & 0.490 $\pm$ 0.115 & 0.470 $\pm$ 0.160 & 0.588 $\pm$ 0.125 & 0.503 $\pm$ 0.074 & 0.468 $\pm$ 0.049 & 0.526 $\pm$ 0.031 & 0.276 $\pm$ 0.069 \\
LVM & \textbf{0.543 $\pm$ 0.076} & \textbf{0.451 $\pm$ 0.088} & \textbf{0.387 $\pm$ 0.087} & \textbf{0.578 $\pm$ 0.073} & 0.480 $\pm$ 0.057 & 0.467 $\pm$ 0.048 & 0.515 $\pm$ 0.024 & 0.245 $\pm$ 0.027 \\
MAE-VQGAN & 0.693 $\pm$ 0.063 & 0.644 $\pm$ 0.079 & 0.585 $\pm$ 0.080 & 0.709 $\pm$ 0.078 & 0.633 $\pm$ 0.062 & 0.643 $\pm$ 0.047 & 0.625 $\pm$ 0.022 & 0.371 $\pm$ 0.033 \\
Painter & 0.588 $\pm$ 0.213 & 0.541 $\pm$ 0.190 & 0.621 $\pm$ 0.208 & 0.756 $\pm$ 0.137 & \textbf{0.284 $\pm$ 0.099} & \textbf{0.213 $\pm$ 0.082} & 0.430 $\pm$ 0.020 & 0.248 $\pm$ 0.031 \\
PromptDiff. & 0.840 $\pm$ 0.069 & 0.739 $\pm$ 0.072 & 0.772 $\pm$ 0.107 & 0.844 $\pm$ 0.078 & 0.719 $\pm$ 0.057 & 0.825 $\pm$ 0.076 & 0.703 $\pm$ 0.028 & 0.655 $\pm$ 0.034 \\
SD-VICL & 0.601 $\pm$ 0.081 & 0.544 $\pm$ 0.080 & 0.562 $\pm$ 0.121 & 0.590 $\pm$ 0.095 & 0.541 $\pm$ 0.080 & 0.528 $\pm$ 0.074 & \textbf{0.399 $\pm$ 0.023} & \textbf{0.194 $\pm$ 0.023} \\
\midrule
Copy target & 0.770 $\pm$ 0.059 & 0.740 $\pm$ 0.057 & 0.789 $\pm$ 0.084 & 0.747 $\pm$ 0.082 & 0.635 $\pm$ 0.057 & 0.766 $\pm$ 0.057 & 0.550 $\pm$ 0.050 & 0.370 $\pm$ 0.039 \\
\bottomrule
\end{tabular}
}
\end{table}

\begin{table}[h]
\centering
\footnotesize
\renewcommand{\arraystretch}{0.88}
\setlength{\tabcolsep}{4pt}
\caption{Inpainting results by dataset (LPIPS $\downarrow$) . Mean $\pm$ standard deviation.}
\label{tab:inpainting_lpips_p2}
\resizebox{\textwidth}{!}{%
\begin{tabular}{lcccccccc}
\toprule
\textbf{Model} & \textbf{IXI (Coronal)} & \textbf{IXI (Sagittal)} & \textbf{LLVIP} & \textbf{NYU-D V2} & \textbf{OTP} & \textbf{RORD} & \textbf{Sea-Thru} & \textbf{SUIM} \\
\midrule
DELVM 1B & 0.300 $\pm$ 0.082 & 0.293 $\pm$ 0.058 & 0.455 $\pm$ 0.058 & \textbf{0.380 $\pm$ 0.039} & 0.290 $\pm$ 0.084 & 0.538 $\pm$ 0.117 & 0.587 $\pm$ 0.067 & 0.443 $\pm$ 0.100 \\
LVM & 0.257 $\pm$ 0.024 & 0.258 $\pm$ 0.020 & 0.479 $\pm$ 0.060 & 0.385 $\pm$ 0.053 & \textbf{0.261 $\pm$ 0.086} & 0.511 $\pm$ 0.089 & 0.563 $\pm$ 0.054 & \textbf{0.420 $\pm$ 0.097} \\
MAE-VQGAN & 0.411 $\pm$ 0.024 & 0.410 $\pm$ 0.024 & 0.595 $\pm$ 0.060 & 0.589 $\pm$ 0.043 & 0.387 $\pm$ 0.121 & 0.679 $\pm$ 0.093 & 0.614 $\pm$ 0.068 & 0.566 $\pm$ 0.105 \\
Painter & \textbf{0.150 $\pm$ 0.022} & 0.197 $\pm$ 0.018 & \textbf{0.327 $\pm$ 0.062} & 0.733 $\pm$ 0.066 & 0.313 $\pm$ 0.171 & \textbf{0.483 $\pm$ 0.151} & 0.521 $\pm$ 0.066 & 0.457 $\pm$ 0.234 \\
PromptDiff. & 0.701 $\pm$ 0.022 & 0.746 $\pm$ 0.032 & 0.600 $\pm$ 0.041 & 0.711 $\pm$ 0.073 & 0.784 $\pm$ 0.041 & 0.742 $\pm$ 0.095 & 0.831 $\pm$ 0.048 & 0.809 $\pm$ 0.065 \\
SD-VICL & 0.157 $\pm$ 0.024 & \textbf{0.186 $\pm$ 0.026} & 0.350 $\pm$ 0.051 & 0.401 $\pm$ 0.062 & 0.328 $\pm$ 0.142 & 0.609 $\pm$ 0.096 & \textbf{0.470 $\pm$ 0.060} & 0.545 $\pm$ 0.109 \\
\midrule
Copy target & 0.416 $\pm$ 0.038 & 0.372 $\pm$ 0.032 & 0.615 $\pm$ 0.093 & 0.676 $\pm$ 0.043 & 0.441 $\pm$ 0.194 & 0.773 $\pm$ 0.071 & 0.551 $\pm$ 0.032 & 0.722 $\pm$ 0.093 \\
\bottomrule
\end{tabular}
}
\end{table}

\begin{table}[h]
\centering
\footnotesize
\renewcommand{\arraystretch}{0.88}
\setlength{\tabcolsep}{4pt}
\caption{Colorization results by dataset (PSNR $\uparrow$) . Mean $\pm$ standard deviation.}
\label{tab:colorization_psnr_p1}
\resizebox{\textwidth}{!}{%
\begin{tabular}{lcccccccc}
\toprule
\textbf{Model} & \textbf{AP-10K} & \textbf{ATLANTIS} & \textbf{Caltech Birds} & \textbf{CDD} & \textbf{DeepCrack} & \textbf{DRAM} & \textbf{LLVIP} & \textbf{NYU-D V2} \\
\midrule
DELVM 1B & 17.659 $\pm$ 3.672 & 17.039 $\pm$ 4.410 & 15.268 $\pm$ 5.966 & 14.450 $\pm$ 4.369 & 19.847 $\pm$ 6.660 & 17.669 $\pm$ 2.567 & 19.026 $\pm$ 4.037 & \textbf{19.684 $\pm$ 2.208} \\
LVM & 17.682 $\pm$ 2.546 & 17.903 $\pm$ 2.812 & \textbf{18.218 $\pm$ 2.931} & \textbf{16.832 $\pm$ 2.625} & 20.312 $\pm$ 3.902 & 17.372 $\pm$ 2.409 & \textbf{19.145 $\pm$ 3.908} & 18.172 $\pm$ 2.402 \\
MAE-VQGAN & 14.814 $\pm$ 1.810 & 14.708 $\pm$ 1.837 & 14.979 $\pm$ 2.031 & 13.758 $\pm$ 1.674 & 17.746 $\pm$ 2.143 & 14.383 $\pm$ 1.676 & 17.746 $\pm$ 3.823 & 13.859 $\pm$ 1.257 \\
Painter & \textbf{18.510 $\pm$ 5.418} & \textbf{18.580 $\pm$ 5.222} & 14.678 $\pm$ 6.234 & 15.510 $\pm$ 3.984 & \textbf{28.247 $\pm$ 4.721} & \textbf{19.681 $\pm$ 3.377} & 12.515 $\pm$ 2.510 & 10.059 $\pm$ 2.503 \\
PromptDiff. & 11.646 $\pm$ 1.882 & 10.664 $\pm$ 1.839 & 10.776 $\pm$ 2.139 & 10.045 $\pm$ 1.504 & 11.836 $\pm$ 3.280 & 10.060 $\pm$ 1.496 & 8.023 $\pm$ 1.332 & 9.658 $\pm$ 1.620 \\
SD-VICL & 13.409 $\pm$ 2.422 & 12.780 $\pm$ 2.495 & 12.690 $\pm$ 3.080 & 13.449 $\pm$ 2.447 & 15.661 $\pm$ 3.451 & 12.510 $\pm$ 2.912 & 15.797 $\pm$ 4.405 & 15.547 $\pm$ 3.240 \\
\midrule
Copy target & 9.509 $\pm$ 1.575 & 9.117 $\pm$ 1.627 & 9.958 $\pm$ 2.123 & 9.378 $\pm$ 1.630 & 14.206 $\pm$ 2.747 & 9.347 $\pm$ 1.761 & 13.397 $\pm$ 4.185 & 8.930 $\pm$ 1.420 \\
\bottomrule
\end{tabular}
}
\end{table}

\begin{table}[h]
\centering
\footnotesize
\renewcommand{\arraystretch}{0.88}
\setlength{\tabcolsep}{4pt}
\caption{Colorization results by dataset (PSNR $\uparrow$) . Mean $\pm$ standard deviation.}
\label{tab:colorization_psnr_p2}
\begin{tabular}{lccc}
\toprule
\textbf{Model} & \textbf{RORD} & \textbf{Sea-Thru} & \textbf{SUIM} \\
\midrule
DELVM 1B & 18.272 $\pm$ 4.866 & 25.669 $\pm$ 4.102 & 14.130 $\pm$ 3.608 \\
LVM & 18.189 $\pm$ 3.184 & 26.776 $\pm$ 3.091 & \textbf{16.112 $\pm$ 3.028} \\
MAE-VQGAN & 15.060 $\pm$ 1.937 & \textbf{28.418 $\pm$ 2.745} & 14.078 $\pm$ 2.164 \\
Painter & \textbf{19.363 $\pm$ 4.847} & 9.873 $\pm$ 2.469 & 14.509 $\pm$ 4.537 \\
PromptDiff. & 10.946 $\pm$ 1.703 & 10.876 $\pm$ 0.500 & 10.227 $\pm$ 1.528 \\
SD-VICL & 13.293 $\pm$ 2.820 & 27.495 $\pm$ 3.116 & 12.369 $\pm$ 2.382 \\
\midrule
Copy target & 9.576 $\pm$ 1.834 & 25.947 $\pm$ 2.798 & 10.467 $\pm$ 1.964 \\
\bottomrule
\end{tabular}
\end{table}

\begin{table}[h]
\centering
\footnotesize
\renewcommand{\arraystretch}{0.88}
\setlength{\tabcolsep}{4pt}
\caption{Colorization results by dataset (SSIM $\uparrow$) . Mean $\pm$ standard deviation.}
\label{tab:colorization_ssim_p1}
\resizebox{\textwidth}{!}{%
\begin{tabular}{lcccccccc}
\toprule
\textbf{Model} & \textbf{AP-10K} & \textbf{ATLANTIS} & \textbf{Caltech Birds} & \textbf{CDD} & \textbf{DeepCrack} & \textbf{DRAM} & \textbf{LLVIP} & \textbf{NYU-D V2} \\
\midrule
DELVM 1B & 0.395 $\pm$ 0.178 & 0.450 $\pm$ 0.190 & 0.490 $\pm$ 0.269 & 0.383 $\pm$ 0.185 & 0.339 $\pm$ 0.213 & 0.449 $\pm$ 0.138 & 0.510 $\pm$ 0.200 & 0.633 $\pm$ 0.094 \\
LVM & 0.382 $\pm$ 0.163 & 0.453 $\pm$ 0.175 & 0.583 $\pm$ 0.179 & 0.409 $\pm$ 0.167 & 0.362 $\pm$ 0.201 & 0.410 $\pm$ 0.126 & \textbf{0.522 $\pm$ 0.181} & 0.578 $\pm$ 0.100 \\
MAE-VQGAN & 0.322 $\pm$ 0.149 & 0.376 $\pm$ 0.166 & 0.463 $\pm$ 0.190 & 0.372 $\pm$ 0.159 & 0.329 $\pm$ 0.195 & 0.308 $\pm$ 0.118 & 0.498 $\pm$ 0.170 & 0.444 $\pm$ 0.096 \\
Painter & \textbf{0.662 $\pm$ 0.217} & \textbf{0.703 $\pm$ 0.189} & \textbf{0.598 $\pm$ 0.319} & \textbf{0.546 $\pm$ 0.171} & \textbf{0.855 $\pm$ 0.113} & \textbf{0.840 $\pm$ 0.078} & 0.411 $\pm$ 0.174 & 0.451 $\pm$ 0.120 \\
PromptDiff. & 0.262 $\pm$ 0.116 & 0.259 $\pm$ 0.101 & 0.338 $\pm$ 0.144 & 0.265 $\pm$ 0.103 & 0.212 $\pm$ 0.124 & 0.277 $\pm$ 0.093 & 0.237 $\pm$ 0.074 & 0.405 $\pm$ 0.107 \\
SD-VICL & 0.385 $\pm$ 0.114 & 0.445 $\pm$ 0.123 & 0.562 $\pm$ 0.128 & 0.421 $\pm$ 0.124 & 0.450 $\pm$ 0.152 & 0.449 $\pm$ 0.102 & 0.399 $\pm$ 0.173 & \textbf{0.635 $\pm$ 0.132} \\
\midrule
Copy target & 0.162 $\pm$ 0.082 & 0.200 $\pm$ 0.112 & 0.266 $\pm$ 0.136 & 0.246 $\pm$ 0.129 & 0.191 $\pm$ 0.120 & 0.160 $\pm$ 0.064 & 0.348 $\pm$ 0.181 & 0.270 $\pm$ 0.068 \\
\bottomrule
\end{tabular}
}
\end{table}

\begin{table}[h]
\centering
\footnotesize
\renewcommand{\arraystretch}{0.88}
\setlength{\tabcolsep}{4pt}
\caption{Colorization results by dataset (SSIM $\uparrow$). Mean $\pm$ standard deviation.}
\label{tab:colorization_ssim_p2}
\begin{tabular}{lccc}
\toprule
\textbf{Model} & \textbf{RORD} & \textbf{Sea-Thru} & \textbf{SUIM} \\
\midrule
DELVM 1B & 0.474 $\pm$ 0.248 & 0.532 $\pm$ 0.192 & 0.454 $\pm$ 0.196 \\
LVM & 0.463 $\pm$ 0.233 & 0.630 $\pm$ 0.087 & 0.473 $\pm$ 0.179 \\
MAE-VQGAN & 0.392 $\pm$ 0.217 & \textbf{0.698 $\pm$ 0.084} & 0.419 $\pm$ 0.178 \\
Painter & \textbf{0.671 $\pm$ 0.190} & 0.238 $\pm$ 0.076 & \textbf{0.657 $\pm$ 0.264} \\
PromptDiff. & 0.317 $\pm$ 0.163 & 0.217 $\pm$ 0.040 & 0.289 $\pm$ 0.117 \\
SD-VICL & 0.412 $\pm$ 0.165 & 0.542 $\pm$ 0.114 & 0.447 $\pm$ 0.121 \\
\midrule
Copy target & 0.221 $\pm$ 0.139 & 0.686 $\pm$ 0.088 & 0.270 $\pm$ 0.143 \\
\bottomrule
\end{tabular}
\end{table}

\begin{table}[h]
\centering
\footnotesize
\renewcommand{\arraystretch}{0.88}
\setlength{\tabcolsep}{4pt}
\caption{Colorization results by dataset (LPIPS $\downarrow$) . Mean $\pm$ standard deviation.}
\label{tab:colorization_lpips_p1}
\resizebox{\textwidth}{!}{%
\begin{tabular}{lcccccccc}
\toprule
\textbf{Model} & \textbf{AP-10K} & \textbf{ATLANTIS} & \textbf{Caltech Birds} & \textbf{CDD} & \textbf{DeepCrack} & \textbf{DRAM} & \textbf{LLVIP} & \textbf{NYU-D V2} \\
\midrule
DELVM 1B & 0.655 $\pm$ 0.109 & 0.567 $\pm$ 0.119 & 0.629 $\pm$ 0.170 & 0.685 $\pm$ 0.107 & 0.539 $\pm$ 0.131 & 0.585 $\pm$ 0.080 & 0.654 $\pm$ 0.084 & 0.461 $\pm$ 0.070 \\
LVM & 0.627 $\pm$ 0.088 & 0.529 $\pm$ 0.096 & \textbf{0.517 $\pm$ 0.122} & 0.643 $\pm$ 0.084 & 0.550 $\pm$ 0.103 & 0.599 $\pm$ 0.077 & 0.655 $\pm$ 0.088 & 0.512 $\pm$ 0.082 \\
MAE-VQGAN & 0.745 $\pm$ 0.072 & 0.696 $\pm$ 0.073 & 0.686 $\pm$ 0.100 & 0.772 $\pm$ 0.089 & 0.639 $\pm$ 0.062 & 0.709 $\pm$ 0.061 & 0.690 $\pm$ 0.096 & 0.634 $\pm$ 0.054 \\
Painter & \textbf{0.533 $\pm$ 0.157} & \textbf{0.451 $\pm$ 0.145} & 0.587 $\pm$ 0.191 & 0.691 $\pm$ 0.142 & \textbf{0.276 $\pm$ 0.093} & \textbf{0.388 $\pm$ 0.098} & 0.598 $\pm$ 0.092 & 0.656 $\pm$ 0.096 \\
PromptDiff. & 0.826 $\pm$ 0.074 & 0.791 $\pm$ 0.061 & 0.760 $\pm$ 0.111 & 0.819 $\pm$ 0.076 & 0.691 $\pm$ 0.049 & 0.783 $\pm$ 0.074 & 0.816 $\pm$ 0.074 & 0.715 $\pm$ 0.078 \\
SD-VICL & 0.627 $\pm$ 0.083 & 0.553 $\pm$ 0.083 & 0.582 $\pm$ 0.119 & \textbf{0.609 $\pm$ 0.099} & 0.492 $\pm$ 0.077 & 0.565 $\pm$ 0.078 & \textbf{0.518 $\pm$ 0.095} & \textbf{0.418 $\pm$ 0.067} \\
\midrule
Copy target & 0.770 $\pm$ 0.059 & 0.740 $\pm$ 0.057 & 0.789 $\pm$ 0.084 & 0.747 $\pm$ 0.082 & 0.635 $\pm$ 0.057 & 0.766 $\pm$ 0.057 & 0.684 $\pm$ 0.115 & 0.676 $\pm$ 0.043 \\
\bottomrule
\end{tabular}
}
\end{table}

\begin{table}[h]
\centering
\footnotesize
\renewcommand{\arraystretch}{0.88}
\setlength{\tabcolsep}{4pt}
\caption{Colorization results by dataset (LPIPS $\downarrow$) . Mean $\pm$ standard deviation.}
\label{tab:colorization_lpips_p2}
\begin{tabular}{lccc}
\toprule
\textbf{Model} & \textbf{RORD} & \textbf{Sea-Thru} & \textbf{SUIM} \\
\midrule
DELVM 1B & 0.608 $\pm$ 0.127 & 0.687 $\pm$ 0.088 & 0.682 $\pm$ 0.097 \\
LVM & 0.616 $\pm$ 0.091 & 0.706 $\pm$ 0.108 & 0.568 $\pm$ 0.133 \\
MAE-VQGAN & 0.729 $\pm$ 0.095 & 0.618 $\pm$ 0.069 & 0.679 $\pm$ 0.106 \\
Painter & \textbf{0.502 $\pm$ 0.124} & 0.664 $\pm$ 0.081 & \textbf{0.544 $\pm$ 0.134} \\
PromptDiff. & 0.722 $\pm$ 0.091 & 0.804 $\pm$ 0.037 & 0.785 $\pm$ 0.064 \\
SD-VICL & 0.622 $\pm$ 0.102 & \textbf{0.459 $\pm$ 0.061} & 0.551 $\pm$ 0.113 \\
\midrule
Copy target & 0.773 $\pm$ 0.071 & 0.551 $\pm$ 0.032 & 0.722 $\pm$ 0.093 \\
\bottomrule
\end{tabular}
\end{table}

\clearpage

\subsubsection{Image Manipulation \& Transformation}
\begin{table}[h!]
\centering
\footnotesize
\renewcommand{\arraystretch}{0.88}
\setlength{\tabcolsep}{4pt}
\caption{Style Transfer results by dataset (PSNR $\uparrow$). Mean $\pm$ standard deviation.}
\label{tab:style_transfer_psnr}
\begin{tabular}{lcccc}
\toprule
\textbf{Model} & \textbf{IXI (Axial)} & \textbf{IXI (Coronal)} & \textbf{IXI (Sagittal)} & \textbf{LLVIP} \\
\midrule
DELVM 1B & 13.425 $\pm$ 2.515 & 11.043 $\pm$ 2.033 & 11.937 $\pm$ 2.112 & 10.389 $\pm$ 2.147 \\
LVM & 14.060 $\pm$ 2.079 & 11.741 $\pm$ 1.559 & 12.648 $\pm$ 1.638 & 10.817 $\pm$ 2.020 \\
MAE-VQGAN & 14.897 $\pm$ 1.164 & 13.577 $\pm$ 0.696 & 14.598 $\pm$ 0.875 & 10.177 $\pm$ 2.299 \\
Painter & 19.791 $\pm$ 1.552 & 14.008 $\pm$ 1.676 & 16.153 $\pm$ 1.428 & 11.518 $\pm$ 1.105 \\
PromptDiff. & 8.268 $\pm$ 0.471 & 9.869 $\pm$ 0.385 & 9.175 $\pm$ 0.248 & 9.828 $\pm$ 1.146 \\
SD-VICL & \textbf{21.038 $\pm$ 2.316} & \textbf{19.262 $\pm$ 2.033} & \textbf{20.495 $\pm$ 1.941} & \textbf{11.976 $\pm$ 2.301} \\
\midrule
Copy target & 12.446 $\pm$ 1.188 & 11.010 $\pm$ 0.725 & 12.721 $\pm$ 0.975 & 12.150 $\pm$ 2.102 \\
\bottomrule
\end{tabular}
\end{table}

\begin{table}[h]
\centering
\footnotesize
\renewcommand{\arraystretch}{0.88}
\setlength{\tabcolsep}{4pt}
\caption{Style Transfer results by dataset (SSIM $\uparrow$). Mean $\pm$ standard deviation.}
\label{tab:style_transfer_ssim}
\begin{tabular}{lcccc}
\toprule
\textbf{Model} & \textbf{IXI (Axial)} & \textbf{IXI (Coronal)} & \textbf{IXI (Sagittal)} & \textbf{LLVIP} \\
\midrule
DELVM 1B & 0.511 $\pm$ 0.101 & 0.386 $\pm$ 0.136 & 0.476 $\pm$ 0.119 & 0.298 $\pm$ 0.162 \\
LVM & 0.473 $\pm$ 0.099 & 0.401 $\pm$ 0.105 & 0.481 $\pm$ 0.085 & 0.352 $\pm$ 0.124 \\
MAE-VQGAN & 0.485 $\pm$ 0.053 & 0.332 $\pm$ 0.045 & 0.411 $\pm$ 0.038 & 0.274 $\pm$ 0.164 \\
Painter & 0.531 $\pm$ 0.080 & 0.617 $\pm$ 0.028 & 0.580 $\pm$ 0.027 & 0.401 $\pm$ 0.102 \\
PromptDiff. & 0.058 $\pm$ 0.025 & 0.098 $\pm$ 0.025 & 0.071 $\pm$ 0.011 & \textbf{0.483 $\pm$ 0.103} \\
SD-VICL & \textbf{0.792 $\pm$ 0.053} & \textbf{0.801 $\pm$ 0.048} & \textbf{0.778 $\pm$ 0.048} & 0.425 $\pm$ 0.127 \\
\midrule
Copy target & 0.436 $\pm$ 0.054 & 0.207 $\pm$ 0.043 & 0.345 $\pm$ 0.043 & 0.427 $\pm$ 0.134 \\
\bottomrule
\end{tabular}
\end{table}

\begin{table}[h]
\centering
\footnotesize
\renewcommand{\arraystretch}{0.88}
\setlength{\tabcolsep}{4pt}
\caption{Style Transfer results by dataset (LPIPS $\downarrow$). Mean $\pm$ standard deviation.}
\label{tab:style_transfer_lpips}
\begin{tabular}{lcccc}
\toprule
\textbf{Model} & \textbf{IXI (Axial)} & \textbf{IXI (Coronal)} & \textbf{IXI (Sagittal)} & \textbf{LLVIP} \\
\midrule
DELVM 1B & 0.293 $\pm$ 0.079 & 0.350 $\pm$ 0.119 & 0.336 $\pm$ 0.103 & 0.728 $\pm$ 0.081 \\
LVM & 0.303 $\pm$ 0.051 & 0.321 $\pm$ 0.043 & 0.317 $\pm$ 0.047 & 0.737 $\pm$ 0.089 \\
MAE-VQGAN & 0.354 $\pm$ 0.034 & 0.403 $\pm$ 0.026 & 0.400 $\pm$ 0.023 & 0.692 $\pm$ 0.048 \\
Painter & 0.237 $\pm$ 0.029 & 0.193 $\pm$ 0.020 & \textbf{0.211 $\pm$ 0.015} & 0.801 $\pm$ 0.089 \\
PromptDiff. & 0.661 $\pm$ 0.028 & 0.689 $\pm$ 0.028 & 0.743 $\pm$ 0.020 & 0.604 $\pm$ 0.036 \\
SD-VICL & \textbf{0.200 $\pm$ 0.029} & \textbf{0.185 $\pm$ 0.031} & 0.225 $\pm$ 0.030 & \textbf{0.585 $\pm$ 0.059} \\
\midrule
Copy target & 0.360 $\pm$ 0.042 & 0.418 $\pm$ 0.040 & 0.367 $\pm$ 0.033 & 0.615 $\pm$ 0.093 \\
\bottomrule
\end{tabular}
\end{table}

\begin{table}[h]
\centering
\footnotesize
\renewcommand{\arraystretch}{0.88}
\setlength{\tabcolsep}{4pt}
\caption{Flip results by dataset (PSNR $\uparrow$). Mean $\pm$ standard deviation.}
\label{tab:flip_psnr_p1}
\resizebox{\textwidth}{!}{%
\begin{tabular}{lcccccccc}
\toprule
\textbf{Model} & \textbf{AP-10K} & \textbf{ATLANTIS} & \textbf{Caltech Birds} & \textbf{CDD} & \textbf{DeepCrack} & \textbf{DRAM} & \textbf{Duke} & \textbf{IXI (Axial)} \\
\midrule
DELVM 1B & 10.780 $\pm$ 2.930 & 10.546 $\pm$ 3.216 & 9.614 $\pm$ 3.580 & 8.944 $\pm$ 2.997 & 15.612 $\pm$ 4.431 & 12.384 $\pm$ 2.303 & 14.351 $\pm$ 2.836 & 14.914 $\pm$ 2.037 \\
LVM & 10.287 $\pm$ 2.550 & 12.241 $\pm$ 2.896 & 11.314 $\pm$ 2.938 & 11.206 $\pm$ 2.659 & 16.396 $\pm$ 4.171 & 12.209 $\pm$ 2.316 & 13.305 $\pm$ 3.090 & 14.724 $\pm$ 1.553 \\
MAE-VQGAN & \textbf{12.886 $\pm$ 2.094} & \textbf{13.552 $\pm$ 2.403} & \textbf{13.575 $\pm$ 2.505} & \textbf{12.738 $\pm$ 2.010} & 15.930 $\pm$ 2.815 & \textbf{13.244 $\pm$ 2.041} & \textbf{14.889 $\pm$ 1.663} & 15.159 $\pm$ 1.282 \\
Painter & 10.840 $\pm$ 2.778 & 11.415 $\pm$ 3.142 & 10.431 $\pm$ 3.598 & 10.625 $\pm$ 2.693 & \textbf{17.533 $\pm$ 3.579} & 12.230 $\pm$ 2.141 & 13.512 $\pm$ 1.742 & 13.832 $\pm$ 1.209 \\
PromptDiff. & 11.468 $\pm$ 1.794 & 10.434 $\pm$ 1.525 & 10.875 $\pm$ 1.966 & 10.105 $\pm$ 1.308 & 11.834 $\pm$ 3.253 & 9.859 $\pm$ 1.387 & 7.981 $\pm$ 0.462 & 7.550 $\pm$ 0.396 \\
SD-VICL & 9.184 $\pm$ 1.528 & 9.231 $\pm$ 1.729 & 9.905 $\pm$ 2.254 & 9.421 $\pm$ 1.601 & 12.126 $\pm$ 2.372 & 8.975 $\pm$ 1.744 & 13.985 $\pm$ 2.365 & \textbf{16.257 $\pm$ 1.436} \\
\midrule
Copy target & 9.509 $\pm$ 1.575 & 9.117 $\pm$ 1.627 & 9.958 $\pm$ 2.123 & 9.378 $\pm$ 1.630 & 14.206 $\pm$ 2.747 & 9.347 $\pm$ 1.761 & 12.301 $\pm$ 2.110 & 13.417 $\pm$ 1.120 \\
\bottomrule
\end{tabular}
}
\end{table}

\begin{table}[h]
\centering
\footnotesize
\renewcommand{\arraystretch}{0.88}
\setlength{\tabcolsep}{4pt}
\caption{Flip results by dataset (PSNR $\uparrow$) . Mean $\pm$ standard deviation.}
\label{tab:flip_psnr_p2}
\resizebox{\textwidth}{!}{%
\begin{tabular}{lcccccccc}
\toprule
\textbf{Model} & \textbf{IXI (Coronal)} & \textbf{IXI (Sagittal)} & \textbf{LLVIP} & \textbf{NYU-D V2} & \textbf{OTP} & \textbf{RORD} & \textbf{Sea-Thru} & \textbf{SUIM} \\
\midrule
DELVM 1B & 11.157 $\pm$ 1.454 & 14.748 $\pm$ 2.447 & 11.964 $\pm$ 2.214 & 10.497 $\pm$ 2.397 & 15.189 $\pm$ 2.298 & 9.889 $\pm$ 2.878 & 20.055 $\pm$ 6.667 & 12.416 $\pm$ 3.870 \\
LVM & 11.751 $\pm$ 0.940 & 15.061 $\pm$ 1.841 & 12.272 $\pm$ 2.040 & 9.598 $\pm$ 2.405 & 14.782 $\pm$ 1.987 & 11.731 $\pm$ 3.028 & 23.867 $\pm$ 4.376 & 12.249 $\pm$ 3.640 \\
MAE-VQGAN & \textbf{13.467 $\pm$ 0.775} & 14.996 $\pm$ 0.952 & \textbf{14.765 $\pm$ 1.826} & \textbf{11.178 $\pm$ 1.901} & \textbf{15.399 $\pm$ 2.183} & \textbf{13.552 $\pm$ 2.390} & \textbf{25.367 $\pm$ 3.302} & \textbf{14.576 $\pm$ 2.829} \\
Painter & 11.804 $\pm$ 0.667 & 15.065 $\pm$ 1.075 & 10.805 $\pm$ 1.672 & 9.159 $\pm$ 1.559 & 13.885 $\pm$ 2.352 & 11.717 $\pm$ 2.803 & 9.535 $\pm$ 2.083 & 12.292 $\pm$ 4.020 \\
PromptDiff. & 8.593 $\pm$ 0.431 & 9.442 $\pm$ 0.169 & 9.670 $\pm$ 1.293 & 9.464 $\pm$ 1.385 & 13.034 $\pm$ 1.959 & 11.124 $\pm$ 1.507 & 10.721 $\pm$ 0.533 & 10.212 $\pm$ 1.506 \\
SD-VICL & 13.288 $\pm$ 0.799 & \textbf{16.554 $\pm$ 1.167} & 11.793 $\pm$ 1.894 & 10.003 $\pm$ 1.746 & 12.892 $\pm$ 3.842 & 9.199 $\pm$ 1.874 & 25.095 $\pm$ 2.945 & 9.780 $\pm$ 1.846 \\
\midrule
Copy target & 12.199 $\pm$ 0.674 & 13.692 $\pm$ 0.825 & 12.150 $\pm$ 2.102 & 8.930 $\pm$ 1.420 & 13.224 $\pm$ 4.048 & 9.576 $\pm$ 1.834 & 25.947 $\pm$ 2.798 & 10.467 $\pm$ 1.964 \\
\bottomrule
\end{tabular}
}
\end{table}

\begin{table}[h]
\centering
\footnotesize
\renewcommand{\arraystretch}{0.88}
\setlength{\tabcolsep}{4pt}
\caption{Flip results by dataset (SSIM $\uparrow$). Mean $\pm$ standard deviation.}
\label{tab:flip_ssim_p1}
\resizebox{\textwidth}{!}{%
\begin{tabular}{lcccccccc}
\toprule
\textbf{Model} & \textbf{AP-10K} & \textbf{ATLANTIS} & \textbf{Caltech Birds} & \textbf{CDD} & \textbf{DeepCrack} & \textbf{DRAM} & \textbf{Duke} & \textbf{IXI (Axial)} \\
\midrule
DELVM 1B & 0.234 $\pm$ 0.146 & 0.300 $\pm$ 0.182 & 0.286 $\pm$ 0.228 & 0.301 $\pm$ 0.184 & 0.307 $\pm$ 0.207 & 0.251 $\pm$ 0.114 & 0.315 $\pm$ 0.075 & 0.440 $\pm$ 0.078 \\
LVM & 0.241 $\pm$ 0.135 & 0.329 $\pm$ 0.176 & 0.382 $\pm$ 0.191 & 0.329 $\pm$ 0.164 & 0.319 $\pm$ 0.209 & 0.260 $\pm$ 0.114 & 0.302 $\pm$ 0.087 & 0.426 $\pm$ 0.069 \\
MAE-VQGAN & \textbf{0.307 $\pm$ 0.145} & \textbf{0.372 $\pm$ 0.172} & \textbf{0.453 $\pm$ 0.193} & \textbf{0.377 $\pm$ 0.161} & \textbf{0.341 $\pm$ 0.203} & \textbf{0.307 $\pm$ 0.122} & \textbf{0.335 $\pm$ 0.075} & 0.453 $\pm$ 0.050 \\
Painter & 0.227 $\pm$ 0.139 & 0.297 $\pm$ 0.180 & 0.293 $\pm$ 0.215 & 0.331 $\pm$ 0.167 & 0.283 $\pm$ 0.200 & 0.248 $\pm$ 0.114 & 0.273 $\pm$ 0.041 & 0.205 $\pm$ 0.057 \\
PromptDiff. & 0.249 $\pm$ 0.115 & 0.241 $\pm$ 0.112 & 0.322 $\pm$ 0.135 & 0.257 $\pm$ 0.109 & 0.202 $\pm$ 0.120 & 0.263 $\pm$ 0.092 & 0.179 $\pm$ 0.023 & 0.051 $\pm$ 0.013 \\
SD-VICL & 0.139 $\pm$ 0.087 & 0.202 $\pm$ 0.132 & 0.278 $\pm$ 0.158 & 0.240 $\pm$ 0.144 & 0.090 $\pm$ 0.083 & 0.122 $\pm$ 0.068 & 0.277 $\pm$ 0.074 & \textbf{0.503 $\pm$ 0.067} \\
\midrule
Copy target & 0.162 $\pm$ 0.082 & 0.200 $\pm$ 0.112 & 0.266 $\pm$ 0.136 & 0.246 $\pm$ 0.129 & 0.191 $\pm$ 0.120 & 0.160 $\pm$ 0.064 & 0.269 $\pm$ 0.055 & 0.429 $\pm$ 0.052 \\
\bottomrule
\end{tabular}
}
\end{table}

\begin{table}[h]
\centering
\footnotesize
\renewcommand{\arraystretch}{0.88}
\setlength{\tabcolsep}{4pt}
\caption{Flip results by dataset (SSIM $\uparrow$). Mean $\pm$ standard deviation.}
\label{tab:flip_ssim_p2}
\resizebox{\textwidth}{!}{%
\begin{tabular}{lcccccccc}
\toprule
\textbf{Model} & \textbf{IXI (Coronal)} & \textbf{IXI (Sagittal)} & \textbf{LLVIP} & \textbf{NYU-D V2} & \textbf{OTP} & \textbf{RORD} & \textbf{Sea-Thru} & \textbf{SUIM} \\
\midrule
DELVM 1B & 0.163 $\pm$ 0.050 & 0.372 $\pm$ 0.100 & 0.412 $\pm$ 0.166 & 0.343 $\pm$ 0.105 & 0.728 $\pm$ 0.235 & 0.291 $\pm$ 0.214 & 0.366 $\pm$ 0.196 & 0.387 $\pm$ 0.218 \\
LVM & 0.157 $\pm$ 0.049 & 0.368 $\pm$ 0.072 & 0.435 $\pm$ 0.129 & 0.321 $\pm$ 0.110 & 0.721 $\pm$ 0.207 & 0.323 $\pm$ 0.221 & 0.442 $\pm$ 0.178 & 0.379 $\pm$ 0.209 \\
MAE-VQGAN & \textbf{0.262 $\pm$ 0.048} & 0.364 $\pm$ 0.043 & \textbf{0.532 $\pm$ 0.141} & 0.374 $\pm$ 0.104 & \textbf{0.739 $\pm$ 0.170} & \textbf{0.381 $\pm$ 0.219} & 0.463 $\pm$ 0.158 & \textbf{0.453 $\pm$ 0.199} \\
Painter & 0.107 $\pm$ 0.023 & 0.270 $\pm$ 0.065 & 0.429 $\pm$ 0.116 & 0.379 $\pm$ 0.087 & 0.696 $\pm$ 0.214 & 0.319 $\pm$ 0.224 & 0.198 $\pm$ 0.057 & 0.317 $\pm$ 0.222 \\
PromptDiff. & 0.067 $\pm$ 0.014 & 0.066 $\pm$ 0.009 & 0.497 $\pm$ 0.105 & \textbf{0.383 $\pm$ 0.101} & 0.312 $\pm$ 0.062 & 0.308 $\pm$ 0.159 & 0.214 $\pm$ 0.040 & 0.259 $\pm$ 0.113 \\
SD-VICL & 0.226 $\pm$ 0.044 & \textbf{0.430 $\pm$ 0.066} & 0.395 $\pm$ 0.131 & 0.354 $\pm$ 0.086 & 0.618 $\pm$ 0.208 & 0.211 $\pm$ 0.155 & \textbf{0.493 $\pm$ 0.106} & 0.249 $\pm$ 0.143 \\
\midrule
Copy target & 0.199 $\pm$ 0.042 & 0.342 $\pm$ 0.039 & 0.427 $\pm$ 0.134 & 0.270 $\pm$ 0.068 & 0.647 $\pm$ 0.240 & 0.221 $\pm$ 0.139 & 0.686 $\pm$ 0.088 & 0.270 $\pm$ 0.143 \\
\bottomrule
\end{tabular}
}
\end{table}

\begin{table}[h]
\centering
\footnotesize
\renewcommand{\arraystretch}{0.88}
\setlength{\tabcolsep}{4pt}
\caption{Flip results by dataset (LPIPS $\downarrow$). Mean $\pm$ standard deviation.}
\label{tab:flip_lpips_p1}
\resizebox{\textwidth}{!}{%
\begin{tabular}{lcccccccc}
\toprule
\textbf{Model} & \textbf{AP-10K} & \textbf{ATLANTIS} & \textbf{Caltech Birds} & \textbf{CDD} & \textbf{DeepCrack} & \textbf{DRAM} & \textbf{Duke} & \textbf{IXI (Axial)} \\
\midrule
DELVM 1B & 0.714 $\pm$ 0.094 & 0.681 $\pm$ 0.125 & 0.741 $\pm$ 0.136 & 0.728 $\pm$ 0.110 & 0.657 $\pm$ 0.126 & 0.646 $\pm$ 0.060 & 0.595 $\pm$ 0.047 & \textbf{0.321 $\pm$ 0.071} \\
LVM & 0.772 $\pm$ 0.098 & \textbf{0.625 $\pm$ 0.107} & 0.723 $\pm$ 0.137 & \textbf{0.724 $\pm$ 0.082} & 0.638 $\pm$ 0.094 & 0.677 $\pm$ 0.085 & 0.599 $\pm$ 0.039 & 0.342 $\pm$ 0.049 \\
MAE-VQGAN & 0.731 $\pm$ 0.064 & 0.679 $\pm$ 0.085 & \textbf{0.647 $\pm$ 0.090} & 0.732 $\pm$ 0.087 & 0.680 $\pm$ 0.068 & 0.685 $\pm$ 0.054 & 0.637 $\pm$ 0.028 & 0.378 $\pm$ 0.035 \\
Painter & \textbf{0.708 $\pm$ 0.091} & 0.644 $\pm$ 0.117 & 0.701 $\pm$ 0.106 & 0.795 $\pm$ 0.117 & \textbf{0.554 $\pm$ 0.042} & \textbf{0.637 $\pm$ 0.060} & 0.590 $\pm$ 0.025 & 0.330 $\pm$ 0.033 \\
PromptDiff. & 0.844 $\pm$ 0.068 & 0.747 $\pm$ 0.066 & 0.806 $\pm$ 0.106 & 0.853 $\pm$ 0.072 & 0.704 $\pm$ 0.049 & 0.833 $\pm$ 0.076 & 0.690 $\pm$ 0.022 & 0.673 $\pm$ 0.030 \\
SD-VICL & 0.765 $\pm$ 0.064 & 0.724 $\pm$ 0.060 & 0.791 $\pm$ 0.090 & 0.740 $\pm$ 0.079 & 0.680 $\pm$ 0.056 & 0.768 $\pm$ 0.060 & \textbf{0.555 $\pm$ 0.027} & 0.347 $\pm$ 0.039 \\
\midrule
Copy target & 0.770 $\pm$ 0.059 & 0.740 $\pm$ 0.057 & 0.789 $\pm$ 0.084 & 0.748 $\pm$ 0.081 & 0.635 $\pm$ 0.057 & 0.766 $\pm$ 0.057 & 0.549 $\pm$ 0.050 & 0.370 $\pm$ 0.039 \\
\bottomrule
\end{tabular}
}
\end{table}

\begin{table}[h]
\centering
\footnotesize
\renewcommand{\arraystretch}{0.88}
\setlength{\tabcolsep}{4pt}
\caption{Flip results by dataset (LPIPS $\downarrow$). Mean $\pm$ standard deviation.}
\label{tab:flip_lpips_p2}
\resizebox{\textwidth}{!}{%
\begin{tabular}{lcccccccc}
\toprule
\textbf{Model} & \textbf{IXI (Coronal)} & \textbf{IXI (Sagittal)} & \textbf{LLVIP} & \textbf{NYU-D V2} & \textbf{OTP} & \textbf{RORD} & \textbf{Sea-Thru} & \textbf{SUIM} \\
\midrule
DELVM 1B & 0.514 $\pm$ 0.090 & 0.375 $\pm$ 0.102 & 0.619 $\pm$ 0.045 & \textbf{0.610 $\pm$ 0.050} & \textbf{0.363 $\pm$ 0.140} & 0.755 $\pm$ 0.124 & 0.679 $\pm$ 0.101 & 0.614 $\pm$ 0.136 \\
LVM & 0.477 $\pm$ 0.053 & 0.361 $\pm$ 0.057 & 0.625 $\pm$ 0.046 & 0.671 $\pm$ 0.095 & 0.387 $\pm$ 0.149 & 0.681 $\pm$ 0.095 & 0.625 $\pm$ 0.081 & 0.635 $\pm$ 0.138 \\
MAE-VQGAN & \textbf{0.431 $\pm$ 0.027} & 0.420 $\pm$ 0.026 & \textbf{0.609 $\pm$ 0.059} & 0.631 $\pm$ 0.046 & 0.405 $\pm$ 0.132 & 0.710 $\pm$ 0.098 & 0.613 $\pm$ 0.078 & \textbf{0.596 $\pm$ 0.115} \\
Painter & 0.457 $\pm$ 0.032 & \textbf{0.309 $\pm$ 0.031} & 0.628 $\pm$ 0.029 & 0.756 $\pm$ 0.066 & 0.379 $\pm$ 0.138 & \textbf{0.668 $\pm$ 0.090} & 0.632 $\pm$ 0.033 & 0.614 $\pm$ 0.121 \\
PromptDiff. & 0.703 $\pm$ 0.027 & 0.753 $\pm$ 0.021 & 0.624 $\pm$ 0.034 & 0.754 $\pm$ 0.067 & 0.795 $\pm$ 0.035 & 0.769 $\pm$ 0.083 & 0.815 $\pm$ 0.037 & 0.814 $\pm$ 0.060 \\
SD-VICL & 0.520 $\pm$ 0.028 & 0.391 $\pm$ 0.038 & 0.635 $\pm$ 0.048 & 0.675 $\pm$ 0.052 & 0.515 $\pm$ 0.122 & 0.770 $\pm$ 0.072 & \textbf{0.609 $\pm$ 0.044} & 0.728 $\pm$ 0.094 \\
\midrule
Copy target & 0.416 $\pm$ 0.038 & 0.373 $\pm$ 0.032 & 0.614 $\pm$ 0.093 & 0.675 $\pm$ 0.043 & 0.441 $\pm$ 0.194 & 0.773 $\pm$ 0.071 & 0.551 $\pm$ 0.032 & 0.722 $\pm$ 0.093 \\
\bottomrule
\end{tabular}
}
\end{table}

\begin{table}[h]
\centering
\footnotesize
\renewcommand{\arraystretch}{0.88}
\setlength{\tabcolsep}{4pt}
\caption{Rotation results by dataset (PSNR $\uparrow$). Mean $\pm$ standard deviation.}
\label{tab:rotation_psnr_p1}
\resizebox{\textwidth}{!}{%
\begin{tabular}{lcccccccc}
\toprule
\textbf{Model} & \textbf{AP-10K} & \textbf{ATLANTIS} & \textbf{Caltech Birds} & \textbf{CDD} & \textbf{DeepCrack} & \textbf{DRAM} & \textbf{Duke} & \textbf{IXI (Axial)} \\
\midrule
DELVM 1B & 9.835 $\pm$ 2.583 & 8.655 $\pm$ 2.264 & 9.271 $\pm$ 3.384 & 7.357 $\pm$ 2.219 & 15.209 $\pm$ 4.627 & \textbf{10.939 $\pm$ 2.155} & 9.944 $\pm$ 2.850 & 12.413 $\pm$ 1.330 \\
LVM & 9.297 $\pm$ 1.963 & 9.422 $\pm$ 1.875 & 10.341 $\pm$ 2.554 & 9.025 $\pm$ 2.017 & 14.976 $\pm$ 3.917 & 9.802 $\pm$ 2.054 & 9.304 $\pm$ 3.145 & 12.674 $\pm$ 1.576 \\
MAE-VQGAN & 9.745 $\pm$ 2.382 & 8.271 $\pm$ 2.163 & \textbf{11.493 $\pm$ 2.650} & 7.947 $\pm$ 2.231 & 13.807 $\pm$ 2.714 & 9.785 $\pm$ 2.121 & 10.340 $\pm$ 1.774 & 13.400 $\pm$ 1.484 \\
Painter & 9.998 $\pm$ 2.488 & 9.022 $\pm$ 2.251 & 10.000 $\pm$ 3.411 & 9.143 $\pm$ 2.496 & \textbf{17.210 $\pm$ 3.633} & 10.925 $\pm$ 1.940 & 9.646 $\pm$ 1.123 & 11.312 $\pm$ 1.065 \\
PromptDiff. & \textbf{11.449 $\pm$ 1.778} & \textbf{9.828 $\pm$ 1.519} & 10.811 $\pm$ 1.962 & \textbf{9.442 $\pm$ 1.446} & 11.786 $\pm$ 3.198 & 9.693 $\pm$ 1.373 & 7.499 $\pm$ 0.636 & 7.180 $\pm$ 0.324 \\
SD-VICL & 8.995 $\pm$ 1.519 & 8.766 $\pm$ 1.625 & 9.816 $\pm$ 2.217 & 9.022 $\pm$ 1.446 & 12.121 $\pm$ 2.401 & 8.733 $\pm$ 1.748 & \textbf{12.723 $\pm$ 1.564} & \textbf{14.216 $\pm$ 1.260} \\
\midrule
Copy target & 9.509 $\pm$ 1.575 & 9.117 $\pm$ 1.627 & 9.958 $\pm$ 2.123 & 9.378 $\pm$ 1.630 & 14.206 $\pm$ 2.747 & 9.347 $\pm$ 1.761 & 12.301 $\pm$ 2.110 & 13.417 $\pm$ 1.120 \\
\bottomrule
\end{tabular}
}
\end{table}

\begin{table}[h]
\centering
\footnotesize
\renewcommand{\arraystretch}{0.88}
\setlength{\tabcolsep}{4pt}
\caption{Rotation results by dataset (PSNR $\uparrow$). Mean $\pm$ standard deviation.}
\label{tab:rotation_psnr_p2}
\resizebox{\textwidth}{!}{%
\begin{tabular}{lcccccccc}
\toprule
\textbf{Model} & \textbf{IXI (Coronal)} & \textbf{IXI (Sagittal)} & \textbf{LLVIP} & \textbf{NYU-D V2} & \textbf{OTP} & \textbf{RORD} & \textbf{Sea-Thru} & \textbf{SUIM} \\
\midrule
DELVM 1B & 10.246 $\pm$ 1.078 & 11.157 $\pm$ 1.260 & 11.640 $\pm$ 2.122 & 9.527 $\pm$ 2.203 & \textbf{14.200 $\pm$ 3.177} & 9.190 $\pm$ 2.575 & 19.984 $\pm$ 6.653 & 11.003 $\pm$ 3.151 \\
LVM & 10.671 $\pm$ 1.481 & 11.422 $\pm$ 1.591 & \textbf{11.761 $\pm$ 1.940} & 8.416 $\pm$ 1.600 & 13.722 $\pm$ 3.085 & 9.702 $\pm$ 2.117 & 20.969 $\pm$ 6.656 & 10.352 $\pm$ 2.631 \\
MAE-VQGAN & 11.773 $\pm$ 0.883 & \textbf{13.635 $\pm$ 1.028} & 10.902 $\pm$ 2.044 & 7.637 $\pm$ 2.155 & 13.624 $\pm$ 3.475 & 9.026 $\pm$ 2.245 & 24.213 $\pm$ 3.054 & \textbf{11.247 $\pm$ 2.749} \\
Painter & 10.790 $\pm$ 0.706 & 10.785 $\pm$ 0.780 & 10.486 $\pm$ 1.655 & 8.874 $\pm$ 1.567 & 13.088 $\pm$ 2.580 & 10.632 $\pm$ 2.584 & 9.494 $\pm$ 2.154 & 10.914 $\pm$ 3.303 \\
PromptDiff. & 7.890 $\pm$ 0.541 & 9.413 $\pm$ 0.168 & 10.000 $\pm$ 1.325 & 9.210 $\pm$ 1.359 & 13.099 $\pm$ 1.842 & \textbf{10.754 $\pm$ 1.443} & 10.728 $\pm$ 0.543 & 9.932 $\pm$ 1.640 \\
SD-VICL & \textbf{12.056 $\pm$ 0.804} & 12.999 $\pm$ 1.043 & 11.706 $\pm$ 1.800 & \textbf{9.914 $\pm$ 1.750} & 13.039 $\pm$ 4.241 & 9.009 $\pm$ 1.694 & \textbf{25.326 $\pm$ 2.928} & 9.664 $\pm$ 1.861 \\
\midrule
Copy target & 12.199 $\pm$ 0.674 & 13.692 $\pm$ 0.825 & 12.150 $\pm$ 2.102 & 8.930 $\pm$ 1.420 & 13.224 $\pm$ 4.048 & 9.576 $\pm$ 1.834 & 25.947 $\pm$ 2.798 & 10.467 $\pm$ 1.964 \\
\bottomrule
\end{tabular}
}
\end{table}

\begin{table}[h]
\centering
\footnotesize
\renewcommand{\arraystretch}{0.88}
\setlength{\tabcolsep}{4pt}
\caption{Rotation results by dataset (SSIM $\uparrow$). Mean $\pm$ standard deviation.}
\label{tab:rotation_ssim_p1}
\resizebox{\textwidth}{!}{%
\begin{tabular}{lcccccccc}
\toprule
\textbf{Model} & \textbf{AP-10K} & \textbf{ATLANTIS} & \textbf{Caltech Birds} & \textbf{CDD} & \textbf{DeepCrack} & \textbf{DRAM} & \textbf{Duke} & \textbf{IXI (Axial)} \\
\midrule
DELVM 1B & 0.210 $\pm$ 0.139 & 0.251 $\pm$ 0.166 & 0.267 $\pm$ 0.223 & 0.250 $\pm$ 0.155 & 0.310 $\pm$ 0.211 & 0.213 $\pm$ 0.101 & 0.224 $\pm$ 0.075 & 0.337 $\pm$ 0.077 \\
LVM & 0.235 $\pm$ 0.126 & \textbf{0.301 $\pm$ 0.150} & 0.375 $\pm$ 0.184 & \textbf{0.311 $\pm$ 0.147} & 0.314 $\pm$ 0.197 & 0.252 $\pm$ 0.109 & 0.221 $\pm$ 0.076 & 0.339 $\pm$ 0.084 \\
MAE-VQGAN & \textbf{0.257 $\pm$ 0.135} & 0.252 $\pm$ 0.148 & \textbf{0.424 $\pm$ 0.191} & 0.294 $\pm$ 0.142 & \textbf{0.344 $\pm$ 0.202} & 0.251 $\pm$ 0.112 & 0.218 $\pm$ 0.065 & \textbf{0.429 $\pm$ 0.051} \\
Painter & 0.204 $\pm$ 0.129 & 0.224 $\pm$ 0.152 & 0.272 $\pm$ 0.208 & 0.286 $\pm$ 0.159 & 0.274 $\pm$ 0.197 & 0.211 $\pm$ 0.102 & 0.198 $\pm$ 0.026 & 0.129 $\pm$ 0.034 \\
PromptDiff. & 0.245 $\pm$ 0.114 & 0.229 $\pm$ 0.107 & 0.317 $\pm$ 0.133 & 0.239 $\pm$ 0.098 & 0.200 $\pm$ 0.121 & \textbf{0.258 $\pm$ 0.090} & 0.157 $\pm$ 0.016 & 0.049 $\pm$ 0.011 \\
SD-VICL & 0.132 $\pm$ 0.085 & 0.180 $\pm$ 0.118 & 0.268 $\pm$ 0.157 & 0.234 $\pm$ 0.135 & 0.086 $\pm$ 0.073 & 0.111 $\pm$ 0.063 & \textbf{0.229 $\pm$ 0.045} & 0.410 $\pm$ 0.051 \\
\midrule
Copy target & 0.162 $\pm$ 0.082 & 0.200 $\pm$ 0.112 & 0.266 $\pm$ 0.136 & 0.246 $\pm$ 0.129 & 0.191 $\pm$ 0.120 & 0.160 $\pm$ 0.064 & 0.269 $\pm$ 0.055 & 0.429 $\pm$ 0.052 \\
\bottomrule
\end{tabular}
}
\end{table}

\begin{table}[h]
\centering
\footnotesize
\renewcommand{\arraystretch}{0.88}
\setlength{\tabcolsep}{4pt}
\caption{Rotation results by dataset (SSIM $\uparrow$). Mean $\pm$ standard deviation.}
\label{tab:rotation_ssim_p2}
\resizebox{\textwidth}{!}{%
\begin{tabular}{lcccccccc}
\toprule
\textbf{Model} & \textbf{IXI (Coronal)} & \textbf{IXI (Sagittal)} & \textbf{LLVIP} & \textbf{NYU-D V2} & \textbf{OTP} & \textbf{RORD} & \textbf{Sea-Thru} & \textbf{SUIM} \\
\midrule
DELVM 1B & 0.135 $\pm$ 0.053 & 0.189 $\pm$ 0.059 & 0.411 $\pm$ 0.159 & 0.308 $\pm$ 0.101 & 0.691 $\pm$ 0.277 & 0.288 $\pm$ 0.222 & 0.385 $\pm$ 0.189 & 0.351 $\pm$ 0.216 \\
LVM & 0.106 $\pm$ 0.042 & 0.183 $\pm$ 0.057 & 0.422 $\pm$ 0.124 & 0.310 $\pm$ 0.095 & 0.675 $\pm$ 0.255 & \textbf{0.318 $\pm$ 0.192} & 0.411 $\pm$ 0.159 & 0.356 $\pm$ 0.180 \\
MAE-VQGAN & \textbf{0.214 $\pm$ 0.040} & \textbf{0.326 $\pm$ 0.042} & 0.243 $\pm$ 0.169 & 0.221 $\pm$ 0.087 & \textbf{0.704 $\pm$ 0.217} & 0.307 $\pm$ 0.206 & 0.354 $\pm$ 0.141 & \textbf{0.389 $\pm$ 0.187} \\
Painter & 0.060 $\pm$ 0.016 & 0.088 $\pm$ 0.014 & 0.415 $\pm$ 0.118 & 0.367 $\pm$ 0.085 & 0.674 $\pm$ 0.240 & 0.293 $\pm$ 0.219 & 0.194 $\pm$ 0.059 & 0.278 $\pm$ 0.210 \\
PromptDiff. & 0.054 $\pm$ 0.012 & 0.065 $\pm$ 0.010 & \textbf{0.499 $\pm$ 0.104} & \textbf{0.373 $\pm$ 0.098} & 0.318 $\pm$ 0.061 & 0.297 $\pm$ 0.153 & 0.213 $\pm$ 0.041 & 0.250 $\pm$ 0.107 \\
SD-VICL & 0.193 $\pm$ 0.034 & 0.300 $\pm$ 0.038 & 0.383 $\pm$ 0.132 & 0.353 $\pm$ 0.086 & 0.615 $\pm$ 0.235 & 0.197 $\pm$ 0.153 & \textbf{0.497 $\pm$ 0.108} & 0.241 $\pm$ 0.141 \\
\midrule
Copy target & 0.199 $\pm$ 0.042 & 0.342 $\pm$ 0.039 & 0.427 $\pm$ 0.134 & 0.270 $\pm$ 0.068 & 0.647 $\pm$ 0.240 & 0.221 $\pm$ 0.139 & 0.686 $\pm$ 0.088 & 0.270 $\pm$ 0.143 \\
\bottomrule
\end{tabular}
}
\end{table}

\begin{table}[h!]
\centering
\footnotesize
\renewcommand{\arraystretch}{0.88}
\setlength{\tabcolsep}{4pt}
\caption{Rotation results by dataset (LPIPS $\downarrow$). Mean $\pm$ standard deviation.}
\label{tab:rotation_lpips_p1}
\resizebox{\textwidth}{!}{%
\begin{tabular}{lcccccccc}
\toprule
\textbf{Model} & \textbf{AP-10K} & \textbf{ATLANTIS} & \textbf{Caltech Birds} & \textbf{CDD} & \textbf{DeepCrack} & \textbf{DRAM} & \textbf{Duke} & \textbf{IXI (Axial)} \\
\midrule
DELVM 1B & 0.744 $\pm$ 0.077 & 0.749 $\pm$ 0.087 & 0.760 $\pm$ 0.119 & 0.779 $\pm$ 0.082 & 0.676 $\pm$ 0.120 & 0.697 $\pm$ 0.059 & 0.643 $\pm$ 0.016 & 0.436 $\pm$ 0.046 \\
LVM & 0.819 $\pm$ 0.075 & 0.777 $\pm$ 0.080 & 0.772 $\pm$ 0.125 & 0.820 $\pm$ 0.074 & 0.685 $\pm$ 0.097 & 0.812 $\pm$ 0.079 & 0.711 $\pm$ 0.077 & 0.450 $\pm$ 0.046 \\
MAE-VQGAN & 0.792 $\pm$ 0.067 & 0.764 $\pm$ 0.073 & \textbf{0.719 $\pm$ 0.100} & 0.794 $\pm$ 0.072 & 0.712 $\pm$ 0.064 & 0.784 $\pm$ 0.066 & 0.643 $\pm$ 0.029 & \textbf{0.433 $\pm$ 0.044} \\
Painter & \textbf{0.742 $\pm$ 0.075} & \textbf{0.729 $\pm$ 0.073} & 0.728 $\pm$ 0.089 & 0.830 $\pm$ 0.081 & \textbf{0.581 $\pm$ 0.041} & \textbf{0.696 $\pm$ 0.058} & 0.666 $\pm$ 0.009 & 0.486 $\pm$ 0.030 \\
PromptDiff. & 0.845 $\pm$ 0.066 & 0.760 $\pm$ 0.059 & 0.810 $\pm$ 0.101 & 0.850 $\pm$ 0.067 & 0.702 $\pm$ 0.050 & 0.837 $\pm$ 0.075 & 0.691 $\pm$ 0.019 & 0.712 $\pm$ 0.024 \\
SD-VICL & 0.770 $\pm$ 0.062 & 0.740 $\pm$ 0.054 & 0.795 $\pm$ 0.089 & \textbf{0.762 $\pm$ 0.067} & 0.692 $\pm$ 0.057 & 0.774 $\pm$ 0.058 & \textbf{0.581 $\pm$ 0.015} & 0.452 $\pm$ 0.032 \\
\midrule
Copy target & 0.771 $\pm$ 0.059 & 0.735 $\pm$ 0.056 & 0.788 $\pm$ 0.083 & 0.744 $\pm$ 0.079 & 0.635 $\pm$ 0.055 & 0.764 $\pm$ 0.056 & 0.533 $\pm$ 0.042 & 0.372 $\pm$ 0.040 \\
\bottomrule
\end{tabular}
}
\end{table}

\begin{table}[h!]
\centering
\footnotesize
\renewcommand{\arraystretch}{0.88}
\setlength{\tabcolsep}{4pt}
\caption{Rotation results by dataset (LPIPS $\downarrow$). Mean $\pm$ standard deviation.}
\label{tab:rotation_lpips_p2}
\resizebox{\textwidth}{!}{%
%
}
\end{table}

\begin{minipage}[h!]{0.3\textwidth}
  \footnotesize
  \renewcommand{\arraystretch}{0.88}
  \setlength{\tabcolsep}{4pt}
  \captionof{table}{Object Removal results by dataset (PSNR $\uparrow$). Mean $\pm$ standard deviation.}
  \label{tab:object_removal_psnr}
  \begin{tabular}{lc}
\toprule
\textbf{Model} & \textbf{RORD} \\
\midrule
DELVM 1B & 14.625 $\pm$ 3.988 \\
LVM & \textbf{15.838 $\pm$ 2.503} \\
MAE-VQGAN & 14.230 $\pm$ 2.058 \\
Painter & 13.506 $\pm$ 4.755 \\
PromptDiff. & 10.675 $\pm$ 1.664 \\
SD-VICL & 12.187 $\pm$ 2.499 \\
\midrule
Copy target & 9.576 $\pm$ 1.834 \\
\bottomrule
\end{tabular}
\end{minipage}\hfill
\begin{minipage}[h!]{0.3\textwidth}
  \footnotesize
  \renewcommand{\arraystretch}{0.88}
  \setlength{\tabcolsep}{4pt}
  \captionof{table}{Object Removal results by dataset (SSIM $\uparrow$). Mean $\pm$ standard deviation.}
  \label{tab:object_removal_ssim}
  \begin{tabular}{lc}
\toprule
\textbf{Model} & \textbf{RORD} \\
\midrule
DELVM 1B & 0.406 $\pm$ 0.230 \\
LVM & 0.423 $\pm$ 0.220 \\
MAE-VQGAN & 0.376 $\pm$ 0.211 \\
Painter & \textbf{0.487 $\pm$ 0.246} \\
PromptDiff. & 0.336 $\pm$ 0.179 \\
SD-VICL & 0.357 $\pm$ 0.158 \\
\midrule
Copy target & 0.221 $\pm$ 0.139 \\
\bottomrule
\end{tabular}
\end{minipage}
´\hfill\begin{minipage}[h!]{0.3\textwidth}
  \footnotesize
  \renewcommand{\arraystretch}{0.88}
  \setlength{\tabcolsep}{4pt}
  \captionof{table}{Object Removal results by dataset (LPIPS $\downarrow$). Mean $\pm$ standard deviation.}
  \label{tab:object_removal_lpips}
  %
\end{minipage}

\clearpage

\subsection{Qualitative results}
In this section, we provide randomly selected qualitative results for different tasks and datasets for all the models we stress-test in~\benchmarkname.

\subsubsection{Perception \& localization tasks}

\begin{figure}[h]
    \scriptsize
    $\phantom{\rotatebox{90}{\scriptsize T}}$
    \hfill
    %

    \centering

    \rotatebox{90}{\scriptsize $\phantom{........}$Target$\phantom{...........}$SD-VICL$\phantom{..........}$Painter$\phantom{.......}$MAE-VQGAN$\phantom{.......}$DeLVM$\phantom{...........}$LVM$\phantom{.......}$PromptDiffusion$\phantom{.......}$Query$\phantom{.................} C_{out}$$\phantom{..............} C_{in}$}
    \hfill
    \includegraphics[width=0.98\textwidth]{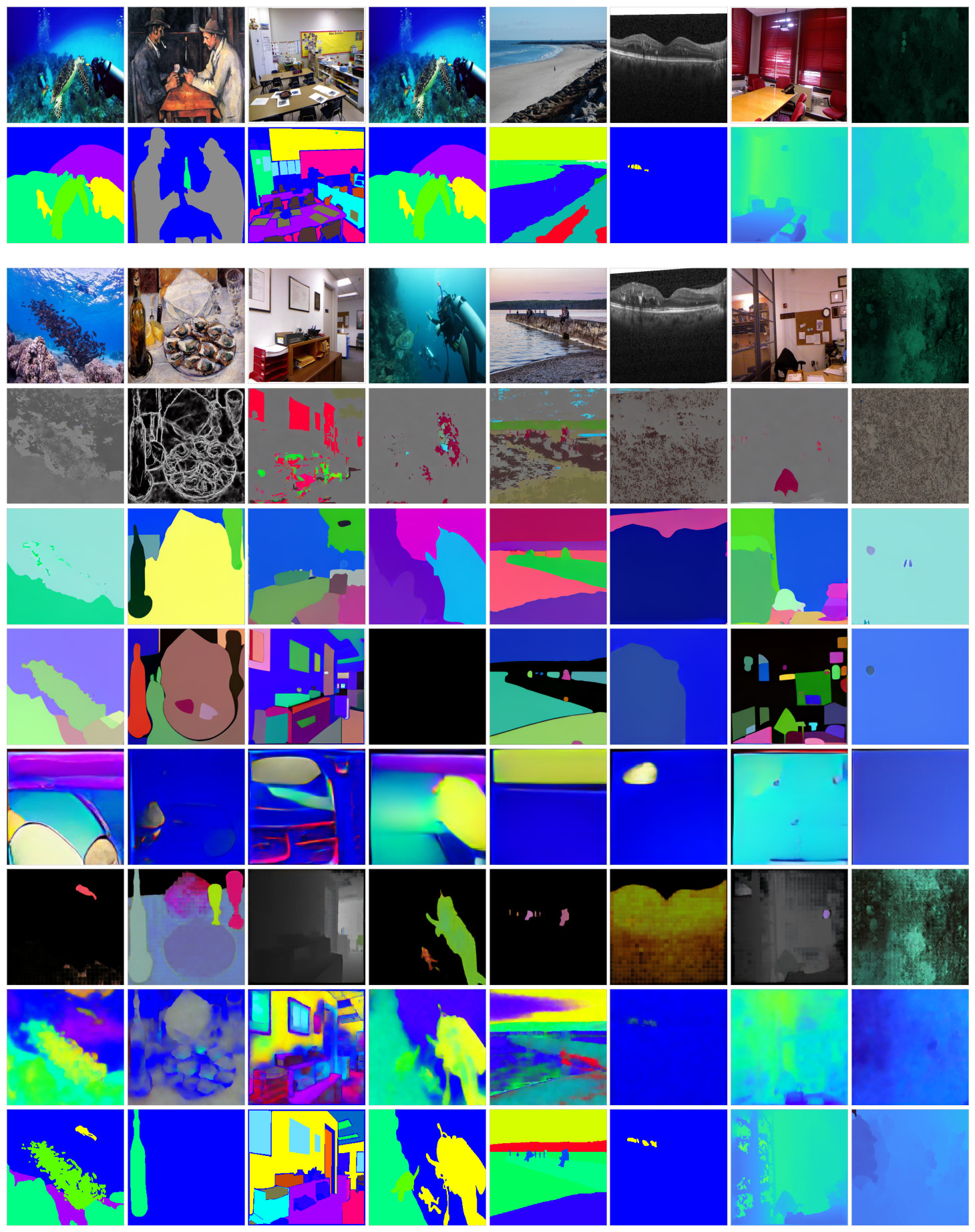}

    \scriptsize
    $\phantom{\rotatebox{90}{\scriptsize T}}$
    \hfill
    %
    
    \caption{Qualitative samples for tasks segmentation and depth estimation.}
    \label{fig:segmentation_depth_appendix}
\end{figure}

\begin{figure}[h]
    \scriptsize
    $\phantom{\rotatebox{90}{\scriptsize T}}$
    \hfill
    %

    \centering

    \rotatebox{90}{\scriptsize $\phantom{........}$Target$\phantom{...........}$SD-VICL$\phantom{..........}$Painter$\phantom{.......}$MAE-VQGAN$\phantom{.....}$DeLVM$\phantom{.............}$LVM$\phantom{.......}$PromptDiffusion$\phantom{.......}$Query$\phantom{................} C_{out}$$\phantom{.............} C_{in}$}
    \hfill
    \includegraphics[width=0.98\textwidth]{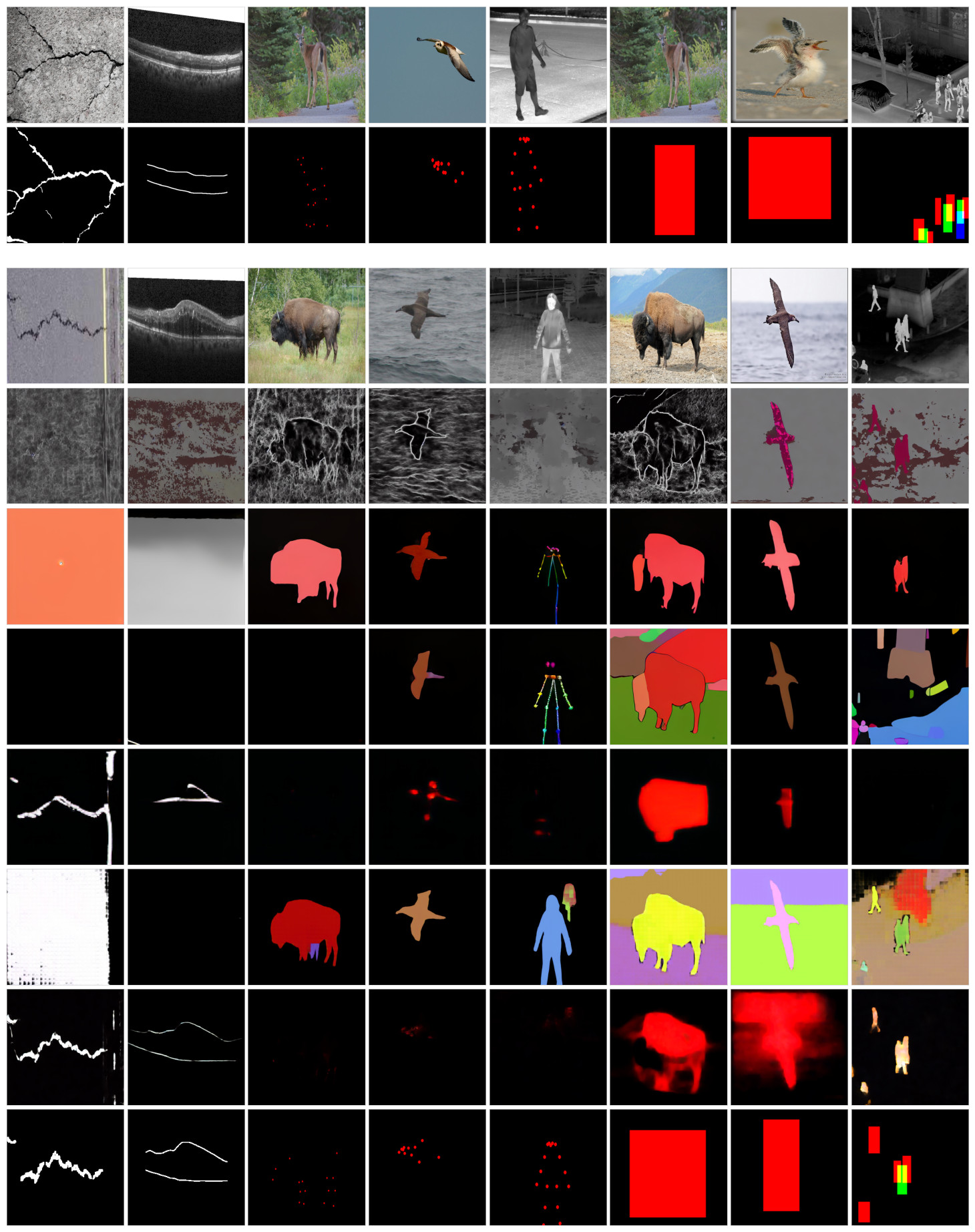}

    \scriptsize
    $\phantom{\rotatebox{90}{\scriptsize T}}$
    \hfill
    %
    \caption{Qualitative samples for edge detection, object detection and keypoint detection.}
    \label{fig:edge_obj_kp_appendix}
\end{figure}

\clearpage
\subsubsection{Image restoration \& enhancement tasks}

\begin{figure}[h]
    \scriptsize
    $\phantom{\rotatebox{90}{\scriptsize T}}$
    \hfill
    %

    \centering

    \rotatebox{90}{\scriptsize $\phantom{........}$Target$\phantom{...........}$SD-VICL$\phantom{..........}$Painter$\phantom{.......}$MAE-VQGAN$\phantom{.......}$DeLVM$\phantom{...........}$LVM$\phantom{.......}$PromptDiffusion$\phantom{.......}$Query$\phantom{.................} C_{out}$$\phantom{.............} C_{in}$}
    \hfill
    \includegraphics[width=0.98\textwidth]{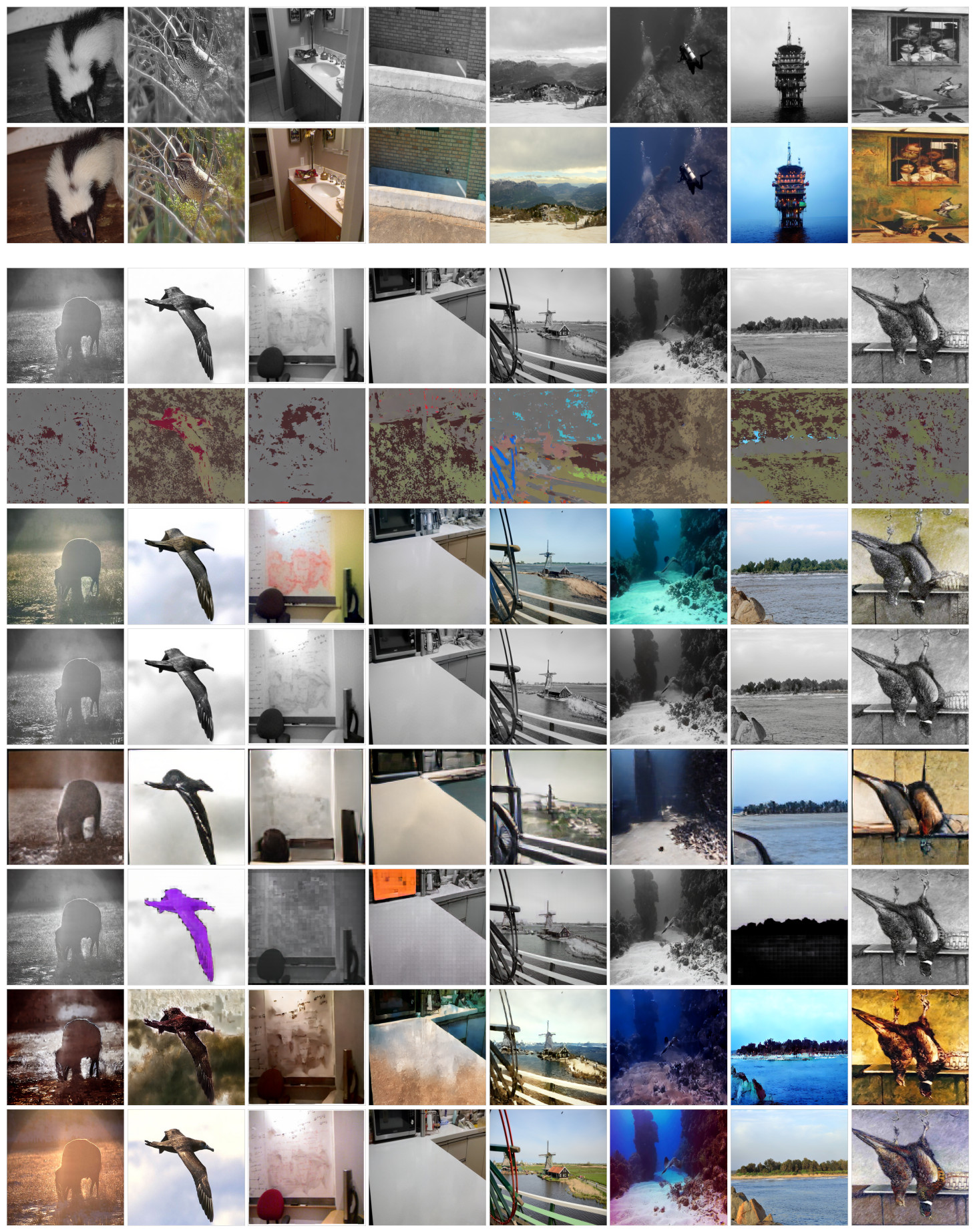}

    \caption{Qualitative samples for the colorization task covering different domains.}
    \label{fig:colorization_appendix}
\end{figure}

\clearpage
\begin{figure}[h]
    \scriptsize
    $\phantom{\rotatebox{90}{\scriptsize T}}$
    $\phantom{...................................}$IXI-Ax.$\phantom{..............}$Duke$\phantom{..........}$OTP$\phantom{..........}$DRAM

    \centering

    \rotatebox{90}{\scriptsize $\phantom{...........}$Target$\phantom{..............}$SD-VICL$\phantom{...............}$Painter$\phantom{............}$MAE-VQGAN$\phantom{.........}$DeLVM$\phantom{.................}$LVM$\phantom{...........}$PromptDiffusion$\phantom{............}$Query$\phantom{....................}$$C_{out}$$\phantom{..................}$$C_{in}$}
    \includegraphics[width=0.6\textwidth]{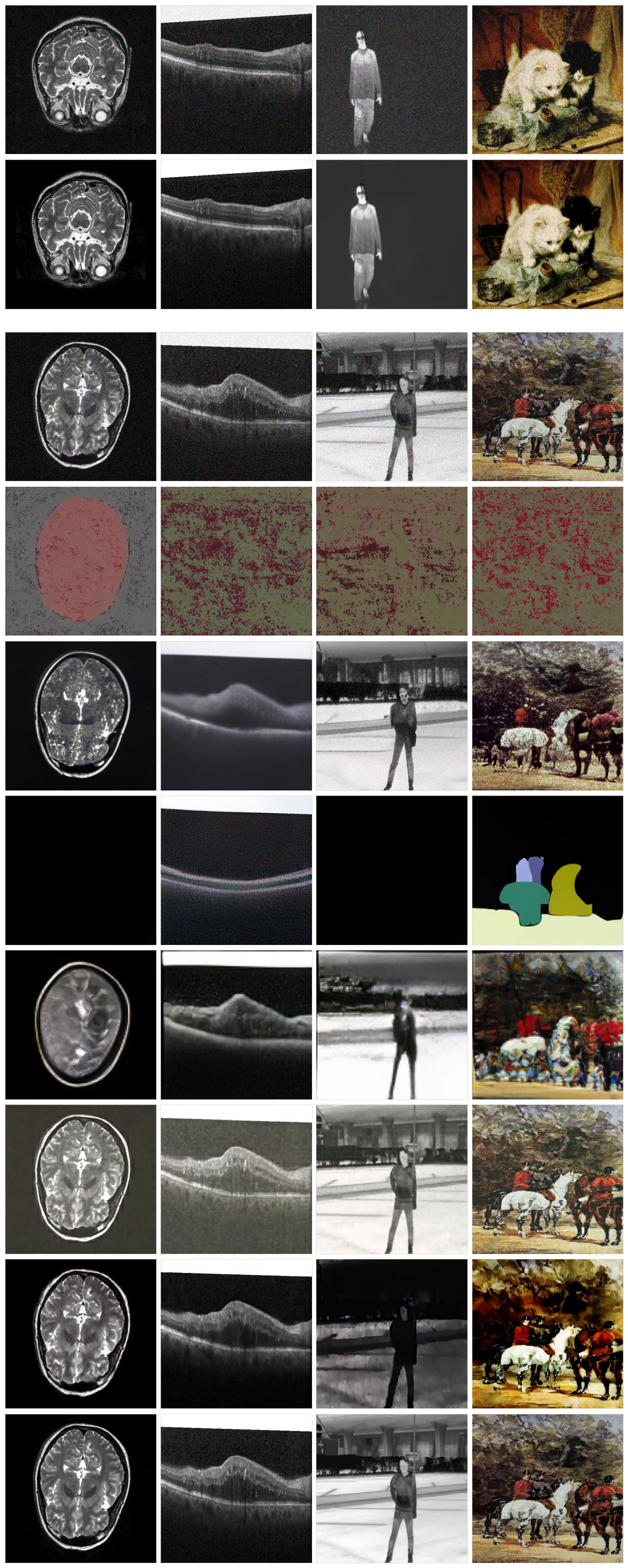}

    \caption{Qualitative samples for the denoising task covering different domains.}
    \label{fig:denoising_appendix}
\end{figure}

\clearpage

\begin{figure}[h]
    \scriptsize
    $\phantom{\rotatebox{90}{\scriptsize T}}$
    \hfill
    \begin{tabular}{p{0.09\textwidth}p{0.09\textwidth}p{0.09\textwidth}p{0.09\textwidth}p{0.09\textwidth}p{0.09\textwidth}p{0.09\textwidth}p{0.09\textwidth}}
         DRAM & CDD & NYUDv2 & IXI-Ax. & RORD & LLVIP & SUIM & CUB
    \end{tabular}

    \centering

    \rotatebox{90}{\scriptsize $\phantom{........}$Target$\phantom{...........}$SD-VICL$\phantom{..........}$Painter$\phantom{.......}$MAE-VQGAN$\phantom{.......}$DeLVM$\phantom{...........}$LVM$\phantom{.......}$PromptDiffusion$\phantom{.......}$Query$\phantom{................} C_{out}$$\phantom{.............} C_{in}$}
    \hfill
    \includegraphics[width=0.98\textwidth]{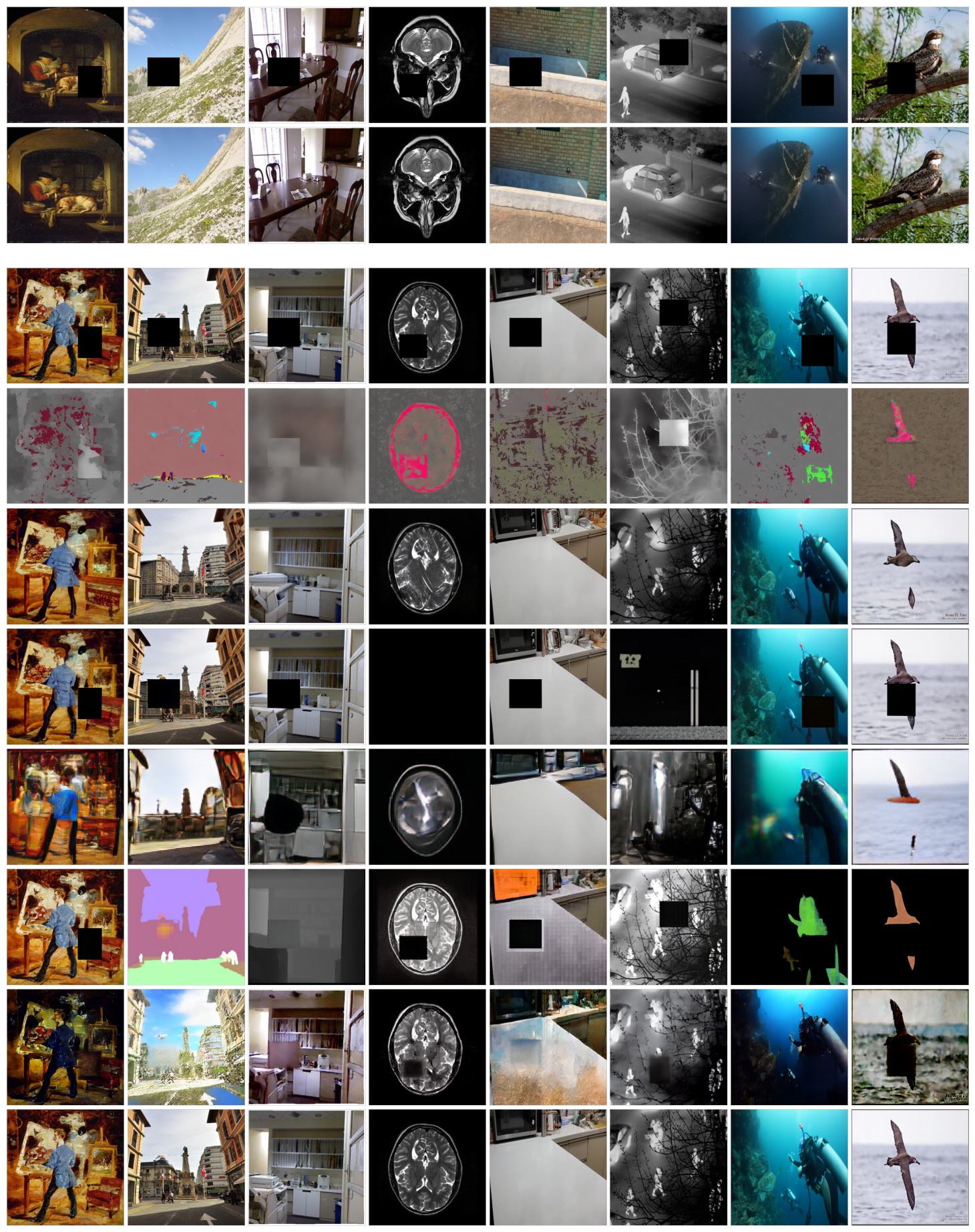}

    \scriptsize
    $\phantom{\rotatebox{90}{\scriptsize T}}$
    \hfill
    \caption{Qualitative samples for the inpainting task covering different domains.}
    \label{fig:inpainting_appendix}
\end{figure}

\clearpage

\begin{figure}[h]
    \scriptsize
    $\phantom{\rotatebox{90}{\scriptsize T}}$

    \centering

    \rotatebox{90}{\scriptsize $\phantom{........}$Target$\phantom{...........}$SD-VICL$\phantom{..........}$Painter$\phantom{......}$MAE-VQGAN$\phantom{.......}$DeLVM$\phantom{...........}$LVM$\phantom{........}$PromptDiffusion$\phantom{.......}$Query$\phantom{................} C_{out}$$\phantom{..............} C_{in}$}
    \includegraphics[width=0.98\textwidth]{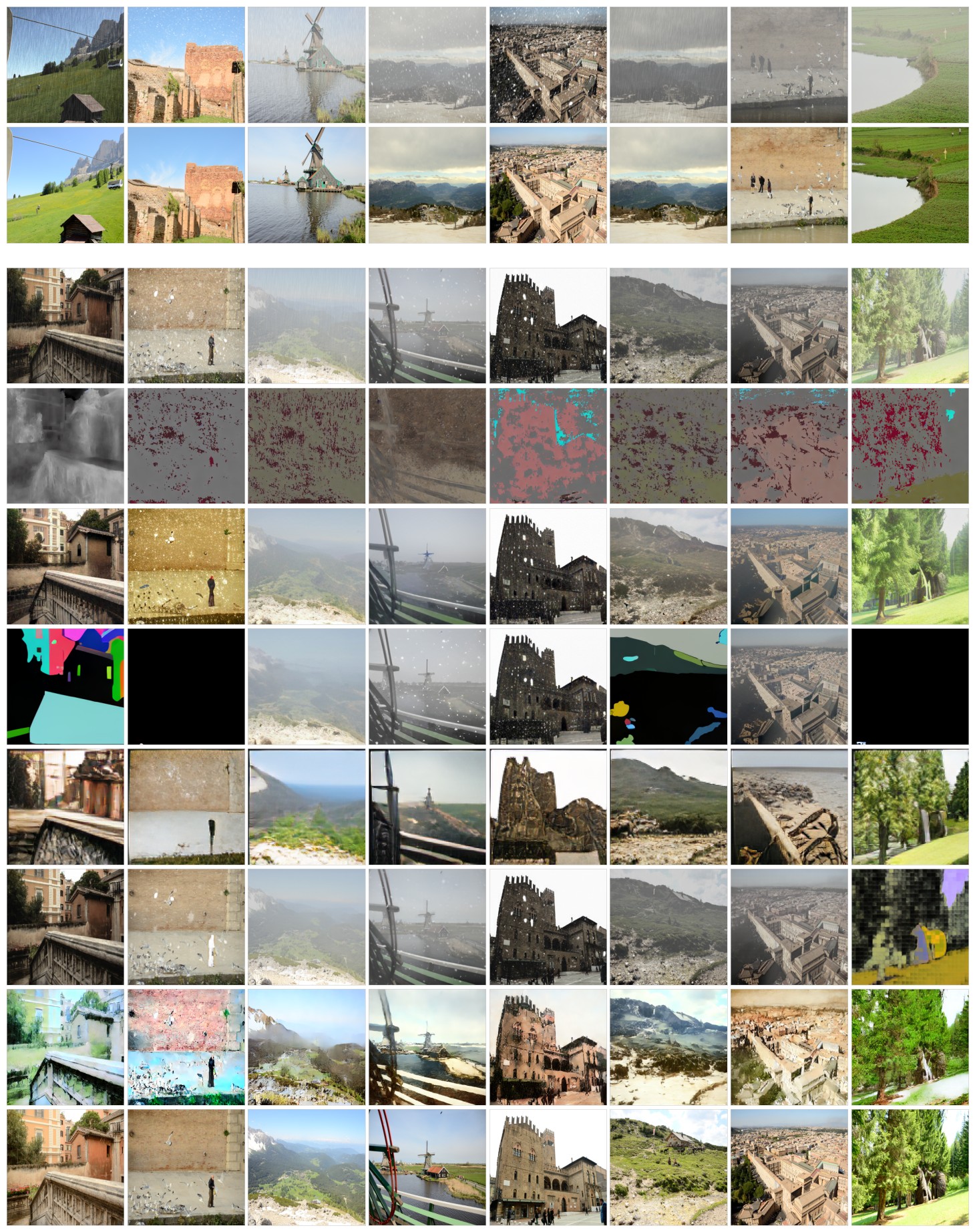}

    \scriptsize
    $\phantom{\rotatebox{90}{\scriptsize T}}$
    \hfill
    \begin{tabular}{p{0.09\textwidth}p{0.09\textwidth}p{0.09\textwidth}p{0.09\textwidth}p{0.09\textwidth}p{0.09\textwidth}p{0.09\textwidth}p{0.09\textwidth}}
         low-rain & snow & haze-rain & l-h-snow & low-snow & l-h-r & low-haze & haze 
    \end{tabular}
    \caption{Qualitative samples for different artifact removal modes in CDD~\cite{guo2024onerestore}.}
    \label{fig:artifact_removal_appendix_qualitative}
\end{figure}

\clearpage

\subsubsection{Image manipulation \& transformation tasks}
\begin{figure}[h]
    \scriptsize
    $\phantom{\rotatebox{90}{\scriptsize T}}$
    \begin{tabular}{p{0.09\textwidth}p{0.09\textwidth}p{0.09\textwidth}p{0.09\textwidth}p{0.09\textwidth}p{0.09\textwidth}p{0.09\textwidth}p{0.09\textwidth}}
         LLVIP & IXI-Sg. & IXI-Ax. & SUIM & CDD & AP10K & DRAM & RORD
    \end{tabular}

    \centering

    \rotatebox{90}{\scriptsize $\phantom{......}$Target$\phantom{...........}$SD-VICL$\phantom{..........}$Painter$\phantom{........}$MAE-VQGAN$\phantom{.......}$DeLVM$\phantom{...........}$LVM$\phantom{........}$PromptDiffusion$\phantom{......}$Query$\phantom{.................} C_{out}$$\phantom{..............} C_{in}$}
    \hfill
    \includegraphics[width=0.98\linewidth]{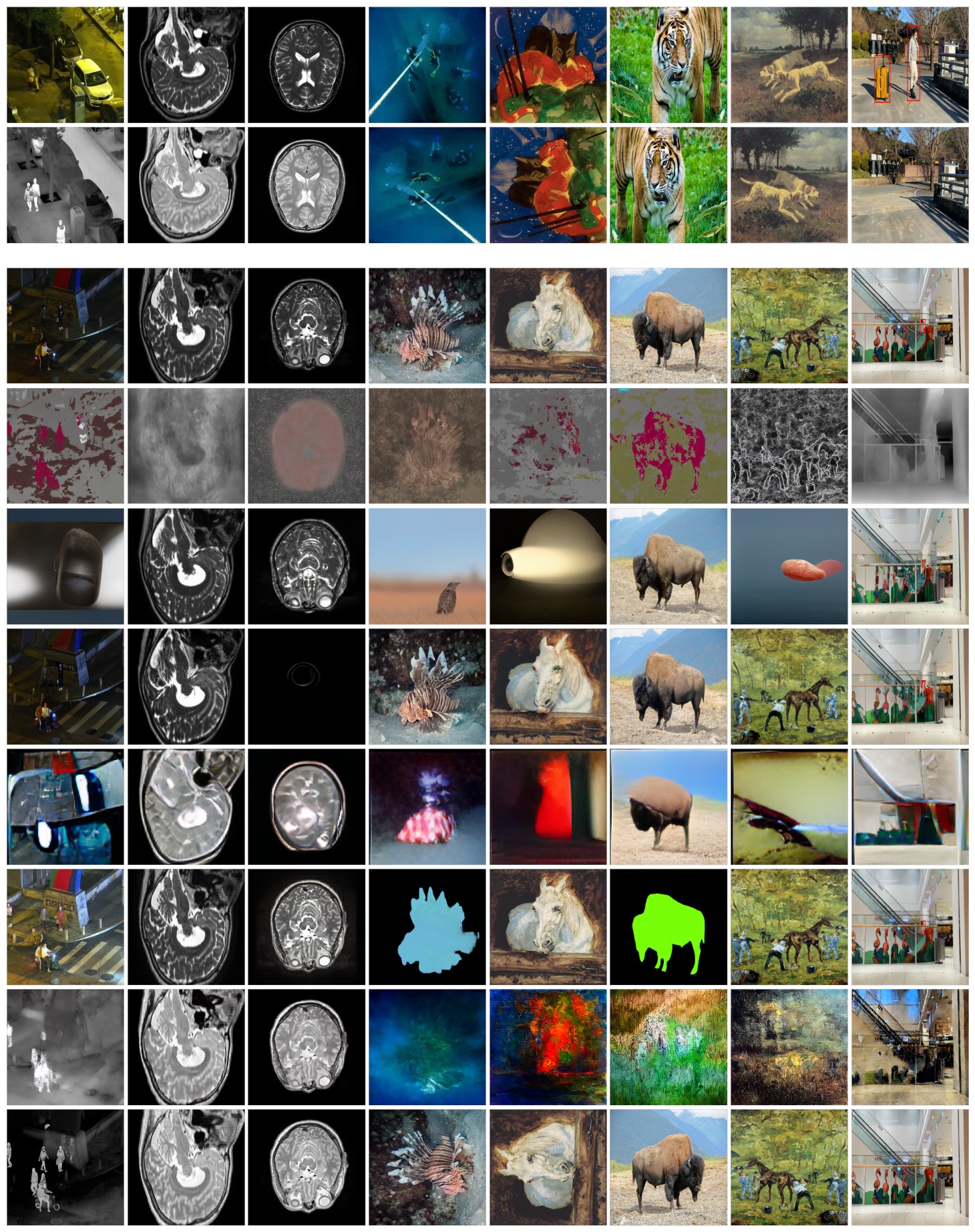}

    \scriptsize
    $\phantom{\rotatebox{90}{\scriptsize T}}$
    \hfill
    \begin{tabular}{p{0.09\textwidth}p{0.09\textwidth}p{0.09\textwidth}p{0.09\textwidth}p{0.09\textwidth}p{0.09\textwidth}p{0.09\textwidth}p{0.09\textwidth}}
        RGB-IR & T2-PD & T2-PD & Rot. & Rot. & Flip & Flip & Obj.rem. 
    \end{tabular}
    \caption{Qualitative samples for modality transfer, rotation, flip and object removal.}
    \label{fig:image_manipulation_and_transformation_qualitative}
\end{figure}

\clearpage

\end{document}